\documentclass[lettersize,journal]{IEEEtran}
\usepackage{amsmath,amsfonts}
\usepackage{algorithmic}
\usepackage{algorithm}
\usepackage{array}
\usepackage[caption=false,font=normalsize,labelfont=sf,textfont=sf]{subfig}
\usepackage{textcomp}
\usepackage{stfloats}
\usepackage{url}
\usepackage{verbatim}
\usepackage{graphicx}
\usepackage{cite}
\usepackage{booktabs}
\usepackage{makecell}
\usepackage{multirow}
\usepackage{overpic}
\usepackage{svg}
\usepackage{hyperref}
\usepackage{xcolor}
\usepackage{lipsum}
\usepackage{longtable}

\hypersetup{
    colorlinks=true,
    linkcolor=blue,     
    urlcolor=cyan,       
    citecolor=blue 
}

\begin{document}

\title{A Survey on Unlearnable Data}

\author{Jiahao Li, Yiqiang Chen, \IEEEmembership{Fellow,~IEEE}, Yunbing Xing, Yang Gu, Xiangyuan Lan
\thanks{J Li (lijiahao@ict.ac.cn), Y Xing (xingyunbing@ict.ac.cn), Y Chen (yqchen@ict.ac.cn), and Y Gu (guyang@ict.ac.cn) are with Institute of Computing Technology, Chinese Academy of Sciences, 100190, Beijing, China. X Lan (lanxy@pcl.ac.cn) are with the Pengcheng Laboratory, 518055, Shenzhen, China.}
\thanks{Corresponding author: Yiqiang Chen (yqchen@ict.ac.cn).}}

\markboth{A Preprint}%
{Shell \MakeLowercase{\textit{et al.}}: A Sample Article Using IEEEtran.cls for IEEE Journals}


\maketitle

\begin{abstract}
Unlearnable data (ULD) has emerged as an innovative defense technique to prevent machine learning models from learning meaningful patterns from specific data, thus protecting data privacy and security. By introducing perturbations to the training data, ULD degrades model performance, making it difficult for unauthorized models to extract useful representations. Despite the growing significance of ULD, existing surveys predominantly focus on related fields, such as adversarial attacks and machine unlearning, with little attention given to ULD as an independent area of study. This survey fills that gap by offering a comprehensive review of ULD, examining unlearnable data generation methods, public benchmarks, evaluation metrics, theoretical foundations and practical applications. We compare and contrast different ULD approaches, analyzing their strengths, limitations, and trade-offs related to unlearnability, imperceptibility, efficiency and robustness. Moreover, we discuss key challenges, such as balancing perturbation imperceptibility with model degradation and the computational complexity of ULD generation. Finally, we highlight promising future research directions to advance the effectiveness and applicability of ULD, underscoring its potential to become a crucial tool in the evolving landscape of data protection in machine learning. Project page: \href{https://github.com/LiJiahao-Alex/Awesome-UnLearnable-Data}{\textcolor{magenta}{https://github.com/LiJiahao-Alex/Awesome-UnLearnable-Data}}.

\end{abstract}


\section{Introduction}
The rapid evolution of deep learning has been fueled by the unprecedented availability of large-scale datasets~\cite{deng2009imagenet,lin2014microsoft,karras2019style,Everingham10}, which in turn has driven remarkable performance improvements across diverse applications~\cite{achiam2023gpt,radford2021learning,ramesh2021zero}. However, as models become more data-dependent, concerns regarding data privacy~\cite{hill2022secretive}, intellectual property protection~\cite{somepalli2023diffusion}, and unauthorized data usage~\cite{birhane2021large} have grown significantly. In response to these issues, techniques aimed at making data unlearnable to machine learning models have emerged in recent years. \emph{Unlearnable Data} (ULD) refers to a category of data that has been deliberately modified through subtle perturbations, preventing models from effectively learning useful representations during training while maintaining perceptual quality for human observers. ULD technique serves as a proactive defense mechanism against unauthorized data collection, data theft, and dataset misuse.

\begin{figure}
    \centering
    \includegraphics[width=0.99\linewidth]{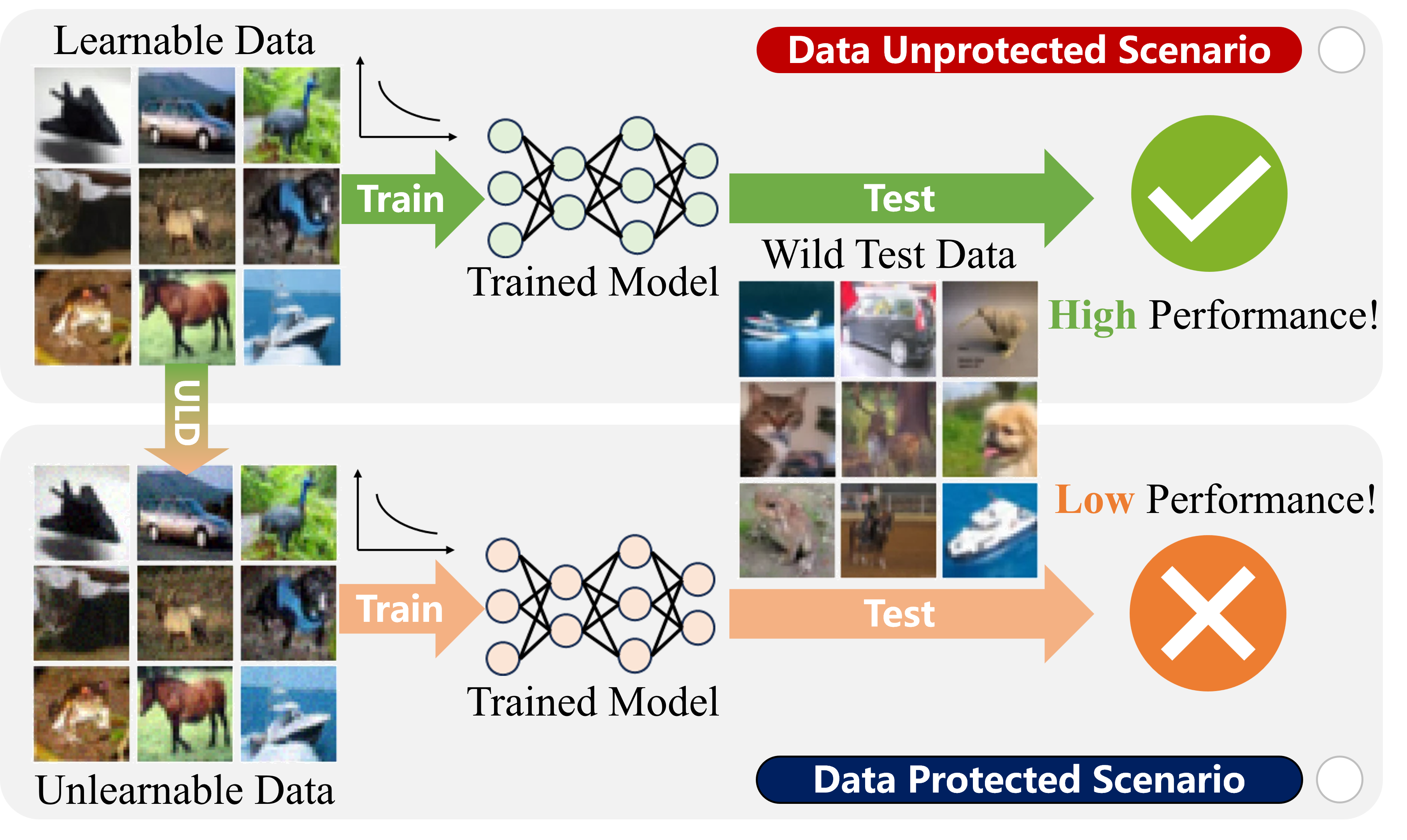}
    \caption{The Illustration of Unlearnable Data in Machine Learning.}
    \label{fig:ULD}
\end{figure}

It is worth noting that the concept of ULD is very similar to machine unlearning~\cite{bourtoule2021machine} and adversarial attacks~\cite{goodfellow2014explaining}, in that all three approaches manipulate data to influence model behavior, but they fundamentally differ in their objectives, timing, and mechanisms. Machine unlearning is primarily concerned with retroactively removing the influence of certain data points from a trained model, often to comply with privacy regulations~\cite{logemann2018art} or to correct for data errors~\cite{thudi2022necessity}. This process is typically performed after the model has been fully trained. In contrast, adversarial attacks focus on introducing carefully crafted noises to test inputs, aiming to mislead the well trained model during inference while leaving human perception basically unaltered. ULD, on the other hand, adopts a proactive strategy as shown in Figure~\ref{fig:ULD}. Rather than noising test inputs or stripping learned information post-training, ULD techniques modify the training data in such a way that the model is hindered from learning useful representations from it from the outset. This means that even when the data is available during training, its contribution to the model’s feature extraction process is deliberately minimized or nullified. In other words, unlike previous research that aim to influence trained models behavior, ULD focus on corrupting the training process itself, ensuring that models trained on such data exhibit degraded performance on generalization. Thus, while all these methods involve data manipulation, ULD is distinct in its preventive approach to data learning, setting it apart from both the post-hoc nature of machine unlearning and the inference-focused methodology of adversarial attacks.

Another closely related concept to ULD is the backdoor attack~\cite{gu2019badnets}, both of which manipulate the training data but with fundamentally different goals and mechanisms. Backdoor attacks aim to implant triggers into the model by injecting carefully crafted samples into the training data. These triggers can take various forms, ranging from imperceptible perturbations, such as subtle pixel modifications~\cite{doan2021backdoor} or watermarks~\cite{li2022untargeted}, to more conspicuous patterns, like distinct shapes~\cite{gu2019badnets} or colors~\cite{jiang2023color}, ensuring reliable activation. A key characteristic of backdoor attacks is that they are designed to preserve the model's performance on clean, unperturbed testing data, ensuring the model behaves as expected in the absence of the trigger. In contrast, ULD does not involve embedding hidden triggering behaviors but rather corrupts the entire learning process from the outset. The objective of ULD is to prevent the model from learning meaningful representations from the data, resulting in degraded performance across all inputs, whether perturbed or clean. While both methods involve manipulating training data, backdoor attacks introduce specific vulnerabilities that affect the model only when the trigger is present, whereas ULD systematically degrades the model's ability to learn effectively. Additionally, backdoor attacks are typically revocable; that is, the attack can be mitigated by removing the trigger. In contrast, ULD often represents an irrevocable disruption of the learning process, making it hard for the model to recover its ability to learn from the data.

Recent studies have explored various ULD approaches, such as adding error-minimizing noise~\cite{EM}, using convolution-based methods~\cite{CUDA}, or leveraging adversarial noise to optimize perturbations~\cite{REM}. These techniques have demonstrated effectiveness in preventing unauthorized model training while preserving the data’s usability for human observation. Despite the promise of ULD techniques for protecting data privacy and preventing unauthorized exploitation, several challenges persist. First, robust learning algorithms and adversarial training can potentially mitigate the effects of unlearnable perturbations, reducing their effectiveness. Second, there exists a critical trade-off between the imperceptibility of the perturbations and the degree of model degradation, as excessive modifications may introduce visible artifacts that limit practical deployment. Third, generating unlearnable data often incurs significant computational overhead, particularly for large-scale datasets. Many state-of-the-art approaches rely on iterative optimization methods, which can be computationally expensive and time-consuming. Finally, the ethical implications of unlearnable data raise concerns regarding its dual-use potential—while it can protect data privacy, it may also be exploited for anti-competitive practices or malicious intent. These challenges underscore the need for a comprehensive survey that not only reviews the current progress in ULD techniques but also provides a detailed analysis of their theoretical foundations, evaluation metrics, and practical applications.

Although research on improving model robustness and protecting data privacy is on the rise~\cite{liu2025threats,zhang2023review,qu2023learn,akhtar2018threat,chakraborty2021survey,akhtar2021advances,li2022backdoor,li2023backdoor,gao2020backdoor,zhang2020adversarial,nguyen2022survey,liu2024machine,wang2024machine}, systematic exploration of ULD is still missing. ULD is a new technique introduced in the recent years that prevents current machine learning model (e.g. deep neural network) from learning the useful features of specified data~\cite{EM}. Yet, many existing surveys mainly focus on related topics such as machine unlearning~\cite{liu2025threats,zhang2023review,qu2023learn}, adversarial attacks~\cite{akhtar2018threat,zhang2020adversarial,chakraborty2021survey,akhtar2021advances}, and backdoor attack~\cite{li2022backdoor,li2023backdoor,gao2020backdoor}, with ULD receiving minimal attention. Even when mentioned, it is often regarded as a special case~\cite{nguyen2022survey,liu2024machine,wang2024machine} rather than being the subject of dedicated investigation. This lack of dedicated attention hinders a comprehensive understanding of the field, making it difficult to discern the evolutionary trajectory, underlying mechanisms, and practical implications of ULD. Therefore, it is crucial to conduct an in-depth survey that consolidates recent advancements, highlights persistent challenges, and delineates future research directions to better inform and support the machine learning community. To bridge this gap, this survey provides a comprehensive review of the current landscape of ULD research.

The main contributions of this survey are as follows: 

\begin{itemize}
    \item \textbf{Comprehensive Review:} The survey provides a holistic and systematic review of unlearnable data (ULD) as an independent and evolving research area, consolidating scattered research efforts into a unified narrative. It covers the full spectrum of ULD  prior to the completion date of this survey—from generation methods and public benchmarks to evaluation metrics, theoretical foundations, and practical applications, etc.

    \item \textbf{Taxonomy Development:} By organizing ULD techniques along several dimensions (e.g., technical intent, data type, task scenario, surrogate model dependency, supervision dependency, perturbation boundedness, etc.), the survey offers a clear and multi-perspective framework that categorizes the diverse approaches in the field.

    \item \textbf{Critical Analysis:} The survey conducts an in-depth analysis of ULD techniques, identifying key strengths, limitations, and trade-offs while offering insights into their practical implications.

    \item \textbf{Challenges and Opportunities:} We highlights open challenges and existing limitations in emerging trends, such as transferability, imperceptibility, scalability, interpretability, revocability, stability, adaptability, ethicality, robustness, etc., which shed light on unresolved issues and offer future exploration directions for advancing ULD techniques toward greater generality, practicality, and usability.
    
\end{itemize}

\begin{figure*}[ht]
    \centering
    \includegraphics[width=0.98\linewidth]{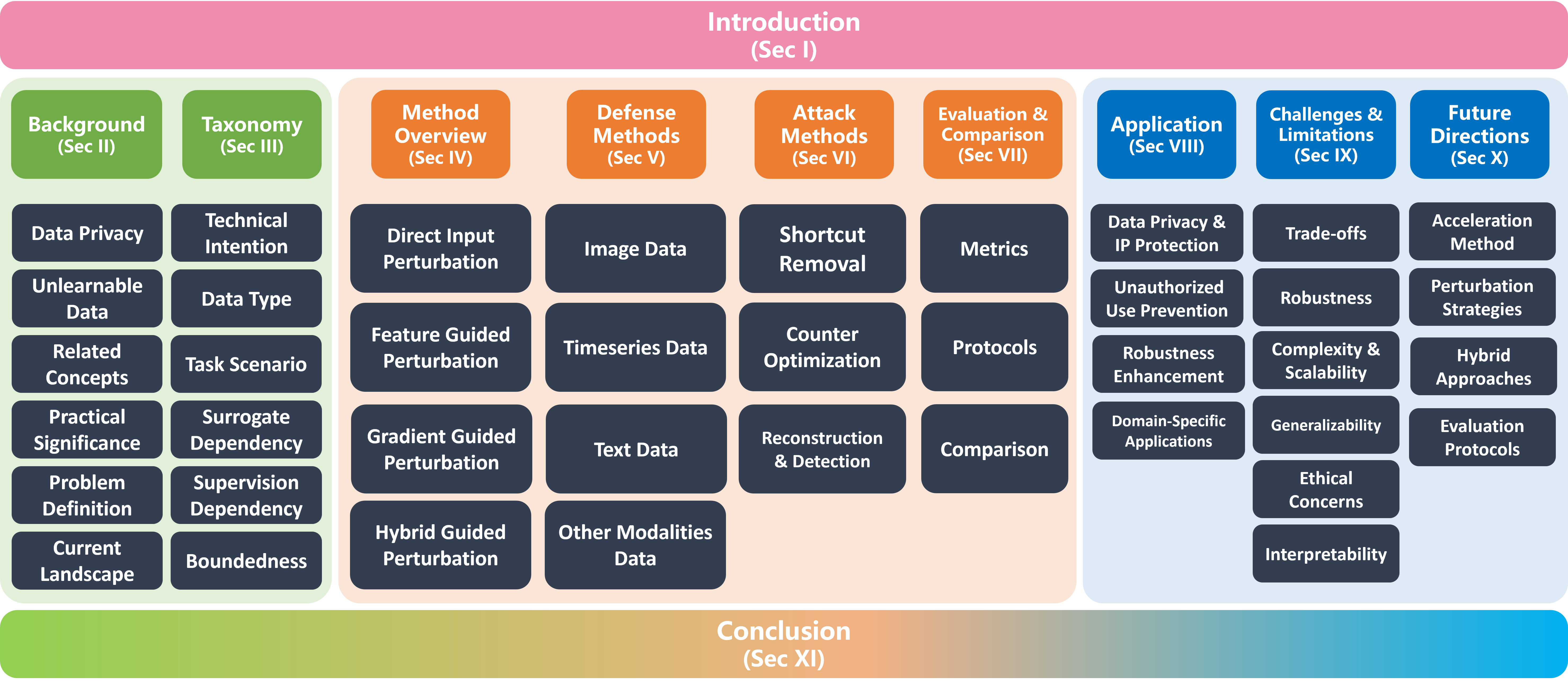}
    \caption{An overview of the structure of the survey.}
    \label{fig:SectionArrangment}
\end{figure*}

The subsequent survey structure is arranged as follows: Section~\ref{sec:Background} presents the Background, offering foundational concepts and contextualizing ULD within the broader landscape of machine learning security. Section~\ref{sec:Taxonomy} presents a comprehensive taxonomy of ULD techniques, categorizing them across multiple dimensions such as technical intent, data type, task scenario, surrogate model dependency, supervision dependency, and boundedness. Section~\ref{sec:Methodologies},~\ref{sec:GenerationStrategies}, and~\ref{sec:SpecificAttackMethods} delves into the methodologies for ULD, detailing key approaches and their underlying principles. Section~\ref{sec:Evaluation} explores the evaluation metrics used to measure unlearnability, imperceptibility, and robustness, alongside a comparative analysis of existing techniques. Section~\ref{sec:Applications} highlights practical applications of ULD, spanning areas like data privacy, intellectual property protection, and adversarial defense. Section~\ref{sec:Challenges} and~\ref{sec:Future} identifies critical challenges in ULD research and outlines promising future directions, such as enhancing scalability, interpretability, and robustness. Finally, Section~\ref{sec:Conclusion} concludes the survey, reflecting on the current state of ULD research and its future trajectory. The overview is shown in Figure~\ref{fig:SectionArrangment}.

\section{Background}
\label{sec:Background}
In recent years, large-scale datasets have become indispensable for training complex machine learning models, particularly deep neural networks. While this data-driven paradigm has fueled remarkable advancements, it has also raised concerns about data privacy, unauthorized access, and the potential misuse of sensitive information. These rising concerns have driven the development of methods to safeguard data from exploitation. One emerging trend is unlearnable data (ULD), which protects data by preventing unauthorized models from learning useful representations while ensuring the data remains accessible for legitimate purposes, such as publication, sharing, or human inspection. This section provides the essential background for understanding ULD within the broader context of machine learning. It first examines the intrinsic dependency of machine learning models on large-scale datasets and the implications of this reliance. Next, it introduces ULD as a response to growing privacy concerns and unauthorized usage. To contextualize ULD, related concepts are discussed, highlighting its distinctions from other data protection techniques. Then it further outlines the practical motivations behind ULD and its significance in real-world applications, followed by a formal definition of the problem it addresses. Finally, it reviews the evolution of ULD research, offering the current landscape in the field.

\subsection{Machine Learning and Data Privacy}
Machine learning (ML)~\cite{jordan2015machine} has revolutionized a wide range of fields, from computer vision to natural language processing, largely due to the availability of massive datasets. The performance of ML models, particularly deep neural networks, is highly dependent on the quality and quantity of training data. As models grow increasingly complex, they require correspondingly larger datasets to generalize well and avoid overfitting. In this data-driven paradigm, the dataset becomes a cornerstone of model success, often determining the upper bound of performance.

This dependency is further reinforced by the scaling laws~\cite{kaplan2020scaling} observed in large-scale models, which reveal a power-law relationship between model size, dataset size, and performance. As model parameters scale into the billions and beyond~\cite{brown2020language,rasley2020deepspeed,touvron2023llama,liu2024sora,guo2025deepseek}, merely increasing the model’s capacity is insufficient to sustain performance improvements — the availability of massive, high-quality datasets becomes equally critical. In fact, recent studies have highlighted the risk of data exhaustion~\cite{villalobos2022will,jones2024ai}, where publicly accessible datasets may no longer be sufficient to support the continued scaling of models, further sparking public concerns~\cite{hill2022secretive,birhane2021large,carlini2021extracting,shokri2017membership} about the unauthorized data exploration or misuse.

However, this growing reliance on data also introduces several challenges. In many cases, the datasets used to train models are collected from publicly available sources or through large-scale web scraping, raising concerns about data privacy~\cite{garhart2023wasn,martin2024artificial}, intellectual property rights~\cite{picht2023ai, oecd2025}, and unauthorized data usage~\cite{li2024digger,salem2018ml}. As machine learning systems become more widely deployed, ensuring that data owners maintain control over how their data is used has become a pressing issue. Unauthorized access to high-quality datasets can provide adversaries with a significant advantage, potentially leading to model theft, competitive exploitation, or privacy breaches.

In response to these challenges, protective mechanisms have emerged to safeguard datasets from misuse, either by limiting access to the data or by rendering the data unlearnable to unauthorized models. In this context, unlearnable data (ULD) has become a promising solution to proactively defend against unauthorized data exploitation. By injecting carefully crafted perturbations into the data, ULD aims to disrupt the training process, preventing models from learning meaningful representations while preserving perceptual quality for human observers. As machine learning continues to expand into sensitive areas such as healthcare, finance, and autonomous systems, the demand for robust data protection techniques like ULD is expected to grow, making data dependency a double-edged sword — both a source of power and a potential vulnerability.

\subsection{Emergence of Unlearnable Data}
The concept of Unlearnable Data (ULD) emerged as a proactive response to the increasing concerns surrounding data privacy, intellectual property protection, and unauthorized model training. Early deep learning models heavily relied on massive datasets to achieve remarkable performance, yet this reliance exposed sensitive data to exploitation, particularly when datasets were scraped from public sources or shared without stringent access control.

The first noTable attempt to introduce ULD was the Error-Minimizing Noise~\cite{EM} technique. Error-Minimizing marked the inception of ULD by injecting subtle perturbations into training data, preventing models from effectively learning useful representations while keeping data visually unchanged for human observers. This pioneering work framed ULD as a defensive measure aimed at protecting personal data, setting a precedent for the broader exploration of unlearnable strategies.

Following the introduction of Error-Minimizing, research into ULD techniques rapidly expanded. Early studies focused on enhancing perturbation effectiveness and robustness, particularly against adversarial training. Over time, the concept grew beyond simple error-minimizing noise, encompassing more sophisticated techniques such as robust unlearnable examples, cluster-based unlearnable methods, and convolution-based perturbations. These advancements aimed to protect data not only from traditional models but also from robust learning techniques and data augmentation.

In parallel, attacks targeting ULD also emerged. Researchers began exploring methods to bypass unlearnability by developing techniques that restored learnability to perturbed data. This cat-and-mouse dynamic between attack and defense has driven continuous innovation in ULD methodologies, giving rise to a diverse landscape of approaches across multiple data modalities, including images, text, audio, and point clouds.

The emergence of ULD has not only reshaped the discourse on data security but has also opened up new lines of inquiry into the very nature of learnability in machine learning. Today, ULD stands as a rapidly evolving field, balancing the need for robust data protection with the ongoing challenge of preserving imperceptibility and scalability.

\subsection{Related Concepts and Distinctions}

Unlearnable Data (ULD) is closely related to several existing concepts in machine learning security, such as adversarial attacks, data poisoning, machine unlearning, and backdoor attacks. While these techniques share the commonality of manipulating data to influence model behavior, their goals, mechanisms, and stages of intervention differ fundamentally. This section clarifies these distinctions to establish a clearer boundary between ULD and related concepts.

\subsubsection{Adversarial Attacks}
Adversarial attacks introduce carefully crafted perturbations to input samples with the goal of misleading a trained model during inference. These perturbations are typically imperceptible to humans but cause the model to produce incorrect predictions. In contrast, ULD intervenes before training, preventing models from learning meaningful representations in the first place. While adversarial attacks target the inference phase, ULD focuses on disrupting the training process itself.

\subsubsection{Data Poisoning}
Data poisoning manipulates training data to deliberately degrade model performance or implant hidden vulnerabilities. Poisoning attacks can take different forms, such as availability attacks, which aim to reduce overall performance, or targeted attacks, which induce misclassification for specific inputs. ULD is conceptually similar to availability poisoning in that both aim to degrade model performance. However, the primary intent behind ULD is data protection, not malicious sabotage, making ULD a more proactive and defensive strategy.

\subsubsection{Machine Unlearning}
Machine unlearning focuses on removing the influence of specific data points from a trained model, often to comply with privacy regulations like \textit{the right to be forgotten}. Unlike ULD, which prevents data from being learned in the first place, machine unlearning is a post-training process that retroactively erases data traces from an already trained model. In essence, ULD is a preventive measure, while machine unlearning serves as a corrective measure.

\subsubsection{Backdoor Attacks}
Backdoor attacks embed hidden triggers into training data, causing the model to behave normally on clean inputs while producing maliciously controlled outputs when the trigger is present. Unlike ULD, which aims to prevent overall learning, backdoor attacks are designed to control model behavior selectively. Additionally, ULD degrades performance across the entire dataset, whereas backdoor attacks maintain clean performance except in the presence of the trigger.

In summary, ULD stands out by taking a preventive stance—disrupting the learning process from the outset to protect data from unauthorized exploitation. This sets it apart from adversarial attacks, data poisoning, and backdoor attacks, which focus on manipulating model behavior either during training or inference. Similarly, ULD differs from machine unlearning by proactively rendering data unlearnable, rather than erasing knowledge after the fact. Understanding these distinctions helps contextualize ULD as a unique and evolving technique in the broader landscape of machine learning security.

\subsection{Motivation and Practical Significance}
The emergence of Unlearnable Data (ULD) is driven by a growing need to protect data in an era where machine learning models are becoming increasingly data-hungry. As models scale to billions of parameters and require massive datasets to train effectively, concerns over data privacy, intellectual property (IP) protection, and unauthorized data usage have become more pronounced. ULD offers a proactive solution to these challenges by preventing models from extracting meaningful representations from data without proper authorization.

One of the primary motivations behind ULD is personal data privacy. With the widespread adoption of data-driven technologies, personal data is often collected, shared, and used for model training without explicit consent. Techniques like ULD empower individuals and organizations to safeguard their data from being exploited by unauthorized parties, aligning with privacy-centric regulations such as the General Data Protection Regulation (GDPR)~\cite{regulation2018generalGDPR} and the California Consumer Privacy Act (CCPA)~\cite{bonta2022californiaCCPA}.

Another key driver is IP protection and data ownership. High-quality datasets are invaluable assets in fields like healthcare, finance, and autonomous systems, where proprietary data provides a competitive advantage. ULD ensures that even if datasets are leaked, scraped, or accessed without permission, unauthorized models trained on such data would exhibit degraded performance, effectively nullifying the value of stolen data.

Furthermore, ULD holds practical significance in defending against model stealing and unauthorized learning. In scenarios where public datasets are released for research purposes, ULD can act as a safeguard to prevent malicious actors from training high-performance models without proper attribution. This extends to protecting open-source datasets while still enabling their use for human-centric applications, maintaining accessibility while restricting machine learning exploitation.

Lastly, the rise of adversarial learning and competitive misuse has further highlighted the importance of ULD. As machine learning becomes deeply integrated into critical infrastructure, malicious entities could exploit public data to train models for harmful purposes. ULD offers a means of controlling access to learning capabilities, ensuring that data remains a controlled resource in high-stakes applications.

In summary, ULD addresses pressing concerns in data privacy, intellectual property protection, and unauthorized model training, offering a robust mechanism to prevent data exploitation while preserving its usability for human interpretation. As machine learning continues to permeate every facet of society, ULD presents itself as a timely and necessary safeguard in the broader landscape of data security.

\subsection{Formal Problem Definition}

\subsubsection{Preliminaries and Notation}  
Let $ \mathcal{X} $ denote the data space and $ \mathcal{Y} $ denote the label space, where the data distribution is represented by $ \mathcal{D} $. A dataset $ D \subseteq \mathcal{X} \times \mathcal{Y} $ consists of $ N $ samples:  

\begin{equation}
    D = \{(x_i, y_i)\}_{i=1}^N, \quad x_i \in \mathcal{X}, \ y_i \in \mathcal{Y}
    \label{eq:dataset}
\end{equation}

In the case of unsupervised tasks, $ \mathcal{Y} = \emptyset $ and $ D $ is composed of unlabeled samples:  

\begin{equation}
    D = \{x_i\}_{i=1}^N
    \label{eq:unsupervised_dataset}
\end{equation}

A machine learning model $ f_\theta: \mathcal{X} \rightarrow \mathcal{Y} $ is parameterized by $ \theta \in \Theta $, trained to minimize an empirical loss function $ \mathcal{L} $ over $ D $:  

\begin{equation}
    \theta^* = \arg \min_{\theta \in \Theta} \mathbb{E}_{(x,y) \sim D} \left[ \mathcal{L}(f_\theta(x), y) \right]
    \label{eq:empirical_loss}
\end{equation}

\subsubsection{Unlearnable Data Objective}  
The goal of Unlearnable Data (ULD) is to craft perturbations $ \delta: \mathcal{X} \rightarrow \mathcal{X} $ to create a perturbed dataset $ D^\prime $, where:  

\begin{equation}
    D^\prime = \{(x_i^\prime, y_i)\}_{i=1}^N, \quad x_i^\prime = x_i + \delta(x_i; y_i)
    \label{eq:supervised_uld}
\end{equation}

In an unsupervised setting, the perturbed dataset is defined as:  

\begin{equation}
    D^\prime = \{x_i^\prime\}_{i=1}^N, \quad x_i^\prime = x_i + \delta(x_i)
    \label{eq:unsupervised_uld}
\end{equation}

A model trained on $ D^\prime $ should fail to extract meaningful features, resulting in performance degradation across tasks such as classification, generation, segmentation, or retrieval. The optimization objective for generating ULD can thus be formulated as:  

\begin{equation}
    \delta^* = \arg \max_{\delta \in \Delta} \mathcal{J}(f_{\theta^\prime}, D_{test})
    \label{eq:uld_objective}
\end{equation}

Where  
$ \theta^\prime = \arg \min_{\theta \in \Theta} \mathbb{E}_{x^\prime \sim D^\prime} \left[ \mathcal{L}(f_\theta(x^\prime), y) \right] $ is the model trained on the unlearnable dataset $ D^\prime $.  
$ \mathcal{J} $ is a performance degradation metric, such as accuracy, loss, or task-specific evaluation measures.  
$ D_{test} $ is a clean test dataset, ensuring the model's degraded generalization ability.  
$ \Delta $ is the perturbation space subject to imperceptibility constraints:  

\begin{equation}
    \lVert \delta(x) \rVert_p \leq \epsilon, \quad \forall x \in D
    \label{eq:imperceptibility_constraint}
\end{equation}



\subsubsection{Properties and Constraints}  
The effectiveness of Unlearnable Data (ULD) hinges on two fundamental properties: Unlearnability and Imperceptibility. These properties serve as the cornerstone of ULD techniques, ensuring that unauthorized models fail to extract meaningful representations while preserving the perceptual quality of the data.  
\textbf{Unlearnability:} The primary objective of ULD is to prevent models from learning useful features from the training data, thereby degrading performance on downstream tasks. Formally, for a model $f_\theta$ trained on a perturbed dataset $D^\prime$, its performance on a clean test set $D_{test}$ should be significantly reduced compared to a model trained on the original dataset $D$.  
\textbf{Imperceptibility:} To ensure the perturbed data remains indistinguishable from the original data by human observers, the perturbations are typically constrained within an $L_p$-norm ball of radius $\epsilon$:

\begin{equation}
    \lVert \delta(x) \rVert_p \leq \epsilon, \quad \forall x \in D
    \label{eq:imperceptibility_constraint2}
\end{equation}

Beyond these fundamental properties, several other characteristics such as transferability, scalability, robustness, etc. have emerged in recent studies, shaping the evolution of ULD. These aspects reflect ongoing challenges and new research directions, which are further discussed in Section~\ref{sec:Future}.

\subsubsection{Generalized ULD Formulation}  
In summary, the formal problem of ULD involves optimizing $ \delta $ under the constraints of imperceptibility while ensuring the learned model $ f_{\theta^\prime} $ exhibits degraded performance across diverse tasks and modalities. The generalized objective can be expressed as:  

\begin{equation}
    \delta^* = \arg \min_{\delta \in \Delta} \mathbb{E}_{(x, y) \sim D} \left[ \mathcal{M}(f_{\theta^\prime}, D_{test}) \right] \quad \text{s.t.} \quad \lVert \delta(x) \rVert_p \leq \epsilon
    \label{eq:generalized_uld}
\end{equation}

Where $ \mathcal{M} $ represents the model’s ability to learn useful representations, measured by performance on task-specific evaluation metrics. This formulation serves as a universal framework to accommodate future advancements in ULD techniques across different domains.

\subsection{Evolution and Current Landscape}
The field of Unlearnable Data (ULD) has evolved significantly over the past few years, driven by the dual motivations of defending machine learning models from attacks and improving adversarial robustness. ULD techniques aim to prevent models from learning certain patterns, either by degrading model performance through unlearnability attacks or by introducing data that confounds learning algorithms. Over time, these techniques have evolved in sophistication, covering a wide range of applications, from defense mechanisms to attacks that exploit model vulnerabilities.

Early work in ULD primarily concentrated on defensive strategies, where the primary technical intention was to enhance model robustness and prevent adversarial exploitation. These early techniques aimed to lock certain learned features, preventing models from overfitting or learning spurious correlations. Methods such as EM~\cite{EM} and GrayAugs~\cite{GoingGrayscale} employed simple data transformations and augmentations to enhance model resilience. However, as the field matured, the focus shifted to more sophisticated techniques such as REM~\cite{REM} and LLock~\cite{peng2022learnabilityLOCK}, which refined defenses by introducing the use of surrogate models and stronger mechanisms to protect against evolving unlearnability attacks.

Simultaneously, unlearnability attacks began to gain prominence, with techniques like JCDP~\cite{jiang2023unlearnableGiveAFalse}, ISS~\cite{liu2023ImageShortcutSqueezing}, and UEraser~\cite{qin2023learningTheUnlearnable} focusing on creating unlearnable examples that degrade model performance by introducing confusion or ambiguity into the data. These attack-based strategies have highlighted the vulnerabilities of machine learning models, sparking a deeper understanding of the risks posed by adversarial settings.

In recent years, a more holistic approach has emerged in ULD research, where the interplay between defense and attack strategies is acknowledged. This dual approach is essential for developing methods that can protect models from adversarial threats while also exploring the possibilities of using unlearnable data to exploit vulnerabilities. Notable works such as AVATAR~\cite{dolatabadi2024devilAdvocate}, EUDP~\cite{zhao2023unlearnableDiffusionModels}, and ASR~\cite{fang2024rethinking} have advanced the field by developing techniques that can be used both for attacking and defending, often tailored to specific application domains such as image classification, text generation, or medical imaging.

As the research landscape broadens, ULD techniques now span a wide variety of data types, including images, audio, text, and time-series data (e.g., EEG). From simple transformations like those used in OPS~\cite{OPS} to more complex models incorporating deep learning and optimization techniques (e.g., ARMOR~\cite{gong2025armor}), ULD methods have diversified significantly. The techniques are applied to a range of domains, including segmentation in medical imaging (UMed~\cite{sun2024medicalUnlearnable}) and 3D object recognition using point cloud data (UPC~\cite{wang2024unlearnable3DpointClouds}), showing the growing domain-specific challenges that ULD aims to address.

The current landscape also reflects a broader understanding of various factors that influence the effectiveness of ULD. The boundedness of transformations is a key consideration, ensuring that unlearnable data does not result in unrealistic or computationally impractical perturbations. Moreover, the research on transferability highlights the importance of ensuring that unlearnable data can generalize across different models, tasks, and scenarios. Recent advancements have also emphasized the need for scalable methods that can handle larger datasets and more complex models efficiently.

Key to this evolution is the recognition of the importance of interpretability and stability in ULD techniques. As ULD becomes more widely applied in real-world settings, understanding how unlearnable data works, and ensuring its stability across various adversarial threats, becomes increasingly critical. Additionally, recent advancements in adaptability and robustness aim to ensure that ULD methods remain effective in the face of new, evolving adversarial techniques and model architectures.

Looking forward, the future of ULD research is centered on creating more robust, adaptable, and scalable methods that strike a balance between effective defenses and realistic attack scenarios. The integration of ULD into real-world applications such as privacy-preserving machine learning, secure AI systems, and enhancing adversarial robustness promises to drive further innovation. As the landscape continues to evolve, these efforts will contribute to building more secure and reliable AI systems capable of resisting both known and unknown adversarial threats.

\begin{figure*}
    \centering
    \begin{overpic}[width=0.99\linewidth]{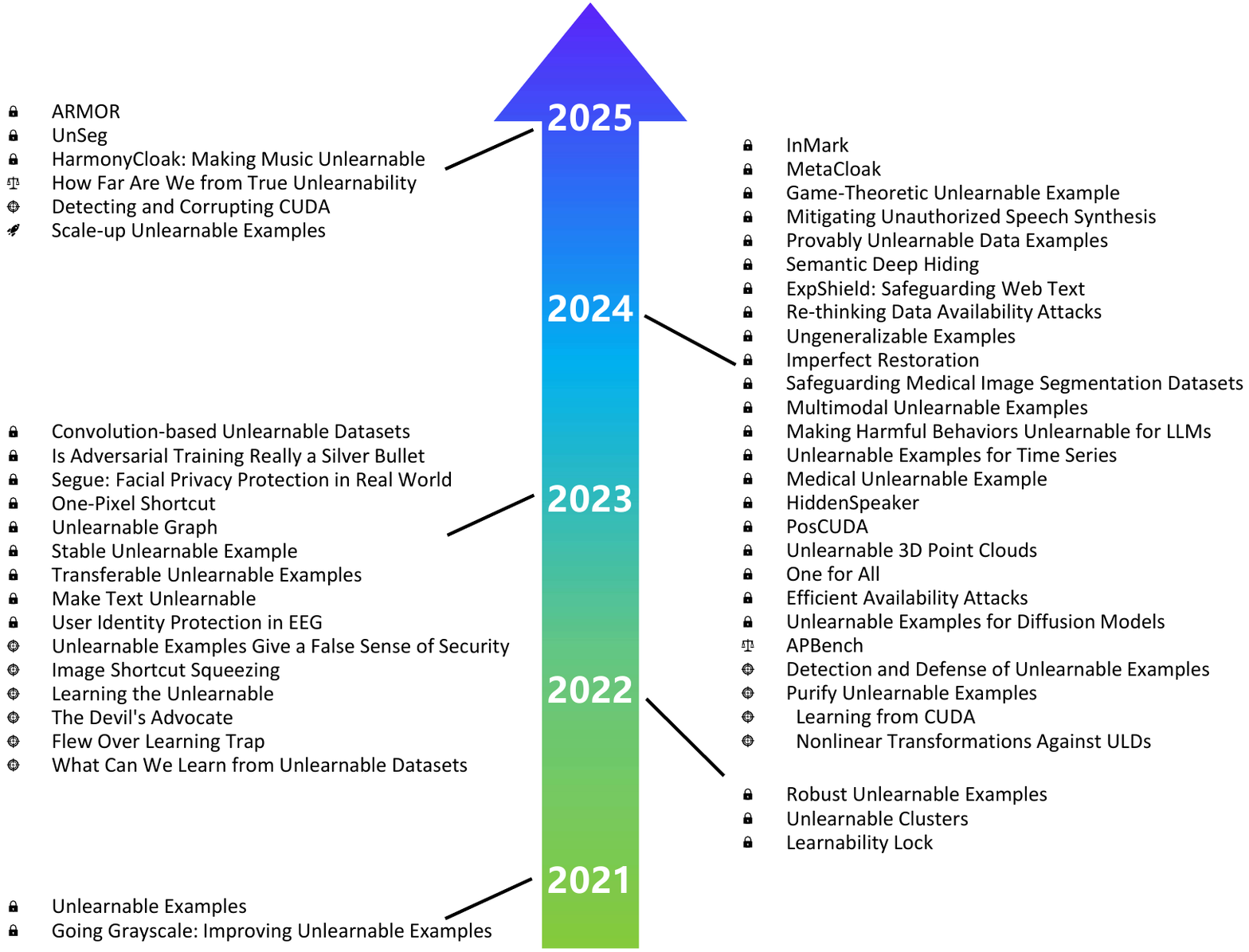}
        \put(4.5,62){\scriptsize{~\cite{gong2025armor}}}
        \put(4.5,60.2){\scriptsize{~\cite{UnSeg}}}
        \put(4.5,58.4){\scriptsize{~\cite{meerza2024harmonycloak}}}
        \put(4.5,56.6){\scriptsize{~\cite{zhu2025scaleupUnlearnableExamples}}}
        \put(4.5,54.8){\scriptsize{~\cite{HowFarAreWeFrom}}}
        \put(4.5,53.0){\scriptsize{~\cite{li2023detectingCorruptingCUDA}}}

        \put(4.5,38.2){\scriptsize{~\cite{CUDA}}}
        \put(4.5,36.43){\scriptsize{~\cite{wen2023is}}}
        \put(4.5,34.66){\scriptsize{~\cite{zhang2025segue}}}
        \put(4.5,32.89){\scriptsize{~\cite{OPS}}}
        \put(4.5,31.12){\scriptsize{~\cite{liu2023unlearnableGraph}}}
        \put(4.5,29.35){\scriptsize{~\cite{liu2024StableUnlearnable}}}
        \put(4.5,27.58){\scriptsize{~\cite{TransferableUEs}}}
        \put(4.5,25.81){\scriptsize{~\cite{li2023makeTextUnlearnable}}}
        \put(4.5,24.04){\scriptsize{~\cite{meng2023user}}}
        \put(4.5,22.27){\scriptsize{~\cite{jiang2023unlearnableGiveAFalse}}}
        \put(4.5,20.50){\scriptsize{~\cite{liu2023ImageShortcutSqueezing}}}
        \put(4.5,18.73){\scriptsize{~\cite{qin2023learningTheUnlearnable}}}
        \put(4.5,16.96){\scriptsize{~\cite{dolatabadi2024devilAdvocate}}}
        \put(4.5,15.19){\scriptsize{~\cite{dang2023flew}}}
        \put(4.5,13.42){\scriptsize{~\cite{sandoval2023WhatCanWeLearn}}}

        \put(4.5,3){\scriptsize{~\cite{EM}}}
        \put(4.5,1){\scriptsize{~\cite{GoingGrayscale}}}

        \put(59.1,59.6){\scriptsize{~\cite{liu2024counteringInMark}}}
        \put(59.1,57.83){\scriptsize{~\cite{liu2024metacloak}}}
        \put(59.1,56.06){\scriptsize{~\cite{liu2024gameUnlearnable}}}
        \put(59.1,54.29){\scriptsize{~\cite{MitigatingUnauthorizedSpeechSynthesis}}}
        \put(59.1,52.52){\scriptsize{~\cite{wang2024provably}}}
        \put(59.1,50.75){\scriptsize{~\cite{meng2024semanticHiding}}}
        \put(59.1,48.98){\scriptsize{~\cite{liu2024expshield}}}
        \put(59.1,47.21){\scriptsize{~\cite{fang2024rethinking}}}
        \put(59.1,45.44){\scriptsize{~\cite{UnlearnableClusters}}}
        \put(59.1,43.67){\scriptsize{~\cite{huang2024leveraging}}}
        \put(59.1,41.90){\scriptsize{~\cite{lin2024safeguarding}}}
        \put(59.1,40.13){\scriptsize{~\cite{liu2024multimodalUnlearnableMM}}}  
        \put(59.1,38.36){\scriptsize{~\cite{zhou2024makingHarmfulBehaviorsUnlearnable}}}
        \put(59.1,36.59){\scriptsize{~\cite{jiang2024UnlearnableTimeSeries}}}
        \put(59.1,34.82){\scriptsize{~\cite{sun2024medicalUnlearnable}}}
        \put(59.1,33.05){\scriptsize{~\cite{zhang2024hiddenspeaker}}}
        \put(59.1,31.28){\scriptsize{~\cite{gokul2024poscuda}}}
        \put(59.1,29.51){\scriptsize{~\cite{wang2024unlearnable3DpointClouds}}}
        \put(59.1,27.74){\scriptsize{~\cite{chen2024oneFroAll}}}
        \put(59.1,25.97){\scriptsize{~\cite{wang2024efficientAvailabilityAttacks}}}
        \put(59.1,24.20){\scriptsize{~\cite{zhao2023unlearnableDiffusionModels}}}
        \put(59.1,22.43){\scriptsize{~\cite{qin2024apbench}}}  
        \put(59.1,20.66){\scriptsize{~\cite{DetectionAndDefense}}}
        \put(59.1,18.89){\scriptsize{~\cite{yu2024purify}}}
        \put(59.1,17.12){\scriptsize{~\cite{LearningFromCUDA}}}
        \put(59.1,15.35){\scriptsize{~\cite{hapuarachchi2024nonlinearTransformationsAgainst}}}

        \put(59.1,11.2){\scriptsize{~\cite{REM}}}
        \put(59.1,9.43){\scriptsize{~\cite{UnlearnableClusters}}}
        \put(59.1,7.66){\scriptsize{~\cite{peng2022learnabilityLOCK}}}

    \end{overpic}
    \caption{The timeline of unlearnable data (ULD) research and related studies. The lock symbol ``\raisebox{-0.3ex}{\includegraphics[height=0.86em]{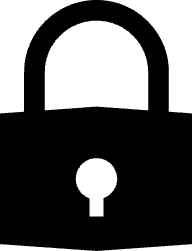}}'' represents the defense method, the cross-star symbol ``\raisebox{-0.3ex}{\includegraphics[height=0.86em]{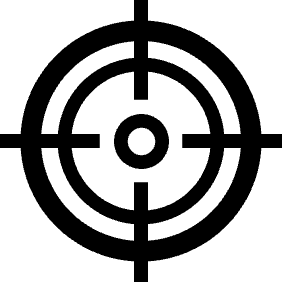}}'' represents the attack method, the balance symbol ``\raisebox{-0.3ex}{\includegraphics[height=0.86em]{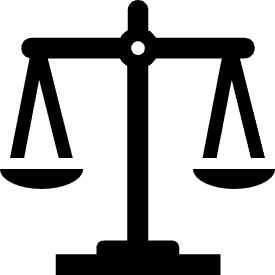}}'' represents the evaluation method, and the rocket symbol ``\raisebox{-0.3ex}{\includegraphics[height=0.86em]{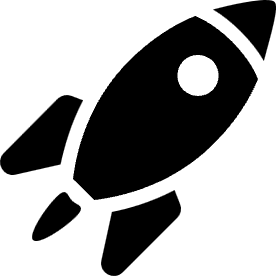}}'' represents the performance acceleration method.}
    \label{fig:Timeline}
\end{figure*}

To provide a comprehensive overview of the ULD techniques and their development timeline, we refer the reader to Table~\ref{tab:ALL} and the corresponding technology timeline presented in Figure~\ref{fig:Timeline}. These resources summarize the key advancements in ULD research and offer a clear visualization of how these techniques have evolved over time.

\section{Taxonomy of Unlearnable Data Techniques}
\label{sec:Taxonomy}
Unlearnable Data (ULD) techniques have rapidly evolved, giving rise to a diverse range of methods aimed at preventing machine learning models from learning useful features during training. Notably, ULD methods classification is inherently multi-faceted, as different studies categorize these methods based on distinct focal points. Depending on different concerns, ULD methods can be classified according to data type (e.g., images, text, audio), task applicability (e.g., classification, generation, segmentation), technical intent (e.g., defense, attack, acceleration), surrogate model dependency (e.g., surrogate-based vs. surrogate-free scenarios), robustness against adversarial countermeasures, etc. This section presents a comprehensive taxonomy that incorporates these diverse perspectives, providing a structured analysis of ULD techniques. Each classification criterion sheds light on different aspects of the technology, offering deeper insights into the evolution and application of ULD methods.

\begin{table*}[ht]
\caption{Overview of ULD techniques.}
\label{tab:ALL}
\resizebox{\linewidth}{!}{
\begin{tabular}{@{}lccccccccc@{}}
\toprule
\textbf{Study}                                                                                                                                                                                      & \textbf{Publication} & \textbf{Year} & \textbf{Data} & \textbf{Task}                & \textbf{Intend} & \textbf{Label} & \textbf{Bounded} & \textbf{Surrogate} & \textbf{Robust} \\ \midrule
EM~\cite{EM}   [\href{https://arxiv.org/pdf/2101.04898}{\textcolor{blue}{Paper}},   \href{https://github.com/HanxunH/Unlearnable-Examples}{\textcolor{magenta}{Code}}]                                        & ICLR                 & 2021          & Image         & Classification               & Defense       & YES              & YES                & YES                  & NO               \\
GrayAugs~\cite{GoingGrayscale}   [\href{https://arxiv.org/pdf/2111.13244}{\textcolor{blue}{Paper}},   \href{https://github.com/liuzrcc/ULE-GrayAug}{\textcolor{magenta}{Code}}]                                           & arXiv                & 2021          & Image         & Classification               & Defense       & YES              & NO                & YES                  & YES               \\
REM~\cite{REM}   [\href{https://arxiv.org/pdf/2203.14533}{\textcolor{blue}{Paper}},   \href{https://github.com/fshp971/robust-unlearnable-examples}{\textcolor{magenta}{Code}}]                                & ICLR                 & 2022          & Image         & Classification               & Defense       & YES              & YES                & YES                  & YES               \\
\bottomrule
\end{tabular}}
\end{table*}

\subsection{Categorization Based on Technical Intention}
In the current landscape of ULD research, there is a conventional consensus that ULD techniques are primarily categorized as defensive methods aimed at preventing unauthorized models from learning meaningful representations from data. This consensus was largely established by the seminal work~\cite{EM} in ULD, which first framed the concept of unlearnable data from the perspective of personal information protection. This work was explicitly designed to defend against unauthorized model training, thereby setting the foundation for viewing ULD as a defensive measure. In contrast, attempts to recover learnability from unlearnable data are often classified as attacks against these defensive measures. This classification aligns with the prevailing view in the community, which this survey adopts for clarity. However, it is worth noting that ULD techniques can also be perceived as attacks on model learnability depending on the deployed scenario and the intentions of the practitioner. This dual perspective reflects the ethical complexity surrounding ULD applications, which will be discussed further in Section~\ref{sec:Challenges}. In this section, we focus on presenting ULD techniques from both the attack and defense perspectives following the conventional consensus.

As summarized in Table~\ref{tab:IntentionType}, ULD techniques can be categorized into four primary technical intentions: defense, attack, evaluate, and computation acceleration. Defense-oriented techniques aim to render data unlearnable to unauthorized models, thereby preventing the extraction of meaningful patterns. Methods like GrayAugs~\cite{GoingGrayscale}, REM~\cite{REM}, and OPS~\cite{OPS} exemplify this category, offering robust data transformations to hinder model learning while preserving data utility for legitimate use cases. These techniques have been applied across diverse data types, including images, text, audio, and multimodal datasets, reflecting the broad applicability of defensive ULD methods.

Conversely, attack-oriented techniques seek to counteract these defensive measures by recovering learnability from unlearnable data or bypassing protective mechanisms. For instance, methods such as ISS~\cite{jiang2023unlearnableGiveAFalse} and Image Shortcut Squeezing~\cite{liu2023ImageShortcutSqueezing} aim to exploit model vulnerabilities, effectively neutralizing the protective effects of ULD. These attack strategies not only challenge the robustness of existing defenses but also provide insights into the development of more resilient protective mechanisms.

Additionally, some techniques focus on computation acceleration, streamlining the process of generating unlearnable data or enhancing scalability. HPC4UE~\cite{zhu2025scaleupUnlearnableExamples} is a notable example, presenting methods to expedite the creation of unlearnable datasets, thereby improving the practical deployment of ULD in large-scale scenarios.

Finally, evaluation-oriented techniques aim to assess the effectiveness and robustness of ULD strategies under various settings. These approaches provide quantitative benchmarks for measuring the degradation of model performance on unlearnable datasets, ensuring a standardized framework for comparison across different methods. Studies such as SALUD~\cite{HowFarAreWeFrom} and APBench~\cite{qin2024apbench} propose evaluation metrics or experimental protocols to systematically analyze the impact of ULD techniques in diverse machine learning scenarios.

Overall, thes four categories provide a structured understanding of ULD methodologies and their implications in different contexts. While the defense-oriented perspective dominates the field, the existence of attack and evaluation strategies highlights the ongoing arms race between protection mechanisms and adversarial countermeasures. Furthermore, the development of computation acceleration techniques signifies the growing need for scalable and efficient ULD generation methods as machine learning applications expand. The technical landscape presented in Table~\ref{tab:IntentionType} provides a comprehensive overview of these techniques, while the broader developmental trends are captured in the technology timeline presented in Figure~\ref{fig:Timeline}. Together, these resources offer valuable insights into the ongoing advancements in ULD research and its multifaceted applications.

In the subsequent sections, we delve deeper into the technical details of each category, analyzing the core principles and mathematical foundations that underpin modern ULD strategies. We begin with defense-oriented methods, which form the backbone of ULD research and continue to drive advancements in protecting data from unauthorized learning.

\begin{table}[ht]
\caption{Overview of ULD techniques categorized by technical intention.}
\label{tab:IntentionType}
\centering
\begin{tabular}{@{}ccc@{}}
\toprule
\textbf{Domain}                 & \textbf{Intention} & \textbf{Reference} \\ \midrule
\multirow{2}{*}{Method} & Defense            & \makecell{
~\cite{EM},~\cite{GoingGrayscale},~\cite{REM},~\cite{UnlearnableClusters},~\cite{peng2022learnabilityLOCK},\\
~\cite{TransferableUEs},~\cite{OPS},~\cite{CUDA},~\cite{liu2024StableUnlearnable},~\cite{zhang2025segue},\\
~\cite{zhao2023unlearnableDiffusionModels},~\cite{li2023makeTextUnlearnable},~\cite{liu2023unlearnableGraph},~\cite{meng2023user},~\cite{wen2023is},\\
~\cite{fang2024rethinking},~\cite{wang2024provably},~\cite{zhou2024makingHarmfulBehaviorsUnlearnable},~\cite{jiang2024UnlearnableTimeSeries},~\cite{sun2024medicalUnlearnable},\\
~\cite{liu2024multimodalUnlearnableMM},~\cite{meng2024semanticHiding},~\cite{zhang2024hiddenspeaker},~\cite{gokul2024poscuda},~\cite{liu2024gameUnlearnable},\\
~\cite{lin2024safeguarding},~\cite{wang2024unlearnable3DpointClouds},~\cite{chen2024oneFroAll},~\cite{liu2024metacloak},~\cite{wang2024efficientAvailabilityAttacks},\\ ~\cite{ye2024ungeneralizable},~\cite{liu2024counteringInMark},~\cite{MitigatingUnauthorizedSpeechSynthesis},~\cite{huang2024leveraging},\\
~\cite{gong2025armor},~\cite{UnSeg},~\cite{meerza2024harmonycloak}
}                   \\ \cmidrule(l){2-3}
                          & Attack             & \makecell{~\cite{jiang2023unlearnableGiveAFalse},~\cite{liu2023ImageShortcutSqueezing},~\cite{qin2023learningTheUnlearnable},~\cite{dolatabadi2024devilAdvocate},~\cite{dang2023flew},~\cite{hapuarachchi2024nonlinearTransformationsAgainst},\\ ~\cite{sandoval2023WhatCanWeLearn},~\cite{yu2024purify},~\cite{LearningFromCUDA},~\cite{DetectionAndDefense},~\cite{li2023detectingCorruptingCUDA}}                   \\ \midrule
Computation               & Accelerate         & ~\cite{zhu2025scaleupUnlearnableExamples}                   \\ \midrule
Metric                  & Evaluate           & ~\cite{qin2024apbench},~\cite{HowFarAreWeFrom}                   \\ \bottomrule

\end{tabular}
\end{table}

\subsection{Categorization Based on Data Type}
In the early stages (2021-2022) of ULD research, the focus was predominantly on image data, making ULD methods almost synonymous with unlearnable image techniques. It wasn’t until 2023 that researchers began exploring unlearnability in other data types, such as text, EEG signals, point clouds, etc. As of the completion of this survey, image-based ULD techniques remain the dominant research focus. However, the emergence of ULD methods targeting diverse data types signals a non-negligible trend that broadens the scope of unlearnable data. In this section, we categorize ULD methods according to the data types they aim to protect, shedding light on how perturbation strategies adapt across different modalities.

As shown in Table~\ref{tab:DataType}, the vast majority of ULD techniques focus on unimodal data, particularly image data, which has seen the earliest and most extensive exploration in this domain. Methods such as EM ~\cite{EM}, REM ~\cite{REM}, and OPS ~\cite{OPS} represent the cornerstone of unlearnable image techniques, targeting generic image datasets. Medical imaging has also garnered attention in recent years, with works like ~\cite{sun2024medicalUnlearnable,lin2024safeguarding} proposing domain-specific perturbation methods.

Beyond images, researchers have extended ULD techniques to other unimodal data types. Time series data, for instance, has seen techniques addressing generic time series ~\cite{jiang2024UnlearnableTimeSeries}, EEG signals ~\cite{meng2023user}, and even music ~\cite{meerza2024harmonycloak}. Audio data has similarly been explored, with efforts to make speech recognition models unlearnable ~\cite{zhang2024hiddenspeaker, gokul2024poscuda, MitigatingUnauthorizedSpeechSynthesis}. Graph-based ULD techniques have emerged as well, targeting graph-structured data ~\cite{liu2023unlearnableGraph}.

Text-based ULD has gradually gained traction, especially with the rise of large language models (LLMs). Methods like ~\cite{li2023makeTextUnlearnable} and ~\cite{zhou2024makingHarmfulBehaviorsUnlearnable} aim to disrupt unauthorized training on textual data. Furthermore, point clouds, which are crucial for 3D object recognition tasks, have become a new frontier for ULD research ~\cite{wang2024unlearnable3DpointClouds}.

In addition to unimodal data, multimodal approaches have begun to emerge. For instance, ~\cite{liu2024multimodalUnlearnableMM} introduces unlearnability across both image and text modalities, marking a step toward more comprehensive ULD strategies that span multiple data types.

The technical timeline presented in Figure X further illustrates the chronological development of these methods, reflecting the field's gradual expansion from images to diverse data types. This evolution not only broadens the applicability of ULD but also challenges researchers to design perturbations that effectively hinder learning across varying data modalities. Table~\ref{tab:DataType} provides a detailed overview of ULD techniques categorized by data type, offering insights into the current state of research and highlighting emerging trends across different data modalities.

\begin{table}[ht]
\caption{Overview of ULD techniques categorized by data type.}
\label{tab:DataType}
\resizebox{\linewidth}{!}{
\begin{tabular}{@{}cccc@{}}
\toprule
\multirow{2}{*}{\textbf{Data   Modality}} & \multicolumn{2}{c}{\textbf{Data Type}}              & \multirow{2}{*}{\textbf{Reference}} \\
                                          & \textbf{Primitive}           & \textbf{Subdivision} &                                     \\ \midrule
\multirow{9}{*}{Unimodal}                 & \multirow{2}{*}{Image}       & Generic              & \makecell{ ~\cite{EM},~\cite{GoingGrayscale},~\cite{REM},~\cite{UnlearnableClusters},~\cite{liu2024multimodalUnlearnableMM},\\ ~\cite{TransferableUEs},~\cite{OPS},~\cite{CUDA},~\cite{liu2024StableUnlearnable},~\cite{zhao2023unlearnableDiffusionModels},\\ ~\cite{jiang2023unlearnableGiveAFalse},~\cite{liu2023ImageShortcutSqueezing},~\cite{dolatabadi2024devilAdvocate},~\cite{li2023detectingCorruptingCUDA},~\cite{HowFarAreWeFrom}\\ ~\cite{sandoval2023WhatCanWeLearn},~\cite{wen2023is},~\cite{wang2024provably},~\cite{fang2024rethinking},~\cite{meng2024semanticHiding},\\ ~\cite{liu2024gameUnlearnable},~\cite{yu2024purify},~\cite{LearningFromCUDA},~\cite{DetectionAndDefense},~\cite{hapuarachchi2024nonlinearTransformationsAgainst},\\ ~\cite{ye2024ungeneralizable},~\cite{chen2024oneFroAll},~\cite{liu2024metacloak},~\cite{wang2024efficientAvailabilityAttacks},~\cite{qin2024apbench},\\ ~\cite{liu2024counteringInMark},~\cite{huang2024leveraging},~\cite{gong2025armor},~\cite{zhu2025scaleupUnlearnableExamples}, ~\cite{UnSeg},\\ ~\cite{dang2023flew},~\cite{qin2023learningTheUnlearnable}}                                    \\ \cmidrule(l){3-4} 
                                          &                              & Facial               & ~\cite{zhang2025segue}                                    \\ \cmidrule(l){3-4} 
                                          &                              & Medical              & ~\cite{sun2024medicalUnlearnable,lin2024safeguarding}                                    \\ \cmidrule(l){2-4} 
                                          & \multirow{4}{*}{Timeserires} & Generic              & ~\cite{jiang2024UnlearnableTimeSeries}                                    \\
                                          &                              & EEG                  & ~\cite{meng2023user}                                    \\
                                          &                              & Music                & ~\cite{meerza2024harmonycloak}                                    \\
                                          &                              & Audio                & ~\cite{zhang2024hiddenspeaker},~\cite{gokul2024poscuda},~\cite{MitigatingUnauthorizedSpeechSynthesis}                                    \\ \cmidrule(l){2-4} 
                                          & Graph                        & -                    & ~\cite{liu2023unlearnableGraph}                                    \\ \cmidrule(l){2-4} 
                                          & Text                         & -                    & ~\cite{li2023makeTextUnlearnable},~\cite{zhou2024makingHarmfulBehaviorsUnlearnable},~\cite{liu2024expshield}                                    \\ \cmidrule(l){2-4} 
                                          & Point Clouds                 & -                    & ~\cite{wang2024unlearnable3DpointClouds}                                    \\ \midrule
Multimodal                                & Image, Text                  & -                    & ~\cite{liu2024multimodalUnlearnableMM}                                    \\ \bottomrule
\end{tabular}}
\end{table}

\subsection{Categorization Based on Task Scenario}
In addition to data type and technical intention, the task scenario is another crucial dimension that shapes the design and evaluation of ULD techniques. Early research in ULD primarily targeted classification tasks—with works such as ~\cite{EM} and ~\cite{GoingGrayscale} demonstrating how carefully designed perturbations can suppress model accuracy by preventing classifiers from learning discriminative features. These foundational studies set the stage for understanding how unlearnable perturbations obstruct supervised learning.

As the field progressed, researchers began to explore additional task scenarios, broadening the impact of ULD beyond mere classification. For instance, in generation tasks, methods like those in ~\cite{zhang2025segue} and ~\cite{zhao2023unlearnableDiffusionModels} are designed to hinder models from accurately modeling data distributions, thereby impeding generative processes. Similarly, segmentation tasks have prompted specialized perturbation strategies; approaches reported in ~\cite{lin2024safeguarding} and ~\cite{UnSeg} not only obscure class boundaries but also maintain spatial coherence to effectively disrupt segmentation performance.

Table~\ref{tab:TaskType} provides an overview of ULD techniques categorized by task scenario, illustrating the evolution from single-task methods to more complex multi-task settings. For single-task applications, the Table lists a comprehensive collection of methods for classification (e.g., ~\cite{REM}, ~\cite{UnlearnableClusters}, ~\cite{peng2022learnabilityLOCK}), as well as for customized, generation, retrieval, segmentation, and verification tasks. Each category reflects distinct challenges; for example, retrieval methods, such as those in ~\cite{liu2024multimodalUnlearnableMM}, focus on disrupting similarity measures and feature matching, while verification approaches (e.g., ~\cite{zhang2024hiddenspeaker}) are designed to obstruct models from reliably validating data authenticity.

Moreover, the emergence of multi-task scenarios, where methods are designed to simultaneously impact tasks such as classification combined with Q\&A or segmentation (see ~\cite{li2023makeTextUnlearnable} and ~\cite{wang2024unlearnable3DpointClouds} in Table~\ref{tab:TaskType}), underscores the increasing complexity of modern ULD research. In these settings, perturbations must be carefully balanced to degrade performance across multiple objectives without sacrificing the effectiveness of any single task.

Collectively, the diversity of task scenarios highlighted in Table~\ref{tab:TaskType} demonstrates that ULD techniques are evolving beyond their initial focus on classification. This evolution reflects a broader trend toward developing comprehensive protection mechanisms that address various learning objectives. As ULD research continues to mature, it is anticipated that methods will further expand into new task domains, potentially heralding an “iPhone moment” where unlearnable data becomes a widely adopted tool across diverse applications.

\begin{table}[ht]
\caption{Overview of ULD techniques categorized by task scenario.}
\label{tab:TaskType}
\resizebox{\linewidth}{!}{
\begin{tabular}{@{}ccc@{}}
\toprule
\textbf{Scenario}            & \textbf{Task}                & \textbf{Reference} \\ \midrule
\multirow{6}{*}{Single task} & Classification               & \makecell{ ~\cite{EM},~\cite{GoingGrayscale},~\cite{REM},~\cite{UnlearnableClusters},~\cite{peng2022learnabilityLOCK},\\ ~\cite{TransferableUEs},~\cite{OPS},~\cite{CUDA},~\cite{liu2024StableUnlearnable},~\cite{jiang2023unlearnableGiveAFalse},\\ ~\cite{liu2023ImageShortcutSqueezing},~\cite{qin2023learningTheUnlearnable},~\cite{dolatabadi2024devilAdvocate},~\cite{dang2023flew},~\cite{sandoval2023WhatCanWeLearn},\\ ~\cite{wen2023is},~\cite{fang2024rethinking},~\cite{wang2024provably},~\cite{meng2024semanticHiding},~\cite{liu2024gameUnlearnable},\\ ~\cite{yu2024purify},~\cite{LearningFromCUDA},~\cite{DetectionAndDefense},~\cite{hapuarachchi2024nonlinearTransformationsAgainst},~\cite{ye2024ungeneralizable},\\ ~\cite{chen2024oneFroAll},~\cite{wang2024efficientAvailabilityAttacks},~\cite{qin2024apbench},~\cite{huang2024leveraging},~\cite{gong2025armor},\\ ~\cite{zhu2025scaleupUnlearnableExamples},~\cite{li2023detectingCorruptingCUDA},~\cite{HowFarAreWeFrom},~\cite{liu2023unlearnableGraph},~\cite{meng2023user},\\ ~\cite{jiang2024UnlearnableTimeSeries},~\cite{sun2024medicalUnlearnable},~\cite{gokul2024poscuda}}                   \\ \cmidrule(l){2-3}
                             & Customized                   & ~\cite{liu2024expshield}                   \\ \cmidrule(l){2-3}
                             & Generation                   & \makecell{~\cite{zhang2025segue},~\cite{zhao2023unlearnableDiffusionModels},~\cite{zhou2024makingHarmfulBehaviorsUnlearnable},~\cite{liu2024metacloak},~\cite{liu2024counteringInMark},\\ ~\cite{MitigatingUnauthorizedSpeechSynthesis},~\cite{meerza2024harmonycloak}}                   \\ \cmidrule(l){2-3}
                             & Retrieval                    & ~\cite{liu2024multimodalUnlearnableMM}                   \\ \cmidrule(l){2-3}
                             & Segmentation                 & ~\cite{lin2024safeguarding},~\cite{UnSeg}                   \\ \cmidrule(l){2-3}
                             & Verification                 & ~\cite{zhang2024hiddenspeaker}                   \\ \midrule
\multirow{2}{*}{Multi-task}  & Classification, Q\&A         & ~\cite{li2023makeTextUnlearnable}                   \\ \cmidrule(l){2-3}
                             & Classification, Segmentation & ~\cite{wang2024unlearnable3DpointClouds}                   \\ \bottomrule
\end{tabular}}
\end{table}

\subsection{Categorization Based on Surrogate Model Dependency}
Another crucial perspective in categorizing ULD techniques lies in their dependency on surrogate models during the ULD generation process. Surrogate models serve as approximations of the target model, providing gradients or training signals that guide the creation of unlearnable perturbations. Based on this dependency, ULD methods can be broadly classified into surrogate-based and surrogate-free approaches.

Early ULD research predominantly relied on surrogate models. For example, seminal works such as ~\cite{EM} and ~\cite{REM} utilized surrogate models to simulate the target model's behavior, thereby enabling the computation of effective gradients for crafting perturbations. These methods leveraged detailed knowledge about model architecture, training dynamics, and data distribution to design perturbations that significantly degrade the performance of the eventual target models. Such surrogate-based methods often achieve high effectiveness in controlled environments, as they can fine-tune the perturbation process by directly optimizing against a representative model. As indicated in Table~\ref{tab:ALL}, many early ULD studies explicitly marked the use of surrogate models (e.g., ~\cite{EM}, ~\cite{REM}, ~\cite{peng2022learnabilityLOCK}) to achieve precise perturbation generation.

However, this dependency also introduces certain limitations. The effectiveness of surrogate-based approaches may degrade when the surrogate model deviates from the actual target model, potentially reducing transferability and robustness. Moreover, in real-world applications, access to the internal details of the target model is often limited or entirely unavailable. These challenges have motivated recent research to explore surrogate-free strategies.

Surrogate-free methods aim to generate unlearnable data without relying on any explicit approximation of the target model. Instead, they often utilize alternative optimization objectives or heuristic strategies that do not require access to gradients from a surrogate model. This approach enhances the generalizability of ULD techniques, as it is less sensitive to the mismatch between surrogate and target models. Although surrogate-free methods may sometimes yield less potent perturbations compared to their surrogate-based counterparts, they offer significant advantages in terms of applicability in black-box scenarios, where model internals are inaccessible. Table~\ref{tab:ALL} further illustrates this trend, with several recent studies explicitly not depending on surrogate models (e.g., ~\cite{OPS}, ~\cite{CUDA}), highlighting their broader applicability.

Overall, the choice between surrogate-based and surrogate-free ULD techniques represents a trade-off between precision and applicability. Surrogate-based methods—with their fine-grained control and tailored perturbation design—excel in environments where target models are well understood. In contrast, surrogate-free approaches promise broader utility across diverse and uncertain settings, a trend that is likely to gain momentum as ULD research moves toward more practical, real-world applications.

This categorization not only underscores the evolution of ULD methodologies—from tightly controlled, model-dependent perturbations to more flexible, broadly applicable techniques—but also highlights the ongoing challenges in balancing effectiveness with generalizability. As the field advances, future research may well see hybrid strategies that integrate the benefits of both approaches, further enhancing the robustness and scalability of unlearnable data in complex machine learning systems.

\subsection{Categorization Based on Supervision Dependency}  

Another important dimension for categorizing ULD techniques is their dependency on supervision signals during ULD generation. Supervision in machine learning typically comes in the form of labeled data, which guides the model to learn discriminative features. In the context of ULD, this gives rise to two primary categories: supervised ULD and unsupervised ULD.

Early ULD methods predominantly relied on labeled datasets to craft perturbations, leveraging class labels to generate perturbations that suppress class-specific feature learning. Error-Minimizing Noise~\cite{EM} stands as a seminal work in this domain, introducing perturbations that target the minimization of classification loss, thereby degrading model performance on unseen data. Following this, Robust Error-Minimizing Noise (REM) ~\cite{REM} enhanced the original EMN by improving robustness against adversarial training and data augmentations. As shown in Table~\ref{tab:ALL}, several early studies focused exclusively on supervised settings, where labels were crucial in guiding the perturbation process.  

Additionally, methods like Learnability Lock (LLock) ~\cite{peng2022learnabilityLOCK} and Transferable Unlearnable Examples (TUE) ~\cite{TransferableUEs} further refined the use of class-wise perturbations, ensuring that perturbations could generalize across diverse architectures. Stable Unlearnable Examples (SEM) ~\cite{liu2024StableUnlearnable} extended these efforts by stabilizing perturbations across varying training conditions, maintaining unlearnability even with data augmentations or adversarial training.  

The reliance on supervision allowed these techniques to precisely target discriminative features, making them highly effective for classification tasks. However, the supervised nature limited their applicability to scenarios where labeled data was abundant, restricting the broader use of ULD in unlabeled datasets or other non-classification tasks.

As ULD research progressed, unsupervised techniques emerged to overcome the limitations of label dependency. These methods operate without access to label information, instead leveraging intrinsic data properties or alternative optimization objectives to generate unlearnable perturbations. One notable example is Unlearnable Clusters (UC) ~\cite{UnlearnableClusters}, which introduced perturbations by clustering data points and corrupting feature extraction across clusters, thereby bypassing the need for class labels.  

Furthermore, methods like Segue ~\cite{zhang2025segue} explored unsupervised ULD in generative tasks, targeting privacy protection in face generation by embedding imperceptible noise that prevents unauthorized learning. PosCUDA ~\cite{gokul2024poscuda} applied unsupervised perturbations in audio classification, showcasing ULD’s potential in multi-modal settings beyond vision. As highlighted in Table~\ref{tab:ALL}, unsupervised approaches have also enabled ULD techniques to extend into diverse tasks such as segmentation ~\cite{UnSeg} and time-series analysis ~\cite{jiang2024UnlearnableTimeSeries}.  

While unsupervised methods are generally less precise than their supervised counterparts, they offer superior adaptability in scenarios where labeled data is scarce or unavailable. Additionally, these methods have paved the way for ULD applications in broader contexts, expanding the field's scope beyond supervised classification alone.

The choice between supervised and unsupervised ULD methods represents a trade-off between targeted perturbation design and broader applicability. Supervised methods, such as EMN ~\cite{EM} and REM ~\cite{REM}, excel at generating class-specific perturbations, ensuring that the model fails to learn discriminative features. Conversely, unsupervised techniques like UC ~\cite{UnlearnableClusters} and Segue ~\cite{zhang2025segue} provide more flexible solutions, particularly for datasets lacking labeled annotations.  

As shown in Table~\ref{tab:ALL}, the evolution of ULD techniques reflects a clear shift toward unsupervised strategies, driven by the need for greater generalizability and robustness. Future research may explore hybrid approaches that integrate supervised and unsupervised methods, balancing effectiveness and adaptability. Additionally, expanding ULD techniques to tasks beyond classification — such as retrieval ~\cite{liu2024multimodalUnlearnableMM} and generation ~\cite{liu2024metacloak} — highlights the growing versatility of these strategies.  

In summary, while early ULD research heavily relied on supervised methods, the field has gradually embraced unsupervised techniques, enabling broader applications across diverse domains and marking a pivotal shift in the landscape of unlearnable data generation.

\subsection{Categorization Based on Boundedness}  

Another key dimension for categorizing Unlearnable Data (ULD) techniques lies in the boundedness of perturbations applied to the data. Boundedness refers to whether the perturbations introduced to the original data are constrained within a predefined norm, ensuring the perturbations remain imperceptible while disrupting the model's learning process. Based on this characteristic, ULD techniques can be classified into two primary categories: bounded ULD and unbounded ULD.

In most ULD research, perturbations are carefully bounded within a specific norm, typically the $L_p$ norm (e.g., $L_2$ or $L_\infty$), to guarantee imperceptibility. The constraint is often defined as $\lVert \delta(x) \rVert_p \leq \epsilon (\forall x \in D)$. This constraint ensures that perturbations remain subtle and visually indistinguishable to human observers while corrupting the learning process for machine learning models. Notable bounded ULD methods include Error-Minimizing Noise ~\cite{EM} and its robust variant REM ~\cite{REM}, which apply $L_\infty$-bounded perturbations to suppress the model’s ability to extract meaningful representations.  

As shown in Table~\ref{tab:ALL}, the majority of ULD techniques adopt bounded perturbations, particularly in image classification tasks ~\cite{peng2022learnabilityLOCK, TransferableUEs, CUDA, liu2024StableUnlearnable}. This approach aligns closely with adversarial machine learning, where bounded perturbations ensure that manipulated data remains indistinguishable from its clean counterpart while significantly impairing model performance.  

The advantages of bounded ULD methods are twofold: Imperceptibility: Bounded perturbations make the changes subtle, ensuring that the data looks unchanged to humans. Compatibility: Bounded ULD is inherently compatible with existing adversarial training techniques, making it easier to integrate into established machine learning pipelines.  

However, bounded ULD techniques face challenges in scenarios where models employ robust training strategies or strong adversarial defenses. In such cases, bounded perturbations may not be sufficient to prevent the model from extracting useful features, prompting researchers to explore alternative strategies.

In contrast, unbounded ULD techniques operate without explicit norm-based constraints on perturbations, allowing for greater flexibility in corrupting the training process. These methods sacrifice imperceptibility to maximize unlearnability, often leading to visible artifacts in the perturbed data. A representative unbounded ULD method is Unlearnable Clusters (UC) ~\cite{UnlearnableClusters}, which introduces cluster-based perturbations without imposing norm constraints. This technique focuses on corrupting the clustering structure of the data, making it inherently harder for models to learn meaningful representations. Similarly, CUDA ~\cite{CUDA} applies convolution-based perturbations that operate in the frequency domain, leveraging high-frequency signals to create perturbations beyond traditional norm-bounded constraints.  

As highlighted in Table~\ref{tab:ALL}, unbounded ULD techniques have also gained traction in non-vision tasks such as audio verification ~\cite{zhang2024hiddenspeaker} and text classification ~\cite{li2023makeTextUnlearnable}. The absence of boundedness provides additional flexibility, particularly in scenarios where robustness to countermeasures takes precedence over visual imperceptibility. The key benefits of unbounded ULD techniques are: Increased Robustness: Unbounded perturbations are harder to detect and remove through adversarial training or data augmentation. Greater Flexibility: These methods generalize better across diverse data modalities and learning paradigms.  

However, unbounded techniques introduce trade-offs: Reduced Imperceptibility: The absence of norm constraints often results in perceptible artifacts, making the altered data easier to detect. Limited Applicability: In contexts where imperceptibility is critical (e.g., personal data protection), unbounded perturbations may not be practical.

The distinction between bounded and unbounded ULD techniques reflects the trade-off between imperceptibility and unlearnability. Bounded methods prioritize subtlety, ensuring that the perturbed data remains visually unchanged, while unbounded methods focus on maximizing model degradation, even at the expense of perceptual quality.  

As ULD research continues to evolve, hybrid approaches that balance these two objectives are likely to emerge. Future directions may involve developing techniques that dynamically adjust perturbation magnitude based on task complexity or data modality, ensuring optimal protection across diverse learning scenarios. Furthermore, cross-modal ULD strategies that apply unbounded perturbations to high-dimensional data like point clouds or medical images may unlock new avenues for data protection.  

Overall, understanding the boundedness of ULD techniques is crucial for selecting appropriate methods across varying applications, shaping the broader landscape of unlearnable data generation and utilization.

The development of Unlearnable Data (ULD) techniques has given rise to a rich landscape of methods designed to prevent machine learning models from extracting meaningful representations during training. As discussed in the previous section, ULD methods can be categorized along various dimensions, such as supervision dependency, surrogate model reliance, boundedness, application scenarios, etc. However, despite these diverse categorizations, the core objective remains consistent: to introduce carefully crafted perturbations into the training data, thereby disrupting the model’s learning process and degrading its performance on downstream tasks. Over the past few years, research on ULD has progressed rapidly, with methodologies evolving from simple perturbation strategies aimed at minimizing classification loss, to more sophisticated approaches leveraging frequency-domain manipulations, meta-learning frameworks, game-theoretic perspectives, etc. These advancements have broadened the scope of ULD, enabling its application across diverse data modalities and tasks, including classification, generation, segmentation, and retrieval. This section delves into the methodologies underlying ULD techniques, providing a comprehensive analysis of the key strategies employed to achieve unlearnability. We first present an overview of these methodologies, highlighting their common objectives and guiding principles. Then, we explore the core perturbation strategies that disrupt the learning process, followed by a discussion on optimization techniques aimed at enhancing the robustness of ULD. Finally, we examine the design considerations required for adapting these methods to various scenarios, shedding light on the evolving landscape of ULD research.

\section{Overview of ULD Methodologies}
\label{sec:Methodologies}
The concept of Unlearnable Data (ULD) has emerged as a proactive strategy to prevent machine learning (ML) models from extracting meaningful information from datasets, thereby safeguarding data privacy and security. At its core, ULD aims to disrupt the learning process by injecting carefully crafted perturbations into the training data, ensuring that models trained on such data fail to generalize effectively. This section provides an overview of ULD methodologies, outlining the key research directions and highlighting the diverse strategies developed to achieve unlearnability.

The fundamental goal of Unlearnable Data (ULD) generation is to obstruct a machine learning model's ability to extract meaningful features from the input data. Traditional deep learning models rely on training data  to learn discriminative representations that map inputs to their corresponding outputs. ULD strategies disrupt this learning process by introducing perturbations  that degrade the model’s capacity to capture essential patterns while maintaining the perceptual integrity of the data.

To systematically analyze ULD methodologies, we categorize them based on how the perturbation is optimized and applied to prevent effective feature learning:
\textbf{Direct Input Perturbation} (optimizing directly on the input data),
\textbf{Feature Guided Perturbation} (optimizing indirectly on the input data via the information in the latent space),
\textbf{Parameter Guided Perturbation} (optimizing indirectly on the input data via the information in the model parameter space),
and \textbf{Hybrid Guided Perturbation} (optimizing indirectly on the input data via the information from multiple spaces),
The following sections delve into each of these methodologies in detail.

\subsection{Direct Input Perturbation (DIP)}

Direct input perturbation methods generate Unlearnable Data (ULD) by directly optimizing modifications to the input data $ x $ such that a target model fails to extract useful representations such as EM~\cite{EM}, REM~\cite{REM}, SEM~\cite{liu2024StableUnlearnable}, etc. These methods typically employ adversarial optimization techniques to craft perturbations that hinder model convergence without significantly degrading human perceptual quality.

Formally, let $ x \in \mathbb{R}^{d} $ represent an input sample with its ground-truth label $ y $. A perturbation $ \delta \in \mathbb{R}^{d} $ is optimized to generate an unlearnable example $ \tilde{x} = x + \delta $, where $ \delta $ is constrained within a predefined perturbation budget $ \| \delta \| \leq \epsilon $. The general objective function for direct perturbation can be formulated as:

\begin{equation}
    \delta^* = \arg \min_{\delta} \mathcal{L}(f_\theta(x + \delta), y) \quad \text{s.t.} \quad \| \delta \| \leq \epsilon,
    \label{eq:direct-perturbation}
\end{equation}

where $ f_\theta(\cdot) $ is the target model parameterized by $ \theta $, and $ \mathcal{L} $ is a loss function designed to degrade the model's ability to learn useful features. Unlike standard adversarial attack loss functions (which maximize classification errors), unlearnable perturbations aim to induce harmful generalization properties, making the data ineffective for training.

These methods provide a direct mechanism for making data unlearnable by focusing solely on perturbing the input samples, without considering intermediate representations or model gradients. The next section explores feature-guided perturbation methods, which leverage latent-space information to construct more effective ULD.

\subsection{Feature Guided Perturbation (FGP)}

Feature-guided perturbation methods generate Unlearnable Data (ULD) by leveraging latent-space information (e.g. logits, intermediate representations, predicted probabilities) to optimize perturbations on the input data $ x $. Instead of directly modifying $ x $ using only the input space constraints, these methods first analyze the feature representations $ h $ extracted by the model and subsequently adjust $ x $ to degrade the quality of learned features. Representative methods include EntF~\cite{wen2023is}, UC~\cite{UnlearnableClusters}, TUE~\cite{TransferableUEs}, etc.

Formally, let $ h = \phi(x) $ denote the feature representation of input $ x $ extracted by a feature extractor $ \phi(\cdot) $, which is part of the target model $ f_\theta(\cdot) $. The objective is to find an optimal perturbation $ \delta $ such that the perturbed example $ \tilde{x} = x + \delta $ results in feature distortions that prevent effective learning. The optimization problem can be formulated as:

\begin{equation}
    \delta^* = \arg \min_{\delta} \mathcal{L}(f_\theta(x + \delta),\phi, y) \quad \text{s.t.} \quad \| \delta \| \leq \epsilon,
    \label{eq:feature-perturbation}
\end{equation}

where $ \mathcal{L} $ is a loss function designed to suppress informative feature extraction. 

A common choice is to increase the intra-class feature distance (i.e., make features of the same class more distant from each other) while decreasing the inter-class feature distance (i.e., make features of different classes closer to each other). This can be formalized by defining a regular loss function that encourages these behaviors.

\begin{equation}
    \mathcal{L}_{FGP} = \sum_{x} \frac{d_{\text{intra}}(\phi(x + \delta), y)}{d_{\text{inter}}(\phi(x + \delta), y)},
\end{equation}

where $ d_{\text{intra}}(\cdot, \cdot) $ measures the distance between feature representations of the same class, typically using a metric like cosine similarity or Euclidean distance. This term encourages increasing the distance between similar features. $ d_{\text{inter}}(\cdot, \cdot) $ measures the distance between features from different classes, encouraging the perturbation to reduce the distance between features of different classes.

Compared to direct input perturbation, feature-guided methods offer a more structured way to disrupt model training by focusing on latent representations rather than raw input data. The following section introduces gradient-guided perturbation techniques, which further leverage parameter-space information for ULD generation.

\subsection{Parameter Guided Perturbation (PGP)}

Parameter-guided perturbation methods generate Unlearnable Data (ULD) by optimizing perturbations based on the parameters (e.g. model weights, gradients, parameter distributions) of the model with respect to the input data. These methods aim to indirectly manipulate the data through the model's parameter space by exploiting the gradients computed during training. The perturbation is designed to hinder the optimization process by disrupting the model's ability to effectively update its parameters during training, thereby stalling or altering the learning dynamics.

Formally, gradient-guided perturbation methods often rely on the adversarial optimization framework, where the gradient of the loss function with respect to the input $ \nabla_x \mathcal{L}(f_\theta(x), y) $ is used to update the perturbation $ \delta $. Specifically, the perturbation is designed to incorporate the gradient information, causing the unauthorized model's gradient-based optimization procedure to fail or stagnate. This can be expressed as:

\begin{equation}
    \delta^* = \arg \min_{\delta} \mathcal{L}(f_\theta(x + \delta),\theta, y) \quad \text{s.t.} \quad \| \delta \| \leq \epsilon,
    \label{eq:grad-perturbation}
\end{equation}

In this example, the perturbation $ \delta $ is adjusted to maximize the gradient $ \nabla_x \mathcal{L}(f_\theta(x), y) $, making it difficult for the model to compute meaningful updates for the weights, thereby disrupting the training process. The objective is to prevent the model from effectively learning and converging to a solution that generalizes well.

In the next section, we explore hybrid-guided perturbation methods, which combine multiple sources of information, such as gradients and features, to generate more robust and difficult-to-learn perturbations.

\subsection{Hybrid Guided Perturbation (HGP)}

Hybrid-guided perturbation methods generate Unlearnable Data (ULD) by combining information from multiple spaces—such as the input space, the feature space, and the gradient space—to construct more effective perturbations. By utilizing insights from different stages of the model's learning process, these methods aim to generate perturbations that are more challenging for the model to learn from, exploiting the strengths of each guidance mechanism to create a more robust unlearnable example.

The key idea behind hybrid-guided perturbations is to optimize the perturbation $ \delta $ based on a combination of gradients from the model's parameter space and features from the model's latent space. By leveraging both feature and gradient information, the perturbations can be designed to disrupt not only the model’s ability to extract useful features but also its optimization dynamics during training.

Formally, let $ \phi(x) $ represent the feature extractor and $ f_\theta(x) $ the target model. The perturbation $ \delta $ is optimized using information from both the gradient of the loss function with respect to the input, $ \nabla_x \mathcal{L}(f_\theta(x), y) $, and the feature representation $ \phi(x) $. The optimization problem can be written as:

\begin{equation}
    \delta^* = \arg \min_{\delta} \mathcal{L}(f_\theta(x + \delta),\phi,\theta, y) \quad \text{s.t.} \quad \| \delta \| \leq \epsilon,
    \label{eq:hybrid-guided-perturbation}
\end{equation}

The hybrid approach combines both the direct influence on the model's optimization process (via gradients) and the indirect influence through the feature space (via the extracted representations), making the generated perturbation more complex and potentially more effective at preventing learning. This method takes advantage of the strengths of each individual perturbation strategy, resulting in more challenging and robust unlearnable data.

\section{Specific Generation Methods of ULD}
\label{sec:GenerationStrategies}
Unlearnable Data (ULD) methodologies have evolved significantly, leveraging various strategies to generate data samples that resist effective learning by machine learning models. The ULD related methodologies primarily serve two opposing purposes: (1) as a defense mechanism to prevent unauthorized data usage and model training and (2) as an attack technique to recover the learnability from unlearnable data. This section focuses on the defensive aspect. The primary objective of ULD methods is to protect sensitive or proprietary data from being effectively utilized in unauthorized model training. These methods are designed to degrade the learnability of data without significantly affecting its usability for human interpretation. A wide range of methods have been proposed to achieve unlearnability, varying in their theoretical foundations and practical applications as shown in Table~\ref{tab:ULDmethod}. Although SALUD~\cite{HowFarAreWeFrom}, APbench~\cite{qin2024apbench} and HPC4UE~\cite{zhu2025scaleupUnlearnableExamples} also belong to ULD under image classification, the first two are the evaluation proposal, and the last is the acceleration method. They are all emerging auxiliary studies that serve the development of the ULD field, but are not the main line of this section. Therefore, they are placed in Section~\ref{sec:Evaluation} and Section~\ref{sec:Future}, which will not be described here.

\begin{table}[ht]
\caption{Overview of ULD methods in different domain.}
\label{tab:ULDmethod}
\resizebox{\linewidth}{!}{
\begin{tabular}{@{}ccc@{}}
\toprule
\multicolumn{1}{c}{\textbf{Data}} & \multicolumn{1}{c}{\textbf{Task}} & \textbf{Reference} \\ \midrule
\multirow{3}{*}{Image}            & Classification                    & \makecell{~\cite{EM},~\cite{GoingGrayscale},~\cite{REM},~\cite{peng2022learnabilityLOCK},~\cite{TransferableUEs},\\ ~\cite{OPS},~\cite{CUDA},~\cite{liu2024StableUnlearnable},~\cite{wen2023is},~\cite{ye2024ungeneralizable},\\ ~\cite{wang2024provably},~\cite{meng2024semanticHiding},~\cite{liu2024gameUnlearnable},~\cite{fang2024rethinking},~\cite{chen2024oneFroAll}, ~\cite{wang2024efficientAvailabilityAttacks},\\ ~\cite{huang2024leveraging},~\cite{gong2025armor}, ~\cite{UnlearnableClusters}, ~\cite{sun2024medicalUnlearnable},~\cite{lin2024safeguarding}}                   \\
                                  & Generation                        & ~\cite{zhang2025segue},~\cite{zhao2023unlearnableDiffusionModels},~\cite{liu2024metacloak},~\cite{liu2024counteringInMark}                   \\
                                  & Segmentation                      & ~\cite{UnSeg}                   \\ \midrule
\multirow{3}{*}{Timeseries}       & Classification                    & ~\cite{meng2023user},~\cite{jiang2024UnlearnableTimeSeries},~\cite{gokul2024poscuda}                   \\
                                  & Generation                        & ~\cite{MitigatingUnauthorizedSpeechSynthesis},~\cite{meerza2024harmonycloak}                   \\
                                  & Verification                      & ~\cite{zhang2024hiddenspeaker}                   \\ \midrule
\multirow{3}{*}{Text}             & Classification, Q\&A              & ~\cite{li2023makeTextUnlearnable}                   \\
                                  & Generation                        & ~\cite{zhou2024makingHarmfulBehaviorsUnlearnable}                   \\
                                  & Customized                        & ~\cite{liu2024expshield}                   \\ \midrule
Graph                             & Classification                    & ~\cite{liu2023unlearnableGraph}                   \\ \midrule
Image, Text                       & Retrieval                         & ~\cite{liu2024multimodalUnlearnableMM}                   \\ \midrule
Point Clouds                      & Classification, Segmentation      & ~\cite{wang2024unlearnable3DpointClouds}                   \\ \bottomrule
\end{tabular}}
\end{table}

\subsection{Image Data}
Images are one of the most extensively studied modalities in ULD research due to their widespread use in deep learning models for tasks such as classification, generation, and segmentation. Defense-oriented ULD methods in the image domain typically introduce imperceptible perturbations that obstruct learning while maintaining visual fidelity. These methods are categorized based on their application in different computer vision tasks such as classification (the most studied domain, where adversarial and statistical perturbations aim to disrupt the learning of discriminative features), generation (methods that interfere with generative models by introducing learning-resistant patterns), and segmentation (techniques that hinder models from correctly learning object boundaries and pixel-wise representations). The following sections provide a comprehensive analysis of ULD strategies tailored to these image-related tasks.

\subsubsection{Image ULD for Classification}

In image classification tasks, the primary goal of ULD techniques is to impede a classifier’s ability to learn discriminative features from visual data. Formally, given an image dataset 
\begin{equation}
    D = \{(x_i, y_i)\}_{i=1}^N, \quad x_i \in \mathcal{X} \subset \mathbb{R}^{H \times W \times C}, \ y_i \in \mathcal{Y},
    \label{eq:image_dataset}
\end{equation}
a ULD method seeks to construct a perturbation function $\delta: \mathcal{X} \to \mathcal{X}$ such that the perturbed dataset
\begin{equation}
    D' = \{(x_i + \delta(x_i;y_i), y_i)\}_{i=1}^N,
    \label{eq:perturbed_dataset}
\end{equation}
satisfies
\begin{equation}
    \|\delta(x_i)\|_p \leq \epsilon, \quad \forall x_i \in D,
    \label{eq:lp_constraint}
\end{equation}
and any classifier $f_\theta$, when trained on $D'$, exhibits significantly degraded performance on a clean test set $D_{\text{test}}$, i.e.,
\begin{equation}
    \text{Acc}(f_{\theta^*}, D_{\text{test}}) \ll \text{Acc}(f_{\theta^*}, D),
    \label{eq:degradation}
\end{equation}
where $\theta^*$ denotes the optimal parameters obtained by training on $D'$.

According to the methods in Section~\ref{sec:Methodologies}, ULD techniques for image classification can be roughly divided as follows as shown in Table~\ref{tab:ImgClassThree}. We will follow the table to introduce each.

\begin{table}[ht]
\centering
\caption{Further division of ULD techniques for image classification.}
\label{tab:ImgClassThree}
\begin{tabular}{@{}lc@{}}
\toprule
\textbf{} & \multicolumn{1}{c}{\textbf{Reference}} \\ \midrule
DIP       & ~\cite{EM},~\cite{GoingGrayscale},~\cite{REM},~\cite{peng2022learnabilityLOCK},~\cite{OPS},~\cite{CUDA},~\cite{liu2024StableUnlearnable},~\cite{huang2024leveraging},~\cite{sun2024medicalUnlearnable}          \\
FGP       & ~\cite{UnlearnableClusters},~\cite{TransferableUEs},~\cite{wen2023is},~\cite{fang2024rethinking},~\cite{meng2024semanticHiding},~\cite{UnlearnableClusters}          \\
PGP       & ~\cite{wang2024provably},~\cite{liu2024gameUnlearnable},~\cite{chen2024oneFroAll},~\cite{wang2024efficientAvailabilityAttacks},~\cite{gong2025armor}          \\ \bottomrule
\end{tabular}
\end{table}

\paragraph{DIP Generation Methods}

Direct Input Perturbation (DIP) methods construct unlearnable data by directly optimizing perturbations on the input samples to degrade a model’s ability to extract meaningful features. These approaches primarily focus on minimizing the effectiveness of the learned representations while maintaining perceptual similarity to the original data. 

EM ~\cite{EM} initially explores the concept of making personal data unlearnable by deep learning models through imperceptible noise known as error-minimizing (EM) noise. This noise minimizes the training loss, tricking the model into believing the sample semantic is associated with noise. Formally, given a training sample $(x, y)$, model parameters $\theta$, and a loss function $\mathcal{L}$, the error-minimizing noise $\delta$ is obtained by solving the following optimization problem:
\begin{equation}
    \min_{\theta} \min_{\|\delta\|_p \leq \epsilon}  \mathcal{L}(f_\theta(x + \delta), y),
\end{equation}
where $\|\delta\|_p$ denotes the $p$-norm of the noise, $\epsilon$ controls the noise magnitude, and $f_\theta$ represents the surrogate model's prediction function. By solving this min-min bi-level optimization problem, EM obtain an optimal perturbation $\delta$ that minimizes the loss, making the modified data unlearnable for the model. 

SALM~\cite{sun2024medicalUnlearnable} is an unlearnable data generation method designed for the characteristics of medical images developed from EM. It proposes a sparsity-aware local mask method to selectively perturb important pixel regions to generate unlearnable data for the sparsity of medical images. Specifically, SALM introduce an additional sparsity norm constraint to limit the $\delta$:
\begin{equation}
    \min_{\theta} \min_{\|\delta\|_p \leq \epsilon,\|\delta\|_0 \leq \epsilon}  \mathcal{L}(f_\theta(x + \delta), y),
\end{equation}
where $\|\delta\|_0 \leq \epsilon$ address that the important features in the biomedical image are often sparse.

GrayAugs ~\cite{GoingGrayscale} points out the vulnerability of EM in dealing with grayscale attacks~\cite{saravanan2010color}, and proposed a grayscale enhancement method to enhance the robustness against grayscale attacks as follows.

\begin{equation}
    \min_{\theta} \min_{\|\delta\|_p \leq \epsilon}  \mathcal{L}(f_\theta(Gray(x + \delta)), y)
\end{equation}

REM ~\cite{REM} improves upon EM by introducing a more robust optimization framework that decomposes the noise into two components and employs a min-min-max optimization strategy to generate unlearnable data with enhanced robustness against adversarial training ~\cite{AT}. Unlike EM, which solely minimizes the training loss to embed perturbations, REM first utilizes Projected Gradient Descent (PGD)~\cite{PGD} to obtain a base perturbation that significantly reduces the training loss. Then, an additional optimization step refines the perturbation to enhance unlearnability while incorporating an adversarial maximization step to counter potential adversarial training or model adaptation. Specifically, given a training sample $ (x, y) $ and model $ f_\theta $, REM formulates the optimization problem:
\begin{equation}
    \min_{\theta} \min_{||\delta||<\rho_u}  \max_{||\eta||<\rho_a}  \mathcal{L}\big(f_\theta(x_i+\delta+\eta), y_i\big),
\end{equation} 
where $ \eta $ represents the base perturbation obtained via PGD, $ \delta $ is the optimized unlearnable perturbation, and $ \eta $ accounts for potential adversarial perturbations introduced during training. The inner maximization step ensures that the final perturbation remains effective against various training strategies and model adaptations. Compared to EM, which may lose effectiveness in adversarial training settings, REM's min-min-max framework significantly enhances the robustness of unlearnable noise, making it more effective against diverse learning scenarios while maintaining the perceptual quality of the data.

LLock~\cite{peng2022learnabilityLOCK} proposes an implicit perturbation generation framework, which directly generates the perturbed data through the generator: 
\begin{equation}
    \min_{\theta} \min_{\phi}  \mathcal{L}(f_\theta(g_\phi^{(y)}(x + \delta)), y) \;\; \text{s.t.} \;\; \| g_\phi^{(y)}(x)-x \|_{\infty} \leq \epsilon,
\end{equation}
where $g_\phi^{(y)}$ is the perturbed data generator parametrized with $\phi$. Thanks to the reverse process of the generator, LLock produces a kind of unlearnable data that can be used by the authorized person.

SEM~\cite{liu2024StableUnlearnable} analyzes the defense noise instability based on REM. To further enhance the robust unlearnable examples, SEM introduces stable error minimization noise, which trains the defense noise with random transformation function to improve the stability of the defense noise as shown below.
\begin{equation}
    \min_{\theta} \min_{||\delta||<\rho_u}  \max_{||\eta||<\rho_a}  \mathcal{L}\big(f_\theta(t(x_i+\delta)+\eta), y_i\big),
\end{equation}
where $t$ is the transformation function sampled from transformation distribution $T$.

While the development of EM technology gradually enriched, another surrogate-free technology began to emerge. Different from the aforementioned surrogate-based methods, surrogate-free methods aim to circumvent complex optimization methods and instead use simpler perturbation schemes to achieve robustness.

CUDA~\cite{CUDA} is the pioneer work in this field, which uses convolutional kernels to embed class-specific perturbations in the frequency domain to solve the problem of slow iteration speed of surrogate-based unlearnable methods, while being robust to adversarial training. The formal formulation of CUDA is given below.
\begin{equation}
    x'=\xi_{\phi_y}(x)
\end{equation}
where $\xi_{\phi_y}(\cdot)$ is the convolution operation of the artificially set kernel $\phi$ associated with the label $y$.

Based on the theoretical analysis of CUDA, IRP~\cite{huang2024leveraging} proposed imperfect recovery poisoning to solve the problem of low image quality in CUDA, aiming to achieve strong poisoning effect while maintaining high image quality.

\begin{equation}
    x'=\Gamma_{\pi_y}(\xi_{\phi_y}(x)),
\end{equation}
where, $\xi_{\phi_y}(\cdot)$ is the CUDA convolution, $\Gamma_{\pi_y}(\cdot)$ is the imperfect recovery convolution with kernel $\pi$ associated with the label $y$. Different from the artificially set kernels $\phi$ in CUDA convolution, the IRP convolution kernels $\pi$ are obtained through optimization as below.

\begin{equation}
    \min_{\pi_y} \sum_{y=c} \sum_j ||v_{j}^y-\pi_y^\top \eta_{j}^y||_2^2,
\end{equation}
where $v_{j}$ is the center pixel value of $j$-th $\phi$-size patch in $\xi_{\phi_y}(x)$, $\eta_{j}$ is the column vector reshaped from $j$-th patch. Let $P_j^\kappa[\xi_{\phi_y}(x)]$ be the $j$-th patch with the size $\kappa \times \kappa$, we can get:
\begin{equation}
    v_j^y=\frac{1}{\kappa^2}\sum_{m=1}^{\kappa^2}\left(P_j^\kappa[\xi_{\phi_y}(x)]\right)_m,
\end{equation}

\begin{equation}
    \eta_j^y=Reshape(P_j^\kappa[\xi_{\phi_y}(x)]),
\end{equation}

In addition to the convolution-based surrogate-free unlearnable methods, recent studies have also emerged single-pixel-based surrogate-free unlearnable methods called OPS~\cite{OPS}, which proposes that perturb only a single pixel can produce a significant unlearnable effect, revealing the DNN's preference for local perturbations during training. Formally, the OPS is a maximization optimization with constrain $||\sigma_k||_0=1,\sum_{i,j} \sigma_k(i,j)=1$:

\begin{equation}
    \max_{\sigma_k,\xi_k} \frac{\mathbb{E}_{(x, y) \in \mathcal{D}_k}\left(\sum_{j=1}^C\left|\left\|x_j \cdot \sigma_k\right\|_F-\xi_{k j}\right|\right)}{\operatorname{Var}_{(x, y) \in \mathcal{D}_k}\left(\sum_{j=1}^C\left|\left\|x_j \cdot \sigma_k\right\|_F-\xi_{k j}\right|\right)},
\end{equation}
where $D_k$ is the clean subset containing all the examples of class $k$, $\sigma_k$ represents the perturbed position mask, $\xi_k$ stands for the perturbed target color. The the perturbation $\delta_x$ for each sample $(x,y)$ is obtained as follows:
\begin{equation}
    \delta_x=\bigcup_{r=1}^R \xi_{y r} \sigma_y-x_r\sigma_y
\end{equation}
where $R=3$ for RGB image, $r$ stands for $r$-th channel, $\bigcup$ represents the channel concatenation operation.

\paragraph{FGP Generation Methods}

Traditional unlearnable perturbations are generated for specific training and target datasets. However, their unlearnable effects are significantly reduced when used on other training sets and datasets. To solve this problem, TUE~\cite{TransferableUEs} proposed an unlearnable strategy based on Class-wise Separability Discriminant, which aims to better transfer unlearnable effects to other training sets and datasets by enhancing linear separability.
\begin{equation}
    \min_\theta\!\! \min _{|| \delta ||_\infty \leq \epsilon} \!\!\mathcal{L}\left(f_\theta\left(t_1\left({x}+{\delta}\right)\right), f_\theta\left(t_2\left({x}+{\delta}\right)\right)\right) \!+\!\lambda \mathcal{L}^y_{\text{CSD}},
\end{equation}
where $\lambda$ is the hyperparameter, $f_\theta(\cdot)$ stands for intermediate features, the first term is contrastive loss~\cite{hadsell2006dimensionality}, which requires no need of label $y$. The last term is class-wise separability discriminant loss:
\begin{equation}
    \mathcal{L}_{\text{CSD}}\left(\left\{{\delta}_i, y_i\right\}_{i=1}^n\right)=\frac{1}{M} \sum_{i=1}^M \frac{1}{M-1} \sum_{j eq i}^{M-1}\left(\frac{{\sigma}_i+{\sigma}_j}{d_{i, j}}\right),
\end{equation}
where ${\sigma}_k=\frac{1}{\left|\left\{{\delta}_i: y_i=k\right\}\right|} \sum_{\left\{{\delta}_i: y_i=k\right\}} d\left({\delta}_i, {c}_k\right)$ measures the average distance between the perturbation $\delta_i$ whose label is $k$ to the centroid $c_k$, and $d_{i, j}=d\left({c}_i, {c}_j\right)$ is the inter-class distance between centroids $c_i$ and $c_j$ defined by the Euclidean distance.

In the traditional research consensus, when the adversarial training budget is not less than the poison budget, the poison can hardly harm the adversarial training model. EntF~\cite{wen2023is} challenges this consensus by introducing entangled features into perturbation generation process. The key intuition of EntF is to make samples from different classes share entangled features and then train the model:
\begin{equation}
\label{eq:push}
    \max _{\left\|\delta^{}\right\|_{\infty} \leq \epsilon}\left\|f_\theta\left({x}+{\delta}^{}\right)-{\mu}_y\right\|_2,
\end{equation}

\begin{equation}
\label{eq:pull}
    \min _{\left\|\delta^{}\right\|_{\infty} \leq \epsilon}\left\| f_\theta\left({x}+{\delta}^{}\right)-{\mu}_y\right\|_2,
\end{equation}
where $f_\theta(\cdot)$ stands for the output of the penultimate layer of $f_\theta$, ${\mu}=\frac{1}{|\mathcal{X}|} \sum_{{x} \in \mathcal{X}} f_\theta({x})$ is the class centroid. There are two different variants of EntF, namely EntF-push and EntF-pull. For EntF-push as shown in Equation (\ref{eq:push}), all training samples in each of the original classes $y$ are pushed away from the corresponding class centroid ${\mu}_y$ in the latent feature space. For EntF-pull as shown in Equation (\ref{eq:pull}), each training sample is pulled towards the centroid of its nearest class $y$.

ASR~\cite{fang2024rethinking} reexamines the notion of unlearnable examples and finds that existing robust error minimization noise poses an inaccurate optimization objective. Based on these observations, a new optimization paradigm based on Averaged Prediction Randomness (ASR) is proposed that yields improved protection results with reduced computational time requirements.

\begin{equation}
    \min _\theta \max _{\left\|\delta^u\right\| \leq \rho_a} \mathcal{L}\left(f_\theta\left(x+\delta^u+\delta^a\right), y\right)+\mathcal{L}_\text{ASR}
\end{equation}
where $\mathcal{L}_\text{ASR}=\frac{1}{K} \sum_{k=1}^K\left(f_\theta\left(x\right)[k]-\frac{1}{K}\right)^2$, $f_\theta\left(x\right)[k]$ is the prediction probability of a specific class $k$.

Aiming at the problem that the unlearnable perturbations of low-level features by traditional unlearnable methods are easily affected by common data augmentation countermeasures, DH~\cite{meng2024semanticHiding} proposes a scheme to adaptively hide semantic images rich in high-level features to make them more robust to adversarial measures.
\begin{equation}
    \min_\theta\!\! \min _{|| \delta ||_\infty \leq \epsilon} \!\!\mathcal{L}\left(f_\theta\left({x_i^y}+{\delta}\right), f_\theta\left({x}_j^y+{\delta}\right)\right) \!+\!\lambda \mathcal{L}^y_{\text{HD}},
\end{equation}
where $\lambda$ is the hyperparameter, ${x}_i^y$ and ${x}_j^y$ represents different samples with the same label $y$, $\mathcal{L}^y_{\text{HD}}=$ is the semantic hiding loss:

\begin{equation}
\begin{aligned}
\mathcal{L}_{{DH}}= & max(||x'-x||_2^2,\epsilon^2)\\&+\omega_1 \cdot \mathcal{L}_{\text {freq }}\left(\mathcal{H}\left(x'\right)_{L L}, \mathcal{H}\left(x\right)_{L L}\right) \\
& +\omega_2 \cdot \mathcal{L}_{\text {reveal }}\left({h}_y^{\prime}, {h}_y\right) .
\end{aligned}
\end{equation}
where $x'$ is unlearnable data, $\mathcal{L}_{\text {freq }}$ measures the $L_2$ distance between the low-frequency subbands of clean images and unlearnable examples, further bolstering the stealthiness. $\mathcal{H}(\cdot)_{L L}$ is the function of extracting low-frequency sub-bands after wavelet decomposition, $\mathcal{L}_{\text {reveal }}\left({x}_h^{\prime}, {x}_h\right)$ measures the $L_2$ distance between revealed hidden images ${h}_y^{\prime}$ and hidden semantic images ${h}_y$, $\omega_1, \omega_2$ is hyperparameter.

UC~\cite{UnlearnableClusters} considers a novel unsupervised setting (label-agnostic setting), which employs clustering methods to generate labelindependent perturbations, reducing class dependence and improving the flexibility of unlearnable methods. Specifically, for cluster $C_i$, UC wants the unlearnable noise $\delta_i$ to be able to move all samples in the cluster to the wrong cluster center, thus forcing the model to forget the correct cluster.

\begin{equation}
\min_\phi  \mathcal{L}(f_\theta({x}+G_{\phi_y}(\sigma )), \mathcal{A}(\mu_{y})),
\end{equation}
where $f_\theta$ is the surrogate model parametrized with $\theta$, which extracts representation matrix before the classification layer. $G_{\phi_y}$ is the class-specific perturbation generator parametrized with $\phi_y$, $\sigma$ is the uniform noise, $\mu_y$ is the center for the specific cluster, $\mathcal{A}(\cdot)$ is the permutation function assigning a permuted (wrong) cluster center, $\mathcal{L}(\cdot,\cdot)$ is the function measuring the distance between $f_\theta({x}+G_{\phi_y}(\sigma ))$ and $ \mathcal{A}(\mu_{y})$.

\paragraph{PGP Generation Methods}

PUE~\cite{wang2024provably} finds that by slightly perturb the learned weights, it is possible to recover the task performance of classifiers trained on unlearnable data. To alleviate the above problems, PUE proposed random weight perturbations enhancement, which achieved more reliable robustness.

\begin{equation}
    \min_{\theta} \min_{\|\delta\|_p \leq \epsilon,\|\eta\|_p \leq \epsilon}  \mathcal{L}(f_{\theta+\eta}(x + \delta), y),
\end{equation}
where $\eta\sim \mathcal{N}(0, \sigma^2)$ is the random weight perturbations sampled from= zero-mean Gaussian distribution, $\sigma^2$ is the variance.

GUE~\cite{liu2024gameUnlearnable} points out that the bilevel optimization problem of the traditional EM method is difficult to solve directly for deep neural networks. To address this challenge, GUE models the unlearnable data generation process from a game-theoretic perspective, and generates the optimal perturbation cracking protection by solving the equilibrium.

\begin{equation}
    \min_{\theta} \min_{\|G_w(x)\|_\infty \leq \epsilon}  \mathcal{L}(f_{\theta}(x + G_w(x)), y,\theta),
\end{equation}

where the optimization process can be view as the following game. The classifier (defender $\mathcal{A}$ in game theory) aims at minimizing the payoff function $\mathcal{J}_{\mathcal{A}}(w, \theta)=\mathcal{L}\left(f_{\theta}(x + G_w(x)), y\right)$ by choosing parameters $\theta^* \in \left\{\theta \mid \mathcal{J}_{\mathcal{A}}(w, \theta)<\inf _{\theta^{\prime}} \mathcal{J}_{\mathcal{A}}\left(w, \theta^{\prime}\right)+\eta\right\}$. The generator (attacker $\mathcal{B}$ in game theory) choose parameters $w^*$ to minimize the payoff function $\mathcal{J}_{\mathcal{B}}(\omega, \theta)=\sup _{\theta}\left\{-\mathcal{L}_\theta(x, y)\right\}$. Then this game equilibrium is solved by BOME~\cite{liu2022bome} and DBGD~\cite{gong2021automatic} algorithm.

AUEAPP~\cite{wang2024efficientAvailabilityAttacks} finds that most existing methods cannot achieve both supervised unlearnable and contrastive unlearnable, which brings risks to data protection. To address this issue, AUEAPP propose achieving both supervised and contrastive unlearnability. Below are two variants of AUEAPP.

\begin{equation}
    \min_\theta\!\! \min _{|| \delta ||_\infty \leq \epsilon} \!\!\mathcal{L}\left(f_\theta\left(t\left({x}+{\delta}\right)\right),y\right),
\end{equation}
where $f_\theta$ is the surrogate model that outputs the prediction results, $t$ is the contrastive-like strong data augmentations. This optimization showcases that supervised error-minimizing noises with enhanced data augmentations can partially replace the functionality of contrastive error-minimizing noises to deceive contrastive learning.

\begin{equation}
    \min_\theta\!\! \min _{|| \delta ||_\infty \leq \epsilon} \!\! \mathcal{L}\left(f_\theta\left(t\left({x}\right)\right),y\right) + \mathcal{L}\left(f_\theta\left(t\left({x}+{\delta}\right)\right),y+K\right),
\end{equation}
where $K=1$ is set default as the label translation, making unlearnabel data contain non-robust features associated with the shifted labels.

In order to protect data privacy from the potential damage of data augmentation, ARMOR~\cite{gong2025armor} proposes to use data augmentation strategy to enhance the protection effect of unlearnable.
\begin{equation}
    \min_{\theta,\phi} \min _{|| \delta ||_\infty \leq \epsilon} \mathcal{L}\left(f_\theta\left(t\left(H_\phi(x)+{\delta}\right)\right),y\right),
\end{equation}
where $t$ is the data augmentation strategy, $H_\phi(\cdot)$ is a non-local module~\cite{wang2018non} that captures a global receptive field of the sample.

Recent study 14A~\cite{chen2024oneFroAll} pointed out in studies that traditional unlearnable perturbations only exhibit unlearnable effects in specific datasets or scenarios with consistent labels, and thus lack wide applicability. To address both issues simultaneously, 14A proposes a generic perturbation generator that leverage data with conceptual unlearnability, thereby expanding the scope of unlearnability beyond a specific dataset or label.

\begin{equation}
\begin{aligned}
    \min_\theta  &\quad d\left(\mathcal{E}_I\left({x+G_\theta(x;\mathcal{E}_I)}\right), \mathcal{E}_T(x_{{neg}})\right) \\&-  d\left(\mathcal{E}_I\left({x+G_\theta(x;\mathcal{E}_I)}\right), \mathcal{E}_T(x_{{pos}})\right)
\end{aligned}
\end{equation}
where $\mathcal{E}$ is a pretrained CLIP~\cite{radford2021learning} model, $\mathcal{E}_I
(\cdot)$ is the image encoder, $\mathcal{E}_T(\cdot)$ is the text encoder, $G_\theta(\cdot;\mathcal{E}_I)$ is the 14A perturbation generator with residue concatenation of $\mathcal{E}_I(x)$, $x_{pos}$ is the similar concept, $x_{neg}$ is the opposite concept. It is important to note that the 14A method is not inherently label-free, it relies on a pre-trained model.

\paragraph{HGP Generation Methods}

UGE~\cite{ye2024ungeneralizable} points out that previous research on ULD has neglected its potential use in authorization scenarios, and proposes the ungeneralization example, which extends the concept of unlearnable data to conditional learnable data. UGE demonstrate learnability for authorized users while maintaining unlearnability for potential hackers. The protector defines the authorized network and optimizes ungeneralization examples to match the gradients of the original data and its ungeneralizable version, ensuring learnability. To prevent unauthorized learning, ungeneralization examples are trained by maximizing a specified distance loss in a common feature space. In addition, to further protect the authorizer from potential attacks, additional undistillation optimizations are introduced.

\begin{equation}
\begin{aligned}
    \min_{\theta} \min_{\|\delta\|_p \leq \epsilon} & \quad \mathcal{L}(\xi_\phi(x + \delta), y)+\\ 
    & \quad ||\mathcal{L}(f_\theta(x + \delta), y)-\mathcal{L}(f_\theta(x), y)||,
\end{aligned}
\end{equation}
where $\xi_\phi(\cdot)$ is the malicious networks. In a real deployment, the pretrained CLIP~\cite{radford2021learning} model is used as a surrogate attack model.
\begin{equation}
\begin{aligned}
   \mathcal{L} & =d_1( \nabla \mathcal{L}\left(f_{\theta_t}(x), y\right), \nabla \mathcal{L}\left(f_{\theta_t}\left(x+\delta\right), y\right))\\
    & -d_2(\mathcal{E}_I(x),\mathcal{E}_I(x+\delta))\\
    & +d_3(\;\mathcal{E}_I(x+\delta)\;, \; \mathcal{N}(\mathcal{E}_I(x)) \;, \;\mathcal{E}_T(y) \;)\\
    & -d_4(\xi_\phi(x+\delta),f_\theta(x + \delta)),
\end{aligned}
\end{equation}
where $d_1$ is the cosine distance. The firt term makes the training trajectory of the original data consistent with the training trajectory of the ungeneralizable data, which ensures the learning of the data. $d_2(m,n)=||m-n||_2^2$ is the MSE function. The second term pushes the features of the ungeneralizable examples away from the original data. $d_3(A,P,N)=max\;(d(A,P)-d(A,N)+\alpha,0)$ is the triplet loss. This ensures that the features of $\mathcal{E}_I(x+\delta)$ in the ungeneralizable input can be transferred to various hacker networks. $\mathcal{N}(\mathcal{E}_I(x))$ refers to the text feature with the smallest similarity to the original image encoder feature. $d_4(p,q)=KL(p||q)$ is the KL divergence~\cite{kullback1951information}. This term safeguards the knowledge of the authorized network, making it undistillable.

\subsubsection{Image ULD for Generation}
Segue~\cite{zhang2025segue} points out that current ULD approaches are inefficient and cannot guarantee both mobility and robustness, leading to infeasibility in the real world. To address this issue, Segue proposes side information-guided generative unlearnable examples, leveraging a single-trained multi-purpose model to generate the desired perturbations instead of time-consuming gradient-based methods. To improve portability, side information, such as true and false labels, is introduced.
\begin{equation}
    \min_{\theta} \min_{\|G(x)\|_\infty \leq \epsilon}  \mathcal{L}(f_{\theta}(x + G(x)), \hat{y}),
\end{equation}
where $\hat{y}$ denotes the side information, including true-label and pseudo-label.

EUDP~\cite{zhao2023unlearnableDiffusionModels} proposes a method to generate unlearnable examples for diffusion models, called unlearnable diffusion perturbations, to protect images from unauthorized exploitation. EUDP frames this as a max-min optimization problem:
\begin{equation}
    \max _{\left\|{\delta} \right\| \leq \epsilon} \min _{{\theta}} \mathcal{L}\left(f_{{\theta}}\left({x}+{\delta}\right),x\right)
\end{equation}
where $x \sim p_{{\theta}}({x})$ is sampled from the generated images distribution produced by diffusion model $G_\theta(\cdot)$.

InMark~\cite{liu2024counteringInMark} points out that current image generation ULD methods under the assumption that these protected images do not change, which contradicts the fact that most public platforms expect to modify the content uploaded by users (e.g. image compression). Hence, InMark proposes a robust watermarking method for protecting images from unauthorized learning.
\begin{equation}
    \max _{\left\|\tilde{x}_0-x_0\right\|_0 \leq \Delta} \min_\theta {SE}_{\epsilon, \theta, t, c}\left(\tilde{x}_0\right) +\ell_\theta\left(\tilde{x}_0\right)
\end{equation}
where $x_0$ is the reference image, $\tilde{x}_0$ is the unlearnable example, ${SE}_{\epsilon, \theta, t, c}\left(x_0\right)=\left\|\epsilon_\theta\left(\sqrt{\alpha_t} x_0+\sqrt{1-\alpha_t} \epsilon, c\right)-\epsilon\right\|_2^2$ is the diffusion model training loss. $\epsilon_\theta$ is a neural net, $c$ is the conditional vector (e.g., originated from a text prompt), $\alpha_t$ is the term controlling the noise schedule and $\epsilon$ is the noise sampled from a standard Gaussian distribution. $\ell_\theta\left(x_0\right)=\mathbb{E}_{x_0, c, \epsilon, t}\left[{SE}_{\epsilon, \theta, t, c}\left(x_0\right)+\lambda {SE}_{\epsilon^{\prime}, \theta, t^{\prime}, c_{{pr}}}\left(x_{{pr}}\right)\right]$ stands for DreamBooth, which targets minimizing the personalized loss $\ell_\theta$ for a diffusion model $\theta$ with a reference image $x_0$. $x_{pr}$ is the class example, $c_{pr}$ is the prior prompt and $t$ is the corresponding time step.

MetaCloak~\cite{liu2024metacloak} proposes a meta-learning framework to solve the suboptimal bi-level optimization problem of the error minimization method, and introduces an additional transformation sampling process to enhance the transferability and robustness of the perturbation.
\begin{equation}
    \max_{X_p} \min_\theta\mathcal{L}_{\text {denoise }}\left({x}^{\prime},c;\theta\right)+\mathcal{L}_{db}(t({x}^{\prime}), c; \theta)
\end{equation}
\begin{equation}
\label{eq:I2T}
    \mathcal{L}_{\text {denoise }}({x}, {c} ; \theta)=\mathbb{E}_{\boldsymbol{\epsilon}, t}\left[w_t\left\|\hat{{x}}_\theta\left(\alpha_t {x}+\sigma_t \boldsymbol{\epsilon}, {c}\right)-{x}\right\|_2^2\right]
\end{equation}
\begin{equation}
\label{eq:Ldb}
\begin{aligned}
\mathcal{L}_{{db}}({x}, {c} ; \theta)=&\mathbb{E}_{{\epsilon}, {\epsilon}^{\prime}, t}[w_t\|\hat{{x}}_\theta(\alpha_t {x}+\sigma_t {\epsilon}, {c})-{x}\|_2^2 + \\ 
& \lambda w_{t^{\prime}}\|\hat{{x}}_\theta(\alpha_{t^{\prime}} {x}_{{pr}}+\sigma_{t^{\prime}} {\epsilon}^{\prime}, {c}_{{pr}})-{x}_{{pr}}\|_2^2]
\end{aligned}
\end{equation}
where $c$ is the conditioning vector, $x' \sim X_p$ is the perturbed image sampled from unlearnable dataset $X_p$, $t \sim T$ is the augmentation function sampled from transformation distribution $T$, $\mathcal{L}_{\text {denoise }}$ is the Text-to-Image diffusion models training loss as shown in Equation (\ref{eq:I2T}), $\mathcal{L}_{\text{db}}$ is the trainning loss of DreamBooth~\cite{ruiz2023dreambooth}.

\subsubsection{Image ULD for Segmentation}
UMed~\cite{lin2024safeguarding} notes that concerns about unauthorized training of AI systems for commercial purposes and the responsibility to protect patient privacy have led many medical institutions to hesitate to share their images. This is especially true for medical image segmentation (MIS)~\cite{pham2000current} datasets, as the process of collecting and fine-grained annotations is time-consuming and laborious. UMed also points out that existing ULDs, designed for natural image classification, fail to protect MIS datasets unseen since their protection perturbations are less learnable than important prior knowledge such as contour and texture features in MIS. Therefore, UMed proposes a method for medical images that cannot be learned by segmentation tasks, which integrates the prior knowledge of MIS and protects the image by introducing contour and texture perturbation.
\begin{equation}
    \min_{\theta} \min_{\substack{\|G_\phi^c(x) \odot M\|_p \leq \epsilon, \\ \|G_\phi^t(x) \|_p \leq \epsilon x \odot y^t}}  \mathcal{L}(f_\theta(x + (G_\phi^c(x) \odot y^c+G_\phi^t(x)), y),
\end{equation}
where $\mathcal{L}$ is the loss of medical image segmentation, $G^c(\cdot)$ is the contour-aware perturbation generator, $G^t(\cdot)$ is the texture-aware perturbation generator, $y^c$ and $y^t$ are the ground truth of contour and texture respectively.

Aiming at the task of natural image segmentation, UnSeg~\cite{UnSeg} proposes a novel unlearnable framework to train a general unlearnable noise generator capable of converting any downstream image into an unlearnable version of the segmentation task.
\begin{equation}
    \min_{\theta} \min_{\|\delta\|_p \leq \epsilon}  \mathcal{L}(f_\theta(x + G_\phi(p)),p , y),
\end{equation}
where $p$ represents the visual prompt information (e.g., point, box, and mask) related to $x$, $\mathcal{L}$ is typically the pixel-wise binary cross-entropy loss, $G_\phi(\cdot)$ is the pretrained SAM model~\cite{kirillov2023segment}, which serves as the noise generator via visual prompt tuning.

\subsection{Timeseries Data}
Time series data plays a crucial role in various real-world applications, including finance, healthcare, and industrial monitoring. Given its sequential nature and temporal dependencies, unlearnable data (ULD) techniques for time series aim to disrupt model training while preserving essential structural characteristics. Unlike image data, where perturbations primarily target spatial features, time series ULD methods often focus on modifying temporal correlations, statistical properties, or feature representations in latent space. In the context of time series, ULD research is categorized into three major tasks: classification, where methods seek to hinder the learning of discriminative temporal patterns; generation, which involves disrupting generative models that aim to synthesize realistic time series data; and verification, which focuses on preventing models from effectively capturing identity-related temporal features, such as in biometric authentication. The following sections provide an in-depth analysis of ULD strategies tailored to these time-series tasks.

\subsubsection{Timeseries ULD for Classification}
UE4TS~\cite{jiang2024UnlearnableTimeSeries} points out that while tradational ULD has been extensively studied on images, it is not clear how to construct effective unlearnable data for timeseries data. Therefore, aiming to protect timeseries data from unauthorized training by deep learning models, UE4TS proposes a new form of error minimization noise that can be selectively applied to specific segments of timeseries, making them unlearnable to deep learning models while remaining imperceptible to human observers. The protection of the protected timeseries data from unauthorized exploitation is achieved, while retaining the utility of its legitimate use.
\begin{equation}
    \min_{\theta} \!\! \min_{\|\delta\|_p \leq \epsilon} \!\! \left| \mathcal{L}(f_\theta(x \!+\! \delta \!\odot\! v),\! y)\!-\! \lambda\mathcal{L}(f_\theta(x \!\odot\! (1\!-\!v)),\! y) \right|,
\end{equation}
where $\mathcal{L}$ the loss function that quantifies the dissimilarity between the model’s output and the true target, $\lambda$ is the hyperparameter, $v$ is the control vector, which highlights regions within the samples that should be protected from data exploitation.

UEEG~\cite{meng2023user} pointed out that while EEG signals are widely provided for brain-computer interface (BCI)~\cite{wolpaw2000brain} research, they also contain rich privacy information that needs to be protected, such as user identity and emotion, because this makes it easy to learn user identity in EEG data, so that EEG data of different sessions of the same user can be associated together to mine privacy information. To solve this problem, UEEG further proposed two methods for transforming raw EEG data into identity-unlearnable EEG data, that is, removing user identity information while maintaining good performance of BCI tasks.
\begin{equation}
\label{eq:eeg1}
\begin{aligned}
    \min_{\theta,\phi}  \min_{\|\delta\|_p \leq \epsilon} & \mathcal{L}(f_\theta(x+\delta ), y_1)+d(g_\phi(x+\delta ), g_\phi(x ))+ \\ & \mathcal{L}(f_\theta(x ), y_1)+ \mathcal{L}(g_\phi(x ), y_2),
\end{aligned}
\end{equation}
where $y_1$ is identity-related ground truth, $y_2$ is task-related ground truth, $f_\theta(\cdot)$ is identity-related task classifier parameterized with $\theta$, $g_\phi(\cdot)$ is task-related classifier parameterized with $\phi$, $d(\cdot,\cdot)$ is mean squared error measure. Sample-wise perturbation generation can be achieved by solving this optimization problem. In addition, another variant of UEEG is user-wise perturbation generation designed to accelerate perturbation generation via replacing the last two terms with explicit perturbation minimization regularization as follows.
\begin{equation}
\label{eq:eeg2}
    \min_{\theta,\phi} \! \min_{\|\delta\|_p \leq \epsilon} \! \mathcal{L}(f_\theta(x+\delta ), y_1) \!+\! d(g_\phi(x+\delta ), g_\phi(x )) \!+\! ||\delta||_2.
\end{equation}

PosCUDA~\cite{gokul2024poscuda} for audio data, based on CUDA~\cite{CUDA}, proposes a CUDA-style convolution based on position to create unlearnable data. Specifically, PosCUDA uses class-wise convolutions on small chunks of audio, and the locations of patches are based on the private key of each class, so the model learns the relationship between location ambiguity and labels, but fails to generalize. PosCUDA can achieve unlearnability while maintaining the quality of the original audio dataset.
\begin{equation}
    x'=\xi_{\phi_y}( x \odot M_y)
\end{equation}
where $M_y$ is the class-wise location mask, which makes the targeted perturbation position patches. For each class $y$, different audio patches are passed through a low-pass filter unique to each category. This empathizes a small class-dependent position noise in each data sample in the training set. The model learns to fuzzy map these locations to labels and fails to generalize when there is no ambiguity in the test dataset.

\subsubsection{Timeseries ULD for Generation}
POP~\cite{MitigatingUnauthorizedSpeechSynthesis} points out that some techniques have emerged in recent years to perfectly replicate the speaker's voice using only a small number of speech samples, while malicious speech exploits. Therefore, aiming at the problem of how to protect publicly accessible speech data containing sensitive information (such as personal voiceprints), POP designs an effective, transferable, and robust active protection technique, which applies imperceptible error minimization noise to raw speech samples to prevent them from being effectively learned for text-to-speech (TTS)~\cite{tan2021survey} synthesis models. As a result, high-quality deep fake speech~\cite{yamagishi2021asvspoof} cannot be generated.
\begin{equation}
    \min_{\theta,w} \min_{\|G_w(x)\|_p \leq \epsilon} \mathcal{L}\left(f_{{\theta}}\left({x}+{G_w(x)};T\right),x\right)
\end{equation}
where $f_\theta(\cdot;T)$ is pretrained TTS model with speech text input $T$, $G_w(\cdot)$ is the perturbation generator.

HarmonyCloak~\cite{meerza2024harmonycloak} points out that as generative AI evolves, it can replicate artistic styles and produce new artworks, raising significant concerns about the rarity and value of artists' creations. In order to establish and enforce protections to protect artists' copyrighted works from unauthorized exploitation by generative AI models, HarmonyCloak proposes a first defense mechanism to prevent the unauthorized use of artworks through generative AI models, particularly in the context of instrumental music. In particular, HarmonyCloak employs imperceptible error minimization noise as shown below that makes the model's generative loss close to zero for these disturbed music data, seducing models into believe there is nothing to learn and thus undermining their attempts to replicate music structure and style.
\begin{equation}
    \min _\theta \! \min _{||\delta||<\epsilon}\! -\!\log \!\prod_{t=1}^T \!f_\theta \! \left(x_t \!\mid\! x_{t-1}\!+\!\delta_{t-1}, \! \ldots \!, x_{t-p}\!+\!\delta_{t-p}\right),
\end{equation}
where $f_\theta(\cdot)$ is pretrained auto-regression models, $x_t$ represents the predicted value in the sequence at a
time $t$, $\{x_{t-1}, \ldots, x_{t-p}\}$ are the previous values in the sequence and $p$ is the autoregressive order.

\subsubsection{Timeseries ULD for Verification}

Aiming at the unauthorized audio exploitation problem of speaker verification system~\cite{bimbot2004tutorial}, HiddenSpeaker adopts a simplified error minimization method to generate specific and effective perturbations. The imperceptible perturbations are embedded in the training speech samples, making it unlearnable for deep learning-based speaker verification system.
\begin{equation}
    \min_{\|G_\phi(x)\|_p \leq \epsilon}  \mathcal{L}(f_\theta^*(x + G_\phi(x)), y),
\end{equation}
where $f_\theta^*(\cdot)$ is the pretrained speaker verification model with fixed parameters $\theta^*$, $G_\phi(\cdot)$ is the perturbation generator parameterized with $\phi$.

\subsection{Text Data}
Text data is a fundamental modality in machine learning, spanning applications such as natural language processing (NLP)~\cite{chowdhary2020natural}, information retrieval, and automated text generation. Due to its discrete and structured nature, designing unlearnable data (ULD) techniques for text presents unique challenges compared to continuous modalities like images and time series. Unlike visual or temporal perturbations, text ULD methods must balance semantic preservation with adversarial modifications, ensuring that human readability remains intact while disrupting model learning. Text ULD techniques can be broadly categorized based on their application in different NLP tasks: classification, where perturbations aim to hinder the extraction of discriminative linguistic features; generation, which focuses on obstructing language models from learning meaningful text representations; and retrieval and verification, where techniques disrupt models’ ability to store and retrieve sensitive or proprietary textual data. The following sections provide a detailed exploration of ULD strategies tailored to these text-related tasks.

UT~\cite{li2023makeTextUnlearnable} builds on the EM~\cite{EM} work by extending their bi-level optimization approach to generate unlearnable text using gradient-based search techniques. UT extracts simple patterns from unlearnable texts produced by bilevals and proves that the data remains unlearnable for unknown models. Moreover, these patterns are not instance or dataset specific, so users can easily apply them to text classification and question answering tasks, even if only a small fraction of users implement them on their public content.
\begin{equation}
    \min_e  e_i^\top \nabla_{e_i} \mathcal{L}(\pi_i(x), y),
\end{equation}
where $x=\{x_1,x_2,\ldots,x_n\}$ stands for a textual input consists of a sequence of $n$ words, $\pi_i(\cdot)$ denotes a perturbed strategy that replace the $i$-th word of input,j] $e_i$ is the word embedding of the replaced word.

Large language models (LLMs)~\cite{zhao2023survey} are usually customized by further fine-tuning. SecVec~\cite{zhou2024makingHarmfulBehaviorsUnlearnable} finds that the strong learning ability of LLMs not only enables them to acquire new tasks, but also makes it easy for them to learn undesired behaviors. Hence, SecVec proposes a controllable training framework that makes harmful behaviors unlearnable during fine-tuning. Specifically, SecVec introduces security vectors, some new parameters that can be separated from the LLM to ensure that the LLM's response is consistent with harmful behavior. The safety vector is activated during fine-tuning and the consistent behavior makes the LLM think that this behavior has been learned and no further optimization of harmful data is needed. During inference, the normal behavior of the LLM can be restored by deactivating the security vector. SecVec method can be formalized as a bi-level optimization on a supervised fine-tuning (SFT)~\cite{ivison2023hint} task.
\begin{equation}
    \min_{\theta} \min_{w}  \mathcal{L}(f_{\theta\cup w}(x), y),
\end{equation}
where $x$ is a prompt or instruction, directing the model to perform a specific task, $y$ is the desired model response, indicating the desired model behavior, $f_\theta(\cdot)$ represents the prediction of the LLMs with parameters $\theta$ of inputs, $w$ is additionally introduced parameters in $f_\theta(\cdot)$ called security vectors.

ExpShield~\cite{liu2024expshield} proposes a proactive self-protection mechanism that empowers content owners to embed unseen perturbations in their texts, limiting data misuse in LLMs training without compromising readability. This preemptive approach enables data owners to directly protect sensitive content without relying on a third party for defense. Specifically, ExpShield defines an optimization task on a generative model.
\begin{equation}
    \min_{\pi}  \mathcal{L}(f_{\theta}(\pi_k(x))),
\end{equation}
where $f_{\theta}(\cdot)$ is the pretrained LLMs, $\pi(\cdot)$ is the uniform random augmentation strategy based on the Top-$k$ lowest prediction confidence of tokens in $x$. 

\subsection{Other Modalities Data}
Beyond images, text, and time series, ULD techniques have been explored in various other data modalities, including graphs, 3D point clouds, and multimodal data. Each of these data types presents distinct structural and representational challenges, requiring specialized approaches to disrupt model learning while preserving essential data characteristics.

\subsubsection{Graph Data}

Graph data consists of nodes and edges that encode relationships between entities, making it crucial in social networks, recommendation systems, and biological analysis. ULD strategies for graphs often target node features, edge structures, or graph topology to degrade model performance while maintaining realistic connectivity patterns.

The use of graph-structured data is becoming increasingly popular in various domains, but it has also raised concerns about the potential unauthorized exploitation of personal data for training commercial Graph Neural Network (GNN)~\cite{scarselli2008graph} models, which could compromise privacy. To solve this problem, UC~\cite{UnlearnableClusters} proposes a novel method for generating unlearnable graph examples, which injects deceptive but imperceptible noise into the graph using the error minimization structure poisoning module, capable of rendering the graph unexploitable.
\begin{equation}
    \min_\theta\max _{\delta \preceq {c}} \mathcal{L}\left(f_{\theta}\left({G}\oplus \delta\right), y\right), \\
\end{equation}
where $\preceq$ represents the budget constraints relationship in graph, $\oplus$ denotes the application of perturbations of node features or topology structure on the original graph $G$.

\subsubsection{Point Clouds Data}

Point clouds Data represents 3D spatial information and is widely used in computer vision, robotics, and autonomous driving. ULD methods in this domain typically involve perturbations that interfere with shape recognition and geometric feature extraction, affecting the learnability of point-based representations.

UPC~\cite{wang2024unlearnable3DpointClouds} points out that as more and more 3D point cloud data contain sensitive information, the unauthorized use of this new type of data has also become a serious problem. To address this issue, UPC proposes the unlearnable framework for 3D point clouds including two processes: data protector and authorized user as shown in Equation (\ref{eq:protector}) and Equation (\ref{eq:user}) repectively. Protector involves a class-wise setting established by a category-adaptive allocation strategy and multi-transformations assigned to samples. Authorized user involves a restoration scheme that utilizes class-wise inverse matrix transformation, thus enabling authorized-only training for unlearnable data.
\begin{equation}
\label{eq:protector}
    \min_\theta \max_{t } \mathcal{L}\left(f_\theta\left(t(x) \right), y\right),
\end{equation}
\begin{equation}
\label{eq:user}
    \min_\theta \min_{\pi} \mathcal{L}\left(f_\theta\left(\pi(t(x)) \right), y\right),
\end{equation}
where $(x,y)$ is the raw point cloud data, $t$ is the 3D transformation matrix that does not seriously damage the visual quality of point clouds, $\pi=t^{-1}$ is the inversion of $t$ received from data protectors.

\subsubsection{Multimodal Data}

Multimodal data integrates multiple data types, such as images with textual descriptions or audio-visual content. ULD techniques for multimodal data must consider cross-modal interactions and disrupt learning in a manner that prevents models from effectively aligning and fusing different modalities.

MEM~\cite{liu2024multimodalUnlearnableMM} points out that hackers may use image-text data for model training without authorization, which may include personal and privacy sensitive information, but traditional ULD methods are designed for single-modal classification. This remains largely unexplored in Multimodal Contrastive Learning (MCL)~\cite{mustafa2022multimodal}. Therefore, MEM proposes multi-step error minimization, a new optimization process for generating multimodal unlearnable samples, which extends the error minimization framework and simultaneously optimizes image noise and additional text trigger words, thereby expanding the optimization space and effectively misleading the model to learn the shortcut between noise features and text trigger words.
\begin{equation}
    \min_{\theta} \min_{\delta,\eta}  \mathcal{L}(f_{\theta}(x_I\oplus\delta;x_T\oplus\eta)),
\end{equation}
where $f_\theta(\cdot;\cdot)$ is the pretrained CLIP~\cite{radford2021learning} model, $(x_I,x_T)$ is the image-text data, $\theta$ and $\eta$ are the image perturbation and text trigger respectively.




\section{Specific Attack Methods targeted ULD}
\label{sec:SpecificAttackMethods}

While defense-oriented ULD techniques are designed to render data unlearnable and hinder a model’s ability to extract useful features, a parallel line of research has emerged on attack methods aimed at countering these defenses. In the context of image classification, such attack strategies seek to recover learnability by neutralizing the effects of ULD perturbations. In this section, we categorize these attack methods into three broad groups based on the mechanisms they employ to invert or bypass the defensive perturbations:
\begin{enumerate}
    \item \textbf{Shortcut Removal/Recovery Approaches}: These methods focus on detecting and eliminating the spurious shortcuts or misleading patterns introduced by ULD defenses. By removing these artifacts, the approaches restore the model’s capacity to learn discriminative features.
    \item \textbf{Adversarial Counter-Optimization Approaches}: In these methods, the attack is formulated as a counter-adversarial optimization problem, in which the attacker designs perturbations or training strategies that directly oppose the ULD objective, thereby recovering the model’s performance.
    \item \textbf{Reconstruction/Detection-Based Approaches}: These strategies involve explicitly identifying the ULD perturbations—using reconstruction frameworks or detection algorithms—and then removing or mitigating them to restore the original data’s learnability.
\end{enumerate}
Together, these attack methods represent critical countermeasures in the ongoing arms race between ULD defenses and adversarial strategies. In the following sections, we provide a detailed analysis of the experimental evaluations and comparative performance of these attack methods. The following subsections provide a detailed discussion of each category.

\subsection{Shortcut Removal/Recovery Approaches}
\label{sec:ShortcutRemoval}

In this category, the attack methods focus on identifying and eliminating the spurious shortcuts induced by defensive ULD techniques. The underlying idea is that defensive perturbations often cause the model to latch onto irrelevant, non-generalizable patterns (shortcuts) that degrade the quality of learned features. By detecting and removing these shortcuts, the attack methods aim to recover the discriminative information that was suppressed. Representative approaches include methods such as Image Shortcut Squeezing (ISS)~\cite{liu2023ImageShortcutSqueezing}, UEraser~\cite{qin2023learningTheUnlearnable}, and JCDP~\cite{jiang2023unlearnableGiveAFalse}. These methods generally employ optimization techniques that reverse the effects of the defensive perturbations, thereby restoring the classifier's performance on clean data.

JDCP~\cite{jiang2023unlearnableGiveAFalse} points out that traditional ULD techniques provide a false sense of security because they do not prevent unauthorized users from exploiting otherwise unprotected data, removing protection by turning unlearnable data into learnable data again. Motivated by this observation, JDCP defines a new threat by introducing learnable unauthorized examples, which are unlearnable data protected by removal. The core of the JDCP approach mainly involves a novel purification process, implemented through a novel joint conditional diffusion model.
\begin{equation}
    \min_{\theta,w}   \mathcal{L}(f_\theta(G_w(x + \delta;y)), y),
\end{equation}
where $G_w(\cdot;y)$ is the DDPM~\cite{ho2020denoising} model parameterized with $y$ and conditioned with $y$.

ISS~\cite{liu2023ImageShortcutSqueezing} work has shown through extensive experiments that multiple ULD methods are susceptible to shortcut compression of images based on simple compression. In further investigation, ISS illustrates that the nature of the perturbation depends on the type of surrogate model used for toxicity generation, which explains why a particular ISS compression yields the best performance for a particular type of perturbation. Based on this, ISS was further tested for more adaptive poisoning and showed that it is not an ideal defense against ISS, providing a meaningful analysis during the subsequent development of ULD technology.

UEraser~\cite{qin2023learningTheUnlearnable} proposes a method designed to combat unlearnable example attacks - a data poisoning technique that adds subtle perturbations to images, preventing deep learning models from effectively learning from such data. Unlike traditional adversarial training, which is resource intensive and may degrade model accuracy, UEraser combines an effective data augmentation strategy with loss maximization adversarial augmentation to counteract the forgetting effect of these attacks. It goes beyond the regular p-norm perturbation constraints assumed by current forgetting attacks and defenses, thus improving the generalization ability of the model without compromising accuracy.

RSK~\cite{LearningFromCUDA} finds that simple transformations such as image sharpening and frequency filtering can significantly improve the utility of CUDA data for training, leading to substantial improvements in test accuracy over adversarial training on CIFAR-10, CIFAR-100, and ImageNet-100 datasets. Our study highlights the need to continuously improve data poisoning techniques to ensure data privacy and opens new avenues for enhancing robustness on unlearnable datasets.

Shortcut removal/recovery approaches target a common vulnerability in ULD defenses: the inadvertent introduction of spurious shortcuts that mislead a model’s feature extraction. In many ULD methods, the perturbations cause models to latch onto superficial, non-generalizable patterns rather than learning robust, discriminative features. Shortcut recovery techniques aim to detect and mitigate these misleading cues, thereby restoring the model's capacity to learn meaningful representations.

Formally, let \( D_{\text{ULD}} \) denote a dataset rendered unlearnable by a defense mechanism, and let \( f_{\theta} \) be a classifier trained on \( D_{\text{ULD}} \). The goal of a shortcut recovery method is to find a transformation \( T: \mathcal{X} \rightarrow \mathcal{X} \) that recovers useful features by eliminating the spurious shortcuts. This can be formulated as:
\begin{equation}
    T^* = \arg \min_{T \in \mathcal{T}} \; \mathbb{E}_{(x,y) \sim D_{\text{ULD}}} \left[ \| \phi(T(x)) - \phi(x) \|^2 \right],
    \label{eq:shortcut_recovery}
\end{equation}
where \( \phi(\cdot) \) represents a feature extraction function (e.g., the output of an intermediate layer), and \( \mathcal{T} \) is a set of candidate transformations that preserve the semantic content of \( x \).

These approaches underscore the ongoing arms race between ULD defenses and attack methods, revealing that even robustly designed unlearnable data may be vulnerable to strategies specifically aimed at removing or neutralizing the induced shortcuts.

\subsection{Adversarial Counter Optimization Approaches}
\label{sec:AdversarialCounter}

Adversarial counter-optimization approaches formulate the recovery process as a min-max optimization problem. Instead of passively removing the perturbations, these methods actively optimize a counter-adversarial objective that directly opposes the ULD defense. For example, AVATAR~\cite{dolatabadi2024devilAdvocate} design objectives that maximize the model’s ability to extract discriminative features despite the presence of ULD-induced perturbations. Similarly, NLT4UD~\cite{hapuarachchi2024nonlinearTransformationsAgainst} adjust the optimization dynamics to neutralize the defensive noise. These methods often rely on ensemble or game-theoretic formulations to enhance the transferability and robustness of the recovery process.

AVATAR~\cite{dolatabadi2024devilAdvocate} critically reviews recent ULD techniques (called availability attacks in the original article), challenging the notion that data can be made permanently unavailable by minor perturbations. Targeting the ULD technique, AVATAR utilizes diffusion models to efficiently denoise such perturbed data, thereby restoring its utility for neural network training, and provides a rigorous analysis demonstrating that the required denoising effort is directly related to the size of the initial data perturbation. This work highlights the need for ongoing research into robust data protection methods.

Challenging the notion that multiple representative ULD methods can make data permanently unlearnable, NLT4UD~\cite{hapuarachchi2024nonlinearTransformationsAgainst} introduces a nonlinear transformation framework designed to combat such data protection techniques. By applying specific nonlinear transformations, our framework enables DNNs to efficiently learn from datasets previously considered unlearnable. NLT4UD provides a rigorous analysis that proves that this approach significantly improves the ability to bypass existing data protection mechanisms. This work highlights the need to develop more robust data protection strategies to prevent unauthorized use of data in machine learning models.

ST~\cite{dang2023flew} observed in the study that the model initially learns the perturbation and semantic features simultaneously, but quickly overfits the perturbation, especially at shallow layers. ST proposes to solve this problem by gradually adjusting the learning rate based on Activation Cluster Measurement (ACM), which evaluates the overfitting state of the model. This method effectively prevents overfitting on perturbed features. It enables the model to learn effective semantic information from unlearnable samples.

OProj~\cite{sandoval2023WhatCanWeLearn} finds that although these perturbations in the ULD method make it difficult for the deep neural network to generalize, the network still learns useful features that can be reweighted to achieve high test performance. In addition, OProj proposes a orthogonal projection attack that can effectively recover learnability from existing unlearnable datasets. In view of the fact that this research mainly explores the attack methods against unlearnable data sets, especially through the orthogonal projection technique to recover the learnability of the data.

Adversarial counter-optimization approaches aim to neutralize the effects of ULD defenses by formulating a counter optimization problem that seeks to recover the learnability of the perturbed data. In contrast to defense-oriented methods—which design perturbations to hinder feature extraction—these attack strategies actively optimize an opposing objective to restore discriminative feature learning, often via a min-max formulation.

Let \( D' = \{(x_i + \delta_i, y_i)\}_{i=1}^N \) be the unlearnable dataset generated by a ULD defense. The goal of an adversarial counter-optimization method is to find a recovery transformation \( T: \mathcal{X} \rightarrow \mathcal{X} \) or an additional recovery perturbation \( \Delta \) such that the recovered dataset
\[
\hat{D} = \{(T(x_i + \delta_i + \Delta), y_i)\}_{i=1}^N
\]
enables a model \( f_\theta \) to regain its ability to learn meaningful features. One representative formulation is:
\begin{equation}
    \Delta^* = \arg \min_{\Delta \in \mathcal{C}} \; \mathbb{E}_{(x,y) \sim D'} \Big[ \mathcal{L}\Big(f_\theta\big(T(x + \delta + \Delta)\big), y\Big) \Big],
    \label{eq:adv_counter_optimization}
\end{equation}
subject to \( \|\Delta\|_p \leq \eta \), where \( \eta \) is a small constant controlling the magnitude of the recovery perturbation, \( \mathcal{L} \) is the standard classification loss, and \( \mathcal{C} \) is the feasible set for \( \Delta \).

These adversarial counter-optimization approaches demonstrate that, despite the protective measures enforced by ULD defenses, the unlearnability can be partially or even fully reversed under adaptive attack conditions. This highlights an ongoing arms race between defensive ULD techniques and methods designed to recover learnability, underscoring the importance of developing more robust data protection strategies.

\subsection{Reconstruction/Detection-Based Approaches}
\label{sec:ReconstructionDetection}

Reconstruction and detection-based approaches focus on explicitly identifying the presence of ULD perturbations and subsequently removing or corrupting them to restore the data’s learnability. Techniques such as DVAE~\cite{liu2024gameUnlearnable} employ variational autoencoder frameworks to reconstruct clean representations from perturbed inputs. Meanwhile, methods like UDP~\cite{DetectionAndDefense} and COIN~\cite{li2023detectingCorruptingCUDA} are designed to detect ULD patterns and apply corrective transformations. By filtering out or reversing the perturbations, these approaches enable the model to recover its original performance, even in the presence of adversarial defenses.

DVAE~\cite{liu2024gameUnlearnable} introduces a novel pretraining purification method to counteract unlearnable samples that degrade model performance through subtle data modifications. They observe that rate-constrained variational autoencoders (vae) inherently suppress perturbations in unlearnable data and provide a theoretical analysis of this phenomenon. Building on these insights, DVAE proposes untangled variational autoencoders to disentangle perturbations with learnable class-level embeddings. This leads to a two-stage purification approach: initially removing the interference and subsequently producing precise, non-toxic data that ensures effectiveness and robustness in a variety of situations.

UDP~\cite{DetectionAndDefense} demonstrates that existing unlearnable data can be efficiently identified using simple network-based detection methods, providing theoretical results for the linear separability of certain unlearnable data sets. Building on these findings, the authors propose a novel defense strategy that combines strong data augmentation with adversarial noise generated by simple networks. This method aims to reduce the detectability of unlearnable data, so as to enhance the resilience of deep learning models to such data poisoning techniques. UDP also establishes a quantitative criterion between unlearnable data and adversarial budgets, providing insights into the conditions under which robust UEs may exist or adversarial defenses may fail.

COIN~\cite{li2023detectingCorruptingCUDA} proposes a mechanism to corrupt such unlearnable data using pixel-based image transformations, thereby restoring the generalization ability of models trained on such data. In addition, COIN introduces two new convolution-based forms of unlearnable, namely horizontal Unlearnable Data Augmentation (HUDA) and vertical unlearnable Data Augmentation (VUDA), to further evaluate the effectiveness of its defense strategies. This work highlights the need to develop powerful methods to detect and neutralize advanced data poisoning techniques that compromise the integrity of machine learning models.

Reconstruction/Detection-Based Approaches aim to explicitly identify and reverse the perturbations introduced by ULD defenses. Rather than counteracting ULD through re-optimization on the perturbed data, these methods focus on recovering the underlying clean representations or directly detecting and mitigating the perturbations. Typically, such approaches employ autoencoder or variational autoencoder (VAE) architectures to learn a mapping \( R: \mathcal{X} \rightarrow \mathcal{X} \) that reconstructs the original input \( x \) from its perturbed version \( \tilde{x} = x + \delta \). This reconstruction objective can be formulated as:
\begin{equation}
    R^* = \arg \min_{R} \; \mathbb{E}_{x \sim \mathcal{D}} \left[ \|R(x + \delta) - x\|^2 \right],
    \label{eq:reconstruction_loss}
\end{equation}
where the goal is to minimize the reconstruction error while maintaining the inherent structure of \( x \).

Alternatively, detection-based approaches design a classifier \( D: \mathcal{X} \rightarrow \{0,1\} \) to distinguish between clean and perturbed samples. The detection process is typically optimized via a binary loss:
\begin{equation}
    \min_{D} \; \mathbb{E}_{(x,\tilde{x}) \sim \mathcal{D}'} \left[ \ell\left(D(\tilde{x}), 1\right) + \ell\left(D(x), 0\right) \right],
    \label{eq:detection_loss}
\end{equation}
where \( \ell(\cdot,\cdot) \) is a standard binary cross-entropy loss, and labels 1 and 0 indicate the presence or absence of ULD perturbations, respectively.

Together, these reconstruction/detection approaches offer an alternative avenue in the arms race against ULD defenses by focusing on the explicit recovery or removal of perturbations, thereby restoring the model's ability to learn meaningful representations.

\section{Evaluation and Comparison}
\label{sec:Evaluation}

In this section, we provide a comprehensive evaluation framework for Unlearnable Data (ULD) techniques, along with a comparative analysis of existing methods. Evaluating ULD methods is challenging due to the need to balance multiple objectives: degrading the learnability of data while preserving perceptual quality, ensuring robustness against adaptive training, and maintaining computational efficiency. We summarize key evaluation metrics, describe common experimental protocols, and compare representative approaches across these dimensions.

\subsection{Evaluation Metrics}
The effectiveness of ULD methods is typically measured by several key metrics.
\paragraph{\textbf{Unlearnability}} This is quantified by the degradation in model performance when trained on perturbed data. Formally, if a model trained on clean data achieves accuracy $\text{Acc}(f_{\theta^*}, D)$ and the same model trained on the unlearnable dataset $D'$ achieves $\text{Acc}(f_{\theta^*}, D_{\text{test}})$, then unlearnability can be measured by the relative drop:
\begin{equation}
\Delta \text{Acc} = \text{Acc}(f_{\theta^*}, D) - \text{Acc}(f_{\theta^*}, D_{\text{test}}).
\label{eq:unlearnability_metric}
\end{equation}
\paragraph{\textbf{Imperceptibility}} The perturbations must remain imperceptible to humans. This is generally ensured by constraining the perturbation norm, e.g., 
\begin{equation}
\|\delta(x)\|_p \leq \epsilon, \quad \forall x \in D.
\label{eq:imperceptibility_constraint_eval}
\end{equation}
Additional perceptual metrics (e.g., SSIM for images) are often used to validate that the modified data appears similar to the original.
\paragraph{\textbf{Robustness}} Robustness measures the persistence of unlearnability when the model is subjected to adaptive training techniques, such as adversarial training or data augmentation. Methods that maintain performance degradation under these conditions are considered more robust.
\paragraph{\textbf{Transferability}} This metric evaluates whether perturbations generated for one model are effective against other architectures. High transferability indicates that the ULD method generalizes well in black-box settings.
\paragraph{\textbf{Computational Efficiency}} The time and resources required for generating ULD are critical for practical deployment, especially for large-scale datasets. Efficiency is measured in terms of the computational cost of the perturbation generation process.

\subsection{Experimental Protocols}
Evaluation of ULD methods is typically performed on standard benchmarks across different modalities (e.g., CIFAR-10, CIFAR-100~\cite{krizhevsky2009learning}, ImageNet-100~\cite{deng2009imagenet} for images) with the following steps:
\begin{itemize}
    \item Train a baseline model on the clean dataset \(D\) and record performance metrics.
    \item Generate the unlearnable dataset \(D'\) using a specific ULD method.
    \item Train the same model architecture on \(D'\) and evaluate its performance on a clean test set \(D_{\text{test}}\).
    \item Compare the performance drop, measure imperceptibility using norm constraints and perceptual metrics, and assess robustness through adversarial or augmented training scenarios.
\end{itemize}

Recent advancements in ULD evaluation have been significantly enhanced by the introduction of APBench~\cite{qin2024apbench}—a unified benchmark for availability poisoning attacks and defenses. APBench standardizes experimental setups, providing a comprehensive suite of poisoning attacks, defense algorithms, and data augmentation techniques. It enables consistent and reproducible evaluations across different models and datasets. Key features of APBench include the following points.

\begin{itemize}
    \item \textbf{Comprehensive Suite:} Incorporates 9 supervised and 2 unsupervised poisoning attack methods, 8 defense strategies, and 4 common data augmentation methods.
    \item \textbf{Standardized Protocols:} Ensures fair and reproducible comparative evaluations by implementing poisoning attacks and defense mechanisms under standardized perturbations and training hyperparameters.
    \item \textbf{Extensive Evaluations:} Conducts experiments across multiple datasets, examining scenarios such as partial poisoning, increased perturbations, and the transferability of attacks across different DNN models under various defenses.
    \item \textbf{Analytical Tools:} Provides visual evaluation tools like t-SNE, Shapley value maps, and Grad-CAM to qualitatively analyze the impact of poisoning attacks.
\end{itemize}

Integrating APBench into ULD research aligns with the experimental protocols outlined above, offering standardized methodologies and evaluation metrics that enhance the reliability and comparability of research findings in the field of data poisoning and protection.

\subsection{Comparative Analysis}
Different methods show the tradeoffs and dependencies of ULD technology in multiple dimensions. We reveal some important trends in ULD technology through comparative analysis.
\begin{itemize}
    \item \textbf{Trade-Off Between Unlearnability and Imperceptibility:} Methods such as EM, REM, and TUE achieve high unlearnability by significantly degrading model performance; however, they must carefully control perturbation magnitudes to avoid perceptible distortions, as enforced by constraints like Equation~\eqref{eq:imperceptibility_constraint_eval}.
    \item \textbf{Impact of Supervision and Surrogate Dependency:} Supervised ULD techniques tend to generate more targeted perturbations, while surrogate-based methods typically achieve higher effectiveness in white-box settings. Unsupervised and surrogate-free approaches, though more generally applicable, often exhibit a lower degree of performance degradation.
    \item \textbf{Robustness and Adaptability:} Recent advancements have focused on enhancing the robustness of ULD methods against adaptive training defenses. Methods that integrate dynamic or hybrid perturbation strategies tend to show improved resistance to adversarial countermeasures.
    \item \textbf{Computational Considerations:} Iterative optimization methods, while effective in generating unlearnable data, may incur significant computational overhead. This trade-off is critical for scalability in real-world applications.
\end{itemize}

In summary, the evaluation of ULD techniques highlights the inherent trade-offs between achieving high unlearnability, maintaining imperceptibility, ensuring robustness, and achieving computational efficiency. Although defense-oriented ULD methods have shown promise in protecting data against unauthorized learning, there remains a significant gap in balancing these competing objectives. The following sections on Applications, Limitations, and Future Research Directions further elaborate on these challenges and outline potential avenues for advancing ULD research.

\section{Applications of Unlearnable Data}
\label{sec:Applications}

Unlearnable Data (ULD) techniques have emerged as a promising solution for safeguarding sensitive information and protecting data assets against unauthorized exploitation. This section surveys the diverse applications of ULD across multiple domains, illustrating how these techniques are leveraged to enhance data privacy, secure intellectual property, and prevent model theft, among other uses.

\subsection{Data Privacy and Intellectual Property Protection}
One of the primary motivations for ULD is to protect personal data and proprietary datasets. By rendering data unlearnable to unauthorized models, ULD techniques prevent malicious actors from effectively extracting useful information. In practice, ULD is applied to publicly released datasets to ensure that even if the data is scraped or leaked, any models trained on such data exhibit significantly degraded performance. This defensive strategy is particularly relevant in light of strict privacy regulations (e.g., GDPR~\cite{regulation2018generalGDPR}, CCPA~\cite{bonta2022californiaCCPA}) and the rising importance of data ownership in industries such as healthcare, finance, and autonomous systems.

\subsection{Prevention of Unauthorized Use}
ULD methods serve as a robust countermeasure to model theft, where adversaries attempt to train competitive models using proprietary data without proper authorization. By injecting carefully crafted perturbations into the training data, ULD techniques ensure that any model trained on this data fails to achieve acceptable performance. This not only preserves the commercial value of the dataset but also deters competitors from benefiting from unauthorized data usage. Such applications are especially critical in environments where large-scale, high-quality datasets constitute a significant competitive advantage.

\subsection{Enhancing Adversarial Robustness}
Beyond data privacy, ULD techniques contribute to improving adversarial robustness by preventing models from overfitting to spurious correlations. In adversarial settings, ULD can be deployed as a defensive mechanism to obstruct the learning process, thereby reducing the risk of adversarial attacks that exploit vulnerable features. By degrading the model’s ability to learn useful representations, ULD methods force adversaries to contend with models that are less sensitive to subtle perturbations—a quality that is beneficial in high-stakes applications such as security and surveillance.

\subsection{Domain-Specific Applications}
The versatility of ULD extends across various data modalities and application domains:
\begin{itemize}
    \item \textbf{Image Data:} ULD methods have been widely applied in computer vision, particularly for image classification, generation, and segmentation. For instance, in medical imaging, techniques like those in ~\cite{sun2024medicalUnlearnable} have been tailored to protect sensitive patient data while preserving image interpretability for diagnostic purposes.
    \item \textbf{Text Data:} In natural language processing, ULD techniques are used to prevent unauthorized training of language models on proprietary or sensitive text corpora. Methods such as those described in ~\cite{li2023makeTextUnlearnable} ensure that published datasets do not inadvertently enable the extraction of private information.
    \item \textbf{Audio and Speech:} In audio applications, ULD is applied to protect voice data and other auditory signals, which is crucial for biometric authentication and speaker verification systems. Studies like ~\cite{zhang2024hiddenspeaker} exemplify the application of ULD in this domain.
    \item \textbf{Multimodal and Time-Series Data:} With the expansion of ULD research, techniques have also been adapted for complex, multimodal datasets and time-series data, addressing challenges in fields such as autonomous driving, finance, and sensor networks.
\end{itemize}

The application of ULD techniques across these varied domains highlights their potential to transform data protection strategies in machine learning. While the primary focus has been on defense, the dual-use nature of ULD also underscores the need for careful ethical and regulatory considerations. As ULD research matures, further integration with real-world systems—along with rigorous evaluation and standardization—will be critical for broad adoption. Overall, ULD represents a versatile toolset for mitigating risks associated with unauthorized data usage, enhancing adversarial robustness, and securing sensitive information in a data-driven world.

\section{Challenges and Limitations}
\label{sec:Challenges}

Despite the promising potential of Unlearnable Data (ULD) techniques for protecting data and mitigating unauthorized model training, several challenges and limitations remain, which hinder their widespread adoption and practical deployment. In this section, we discuss these key issues:

\subsection{Trade-off Between Imperceptibility and Unlearnability}
A core challenge in ULD methods is balancing the perturbation strength with perceptual quality. Perturbations must be sufficiently strong to degrade model performance yet remain imperceptible to human observers. This trade-off is formalized by norm constraints (e.g., \( \| \delta(x) \|_p \leq \epsilon \)), which often limit the effectiveness of ULD under robust training scenarios. As defense methods become more sophisticated, achieving an optimal balance remains a significant technical hurdle.

\subsection{Robustness Against Adaptive Training}
Many ULD techniques, particularly those that rely on direct optimization of perturbations, are vulnerable to adaptive training strategies such as adversarial training or data augmentation. Such methods can partially mitigate the impact of ULD perturbations, enabling models to recover some of the suppressed features. Developing ULD methods that are robust to these adaptive defenses is an ongoing challenge in the field.

\subsection{Computational Complexity and Scalability}
Generating unlearnable data typically involves iterative optimization procedures, which can be computationally expensive—especially for large-scale datasets and complex model architectures. The high computational overhead not only limits the scalability of ULD techniques but also poses challenges for real-time or resource-constrained applications. Efficient algorithms and high-performance computing strategies are needed to bridge this gap.

\subsection{Generalizability Across Modalities and Tasks}
While many ULD methods have been developed for image data, extending these techniques to other modalities (e.g., text, audio, time series, and multimodal data) remains challenging. Each modality presents unique characteristics, and methods that work well for images may not directly translate to text or audio without significant modifications. Additionally, adapting ULD approaches to various tasks—such as classification, generation, and segmentation—requires careful consideration of task-specific constraints and evaluation metrics.

\subsection{Ethical and Dual-Use Concerns}
ULD techniques are inherently dual-use: while they can protect sensitive data, they may also be misused to obstruct legitimate learning or to facilitate anti-competitive practices. This raises ethical and regulatory questions about the deployment of ULD methods in practice. Establishing clear guidelines and frameworks to govern the use of ULD is essential to ensure that these technologies are used responsibly.

\subsection{Interpretability and Theoretical Understanding}
Although significant progress has been made in developing ULD techniques, the theoretical underpinnings of why certain perturbations render data unlearnable remain partially understood. Enhanced interpretability of ULD mechanisms is needed to gain deeper insights into their behavior, predict their performance under different conditions, and design more effective countermeasures against adaptive attacks.

In summary, while ULD represents a novel and promising approach to data protection in machine learning, addressing these challenges is crucial for improving their robustness, scalability, and general applicability. Future research must focus on developing more efficient, interpretable, and ethically sound ULD methods that can withstand adaptive adversarial strategies across a wide range of applications.

\section{Future Research Directions}
\label{sec:Future}

As the field of Unlearnable Data (ULD) continues to mature, several promising avenues for future research have emerged. In this section, we outline key directions that could drive the next generation of ULD techniques and expand their practical applicability, while ensuring that critical attributes like Transferability, Imperceptibility, Unlearnability, Scalability, Interpretability, Revocability, Stability, Adaptability, and Robustness are fully considered.

\begin{itemize}
    \item \textbf{Adaptive Perturbation Strategies:} Future work should explore methods that dynamically adjust the perturbation budget based on the complexity of the data and task. Developing adaptive algorithms that balance imperceptibility with effective unlearnability remains a critical challenge.
    
    \item \textbf{Robustness Against Adaptive Defenses:} As adversaries continually improve their countermeasures (e.g., adversarial training, data augmentation), ULD methods must be designed to withstand these adaptive defenses. This includes improving the stability of ULD techniques under different conditions, ensuring that the data remains unlearnable despite variations in the attack strategies employed by adversaries. Additionally, ULD methods should be designed with robustness to variations in data distribution and model architectures, ensuring consistent performance across tasks.
    
    \item \textbf{Scalability and Efficiency:} The computational cost of generating ULD—especially for large-scale datasets—poses a significant barrier to real-world deployment. Future research should focus on developing more efficient algorithms, potentially leveraging high-performance computing, model compression, or transfer learning to scale ULD generation. These methods should not only be scalable but also robust, ensuring that they maintain the desired unlearnability even when applied to vast and diverse datasets.
    
    \item \textbf{Generalizability Across Modalities and Tasks:} Although much of the current work has focused on image data, extending ULD techniques to other modalities (e.g., text, audio, time series, and multimodal data) is essential. Future studies should investigate modality-specific challenges and design unified frameworks that generalize across diverse tasks. This involves ensuring that ULD methods are adaptable to different types of data while maintaining their core properties, such as imperceptibility and unlearnability, across modalities.
    
    \item \textbf{Theoretical Insights and Interpretability:} A deeper theoretical understanding of why certain perturbations render data unlearnable is still lacking. Advancing the interpretability of ULD mechanisms—through rigorous analysis of feature and gradient behavior—can lead to more principled and effective designs. Future work should aim to unravel the underlying principles that govern data perturbations, ensuring that the processes are not only interpretable but also transferable to new domains and datasets.
    
    \item \textbf{Hybrid Approaches:} Combining multiple perturbation strategies (direct, feature-guided, gradient-guided) may yield more robust ULD methods. Future research should explore hybrid approaches that leverage the strengths of each method while mitigating their individual limitations. This will require ensuring that these hybrid strategies are both stable and adaptable, providing a robust defense across different adversarial conditions.
    
    \item \textbf{Revocability of Unlearnable Data:} A crucial area for future research is the potential for the revocation of unlearnable data once it is no longer needed for privacy protection or other purposes. Investigating mechanisms that allow for the reversal of ULD transformations or the unlearning of data could pave the way for more flexible data protection methods that allow users to retain control over their data throughout its lifecycle.
    
    \item \textbf{Ethical and Regulatory Considerations:} Given the dual-use nature of ULD, establishing ethical guidelines and regulatory frameworks is critical. Future work should address the potential for misuse, ensuring that ULD technologies are deployed in a manner that protects data privacy without enabling malicious applications. Furthermore, as ULD methods evolve, they must be designed with careful consideration of societal and ethical implications, ensuring that they are not only secure but also fair and transparent.
    
    \item \textbf{Standardized Evaluation Protocols:} Developing comprehensive benchmarks and standardized evaluation metrics for ULD will facilitate more consistent comparisons across methods. This includes assessing unlearnability, imperceptibility, robustness, scalability, and adaptability in a unified experimental framework. Ensuring that these evaluation metrics cover all essential attributes of ULD methods will provide the necessary foundation for future development and deployment.

\end{itemize}

By addressing these research directions, the ULD community can advance towards more robust, scalable, and interpretable methods that not only protect sensitive data but also integrate seamlessly into real-world machine learning systems.

\section{Conclusion}
\label{sec:Conclusion}

In this survey, we have provided a comprehensive review of Unlearnable Data (ULD) techniques as a distinct research area within machine learning security. We began by discussing the motivations behind ULD—primarily the need to protect sensitive data and intellectual property in an era dominated by data-driven models—and established the conceptual foundations that differentiate ULD from related fields such as adversarial attacks, data poisoning, and machine unlearning.

This survey systematically categorized ULD methods along multiple dimensions, including technical intention, data modality, task scenario, supervision and surrogate dependency, as well as boundedness constraints. We then delved into the methodologies underpinning ULD generation, with a detailed examination of strategies such as direct input perturbation, feature-guided perturbation, gradient-guided perturbation, and hybrid approaches. Additionally, we discussed specific attack methods targeting ULD defenses, highlighting the ongoing arms race between protection mechanisms and countermeasures.

The evaluation and comparative analysis further underscored the critical trade-offs between unlearnability, imperceptibility, robustness, transferability, and computational efficiency. Finally, we identified several promising future research directions that aim to enhance the adaptability, scalability, and interpretability of ULD techniques, while also addressing emerging ethical and regulatory challenges.

Overall, the evolving landscape of ULD offers powerful tools for mitigating unauthorized model training and safeguarding data integrity. As machine learning continues to integrate into critical applications across various domains, further advancements in ULD will be essential for building secure and resilient AI systems. We hope this survey serves as a valuable resource and roadmap for researchers and practitioners striving to advance the state-of-the-art in unlearnable data generation.


\begin{table*}[ht]
\caption{Overview of ULD Methodology.}
\label{tab:Methodology}
\resizebox{\linewidth}{!}{
\small
\begin{tabular}{@{} >{\centering\arraybackslash}m{0.19\linewidth} >{\centering\arraybackslash}m{0.08\linewidth} >{\centering\arraybackslash}m{0.05\linewidth} p{0.88\linewidth} @{}}
\toprule
\multicolumn{1}{c}{\textbf{Study}} & \textbf{Publication} & \textbf{Year} & \multicolumn{1}{c}{\textbf{Overview}} \\ \midrule
EM   [\href{https://arxiv.org/pdf/2101.04898}{\textcolor{blue}{Paper}},   \href{https://github.com/HanxunH/Unlearnable-Examples}{\textcolor{magenta}{Code}}]                                        & ICLR                 & 2021          & Propose error-minimizing noise to keep data visually unchanged while   making the trained model behave nearly randomly, protecting personal data.                                                                                                                                                                                                                                                                             \\
GrayAugs   [\href{https://arxiv.org/pdf/2111.13244}{\textcolor{blue}{Paper}},   \href{https://github.com/liuzrcc/ULE-GrayAug}{\textcolor{magenta}{Code}}]                                           & arXiv                & 2021          & Use grayscale and data augmentation to analyze the impact of noise on   unlearnable data and validate the effectiveness of the preliminary method.                                                                                                                                                                                                                                                                            \\
REM   [\href{https://arxiv.org/pdf/2203.14533}{\textcolor{blue}{Paper}},   \href{https://github.com/fshp971/robust-unlearnable-examples}{\textcolor{magenta}{Code}}]                                & ICLR                 & 2022          & Propose robust error-minimizing noise to enhance the protection of data   under adversarial training.                                                                                                                                                                                                                                                                                                                         \\
UC   [\href{https://arxiv.org/pdf/2301.01217}{\textcolor{blue}{Paper}},   \href{https://github.com/jiamingzhang94/Unlearnable-Clusters}{\textcolor{magenta}{Code}}]                                 & CVPR                 & 2022          & Use clustering methods to generate perturbations independent of labels,   reducing category dependence and enhancing flexibility.                                                                                                                                                                                                                                                                                             \\
LLock   [\href{https://arxiv.org/pdf/2202.03576}{\textcolor{blue}{Paper}},   \href{https://github.com/jinghuichen/Learnability-Lock}{\textcolor{magenta}{Code}}]                                    & ICLR                 & 2022          & Propose adversarial reversible transformations, which can be considered   as image-to-image mappings to encrypt data samples. Authorized clients can   use specific keys to unlock the learnability of protected datasets and train   models normally.                                                                                                                                                                        \\
TUE   [\href{https://arxiv.org/pdf/2210.10114}{\textcolor{blue}{Paper}},   \href{https://github.com/renjie3/TUE}{\textcolor{magenta}{Code}}]                                                        & ICLR                 & 2023          & Transferable unlearnable perturbations across models and datasets.                                                                                                                                                                                                                                                                                                                                                            \\
OPS   [\href{https://arxiv.org/pdf/2205.12141}{\textcolor{blue}{Paper}},   \href{https://github.com/cychomatica/One-Pixel-Shotcut}{\textcolor{magenta}{Code}}]                                      & ICLR                 & 2023          & Propose that perturbing a single pixel can produce significant effects,   revealing DNN's preference for local perturbations during training.                                                                                                                                                                                                                                                                                 \\
CUDA   [\href{https://arxiv.org/pdf/2303.04278}{\textcolor{blue}{Paper}},   \href{https://github.com/vinusankars/Convolution-based-Unlearnability}{\textcolor{magenta}{Code}}]                      & CVPR                 & 2023          & Embed category-specific perturbations in the frequency domain using   convolution kernels to address the slow iteration issue in traditional   unlearnable methods.                                                                                                                                                                                                                                                           \\
SEM   [\href{https://arxiv.org/pdf/2311.13091}{\textcolor{blue}{Paper}},   \href{https://github.com/liuyixin-louis/Stable-Unlearnable-Example}{\textcolor{magenta}{Code}}]                          & AAAI                 & 2023          & Propose a stable version of error-minimizing noise to ensure unlearnable   data remains effective under different training conditions.                                                                                                                                                                                                                                                                                        \\
Segue   [\href{https://arxiv.org/pdf/2310.16061v1}{\textcolor{blue}{Paper}}]                                                                                                                        & arXiv                & 2023          & Use auxiliary information to guide the generation of unlearnable samples   for facial privacy protection.                                                                                                                                                                                                                                                                                                                     \\
UT   [\href{https://arxiv.org/pdf/2307.00456}{\textcolor{blue}{Paper}}]                                                                                                                             & ACLW                 & 2023          & Generate unlearnable text using gradient-based search techniques.                                                                                                                                                                                                                                                                                                                                                             \\
EMinS   [\href{https://arxiv.org/pdf/2303.02568}{\textcolor{blue}{Paper}}]                                                                                                                          & NDSS                 & 2023          & This paper proposes a new method for generating unlearnable graph   examples.                                                                                                                                                                                                                                                                                                                                                 \\
JCDP   [\href{https://arxiv.org/pdf/2305.09241}{\textcolor{blue}{Paper}},   \href{https://github.com/jiangw-0/LE_JCDP}{\textcolor{magenta}{Code}}]                                                  & MM                   & 2023          & Use diffusion models to restore unlearnable data to a learnable state,   revealing security risks in current unlearnable methods.                                                                                                                                                                                                                                                                                             \\
ISS   [\href{https://arxiv.org/pdf/2301.13838}{\textcolor{blue}{Paper}},   \href{https://github.com/liuzrcc/ImageShortcutSqueezing}{\textcolor{magenta}{Code}}]                                     & ICML                 & 2023          & Use image compression techniques to weaken the shortcuts formed by perturbations, enabling the model to recapture the original   semantic information.                                                                                                                                                                                                                                                                    \\
UEraser   [\href{https://arxiv.org/pdf/2303.15127}{\textcolor{blue}{Paper}},   \href{https://github.com/lafeat/ueraser}{\textcolor{magenta}{Code}}]                                                 & arXiv                & 2023          & Suppress the impact of unlearnable perturbations through adversarial data   augmentation and restore data learnability.                                                                                                                                                                                                                                                                                                       \\
AVATAR   [\href{https://arxiv.org/pdf/2303.08500}{\textcolor{blue}{Paper}},   \href{https://github.com/hmdolatabadi/AVATAR}{\textcolor{magenta}{Code}}]                                             & SatML                & 2023          & Recover learnability from unlearnable data based on diffusion models,   breaking through unlearnability protection.                                                                                                                                                                                                                                                                                                           \\
ST   [\href{https://arxiv.org/pdf/2306.02064}{\textcolor{blue}{Paper}},   \href{https://github.com/CherryBlueberry/ST}{\textcolor{magenta}{Code}}]                                                  & arXiv                & 2023          & Use a staged training strategy to prevent the model from falling into   perturbation feature traps too early, restoring learning of data semantics.                                                                                                                                                                                                                                                                           \\
OProj   [\href{https://arxiv.org/pdf/2305.19254}{\textcolor{blue}{Paper}},   \href{https://github.com/psandovalsegura/learn-from-unlearnable}{\textcolor{magenta}{Code}}]                           & NIPS                 & 2023          & Provide a deep theoretical and experimental analysis of the nature and   limitations of unlearnable data, proposing Orthogonal Projection to restore   learnability.                                                                                                                                                                                                                                                          \\
UEEG   [\href{https://arxiv.org/pdf/2412.09854}{\textcolor{blue}{Paper}}]                                                                                                                           & TNNLS                & 2023          & Proposes two methods to convert raw EEG data into identity-unlearnable EEG data, removing user identity information while maintaining good performance for brain-computer interface (BCI) tasks. \\
EntF   [\href{https://openreview.net/pdf?id=zKvm1ETDOq}{\textcolor{blue}{Paper}},   \href{https://github.com/WenRuiUSTC/EntF}{\textcolor{magenta}{Code}}]                                           & ICLR                 & 2023          & Propose a poisoning method based on indifferentiable features to   significantly reduce the impact of adversarial training.                                                                                                                                                                                                                                                                                                   \\
ASR   [\href{https://arxiv.org/pdf/2305.10691}{\textcolor{blue}{Paper}},   \href{https://github.com/EuterpeK/Rethinking-Data-Availability-Attacks}{\textcolor{magenta}{Code}}]                      & CVPR                 & 2024          & Discuss the limitations of existing unlearnable methods from the   perspective of data availability attacks, proposing new ideas for data   protection.                                                                                                                                                                                                                                                                       \\
PUE   [\href{https://arxiv.org/pdf/2405.03316}{\textcolor{blue}{Paper}},   \href{https://github.com/NeuralSec/certified-data-learnability}{\textcolor{magenta}{Code}}]                              & NDSS                 & 2024          & This paper proposes a theoretical mechanism to evaluate and verify the   learnability of unlearnable datasets through parameter smoothing.                                                                                                                                                                                                                                                                                    \\
SecVec   [\href{https://arxiv.org/pdf/2311.02105}{\textcolor{blue}{Paper}}]                                                                                                                         & ACLF                 & 2024          & Propose a controllable training framework that leverages the concept of   safe vectors to make harmful behaviors unlearnable during fine-tuning.                                                                                                                                                                                                                                                                              \\
UE4TS   [\href{https://arxiv.org/pdf/2402.02028}{\textcolor{blue}{Paper}}]                                                                                                                          & PAKDD                & 2024          & Propose a method for generating unlearnable examples to protect   time-series data from unauthorized training by deep learning models.                                                                                                                                                                                                                                                                                        \\
SALM   [\href{https://arxiv.org/pdf/2403.10573}{\textcolor{blue}{Paper}}]                                                                                                                           & ICMLW                & 2024          & Propose a sparse-aware local masking method for medical images,   selectively perturbing important pixel regions to generate unlearnable data.                                                                                                                                                                                                                                                                                \\
EUDP   [\href{https://arxiv.org/pdf/2306.01902}{\textcolor{blue}{Paper}}]                                                                                                                           & ICLRW                & 2024          & Unlearnable data for diffusion models, protecting unauthorized data   generation.                                                                                                                                                                                                                                                                                                                                             \\

MEM   [\href{https://arxiv.org/pdf/2407.16307}{\textcolor{blue}{Paper}},   \href{https://github.com/thinwayliu/Multimodal-Unlearnable-Examples}{\textcolor{magenta}{Code}}]                         & MM                   & 2024          & This paper proposes multi-step error minimization, a new optimization   process for generating multimodal unlearnable samples.                                                                                                                                                                                                                                                                                                \\
DH   [\href{https://arxiv.org/pdf/2406.17349}{\textcolor{blue}{Paper}}]                                                                                                                             & TIFS                 & 2024          & Traditional unlearnable perturbations targeting low-level features are   easily affected by common data augmentation strategies. This paper proposes   an adaptive approach to hide semantic images with rich high-level features,   making them more robust to adversarial measures.                                                                                                                                         \\
HiddenSpeaker   [\href{https://arxiv.org/pdf/2405.15655v2}{\textcolor{blue}{Paper}}]                                                                                                                & IJCNN                & 2024          & This paper embeds imperceptible perturbations into training speech   samples, making them unlearnable for deep learning-based speaker verification   systems.                                                                                                                                                                                                                                                                 \\
PosCUDA   [\href{https://arxiv.org/pdf/2401.02135}{\textcolor{blue}{Paper}}]                                                                                                                        & arXiv                & 2024          & Propose a location-based class-level convolution to create unlearnable   audio datasets.                                                                                                                                                                                                                                                                                                                                      \\
GUE   [\href{https://arxiv.org/pdf/2401.17523}{\textcolor{blue}{Paper}},   \href{https://github.com/hong-xian/gue}{\textcolor{magenta}{Code}}]                                                      & AAAI                 & 2024          & Model the unlearnable data generation process from a game-theoretic   perspective, solving for equilibrium to generate optimal perturbations that   break protection.                                                                                                                                                                                                                                                         \\
DVAE   [\href{https://arxiv.org/pdf/2405.01460}{\textcolor{blue}{Paper}},   \href{https://github.com/yuyi-sd/D-VAE}{\textcolor{magenta}{Code}}]                                                     & ICML                 & 2024          & Use constrained VAE pretraining purification to remove perturbations and   restore the learnability of unlearnable data.                                                                                                                                                                                                                                                                                                      \\
RSK   [\href{https://arxiv.org/pdf/2411.01742}{\textcolor{blue}{Paper}},   \href{https://github.com/aseriesof-tubes/RSK}{\textcolor{magenta}{Code}}]                                                & NIPSW                & 2024          & Propose a method to restore learnability in CUDA samples through   sharpening and DCT frequency filtering.                                                                                                                                                                                                                                                                                                                    \\
UDP   [\href{https://arxiv.org/pdf/2312.08898}{\textcolor{blue}{Paper}},   \href{https://github.com/hala64/udp}{\textcolor{magenta}{Code}}]                                                         & AAAI                 & 2024          & Use network-based detection methods to identify unlearnable examples.                                                                                                                                                                                                                                                                                                                                                         \\
NLT4UD   [\href{https://arxiv.org/pdf/2406.02883}{\textcolor{blue}{Paper}}]                                                                                                                         & arXiv                & 2024          & Propose an effective nonlinear transformation framework that enables   effective learning from traditionally unlearnable data.                                                                                                                                                                                                                                                                                                \\
UGE   [\href{https://arxiv.org/pdf/2404.14016}{\textcolor{blue}{Paper}}]                                                                                                                            & CVPR                 & 2024          & Extend the concept of unlearnable data to conditional data learnability,   showcasing learnability for authorized users while maintaining unlearnability   for potential hackers.                                                                                                                                                                                                                                             \\
UPC   [\href{https://arxiv.org/pdf/2410.03644}{\textcolor{blue}{Paper}},   \href{https://github.com/CGCL-codes/UnlearnablePC}{\textcolor{magenta}{Code}}]                                           & NIPS                 & 2024          & Propose the first overall unlearnable framework for 3D point clouds.                                                                                                                                                                                                                                                                                                                                                          \\
14A   [\href{https://openreview.net/pdf?id=vSerUPYFtB}{\textcolor{blue}{Paper}},   \href{https://github.com/xiye7lai/UnlearnableConcept}{\textcolor{magenta}{Code}}]                                & ICML                 & 2024          & Propose a universal perturbation generator using conceptually unlearnable   data.                                                                                                                                                                                                                                                                                                                                             \\
MetaCloak   [\href{https://arxiv.org/pdf/2311.13127}{\textcolor{blue}{Paper}},   \href{https://github.com/liuyixin-louis/MetaCloak}{\textcolor{magenta}{Code}}]                                     & CVPR                 & 2024          & Propose a meta-learning framework to solve the bi-level optimization   suboptimal problem of error-minim                                                                                                                                                                                                                                                                                                                      \\
AUEAPP   [\href{https://arxiv.org/pdf/2402.04010}{\textcolor{blue}{Paper}},   \href{https://github.com/EhanW/AUE-AAP}{\textcolor{magenta}{Code}}]                                                   & NPIS                 & 2024          & Propose achieving both supervised and contrastive unlearnability   simultaneously.                                                                                                                                                                                                                                                                                                                                            \\
APBench   [\href{https://arxiv.org/pdf/2308.03258}{\textcolor{blue}{Paper}},   \href{https://github.com/lafeat/apbench}{\textcolor{magenta}{Code}}]                                                 & TMLR                 & 2024          & Propose the first benchmark for usability poisoning attacks and defenses.                                                                                                                                                                                                                                                                                                                                                     \\
UMed   [\href{https://arxiv.org/pdf/2403.14250}{\textcolor{blue}{Paper}}]                                                                                                                           & arXiv                & 2024          & Propose a method for generating unlearnable medical images, incorporating   prior knowledge of the data and protecting images by introducing contour and   texture perturbations.                                                                                                                                                                                                                                             \\
InMark   [\href{https://openaccess.thecvf.com/content/CVPR2024/papers/Liu_Countering_Personalized_Text-to-Image_Generation_with_Influence_Watermarks_CVPR_2024_paper.pdf}{\textcolor{blue}{Paper}}] & CVPR                 & 2024          & Propose unlearnable examples for diffusion models by embedding watermarks   in influential pixels.                                                                                                                                                                                                                                                                                                                            \\
POP   [\href{https://arxiv.org/pdf/2410.20742}{\textcolor{blue}{Paper}},   \href{https://github.com/wxzyd123/Pivotal_Objective_Perturbation}{\textcolor{magenta}{Code}}]                            & CCSW                 & 2024          & Apply imperceptible error-minimizing noise to raw speech samples to   prevent them from being effectively learned for text-to-speech synthesis   models, thus preventing the generation of high-quality deepfake speech.                                                                                                                                                                                                      \\
ExpShield   [\href{https://arxiv.org/pdf/2412.21123}{\textcolor{blue}{Paper}}]                                                                                                                      & arXiv                & 2024          & Restrict data abuse during LLM training without affecting readability.                                                                                                                                                                                                                                                                                                                                                        \\
IRP   [\href{https://dr.ntu.edu.sg/bitstream/10356/179131/3/IRP_Main_ECCV_2024.pdf}{\textcolor{blue}{Paper}},   \href{https://github.com/lyumingzhi/IRP}{\textcolor{magenta}{Code}}]                & ECCV                 & 2024          & Based on CUDA theoretical analysis, propose imperfect recovery poisoning   aimed at achieving strong poisoning effects while maintaining high image   quality.                                                                                                                                                                                                                                                                \\
ARMOR   [\href{https://arxiv.org/pdf/2501.08862}{\textcolor{blue}{Paper}}]                                                                                                                          & arXiv                & 2025          & Propose using data augmentation strategies to disrupt the detectability   of perturbations in unlearnable data, enhancing protective effects.                                                                                                                                                                                                                                                                                 \\
HPC4UE   [\href{https://arxiv.org/pdf/2501.06080}{\textcolor{blue}{Paper}},   \href{https://github.com/hrlblab/UE_HPC}{\textcolor{magenta}{Code}}]                                                  & arXiv                & 2025          & Study the feasibility of unlearnable data in high-performance computing,   exploring the impact of batch size on data unlearnability.                                                                                                                                                                                                                                                                                         \\
UnSeg   [\href{https://arxiv.org/pdf/2410.09909}{\textcolor{blue}{Paper}},   \href{https://github.com/sunye23/UnSeg}{\textcolor{magenta}{Code}}]                                                    & NIPS                 & 2025          & For segmentation tasks, propose a new unlearnable framework to train a   universal unlearnable noise generator that can convert any downstream image   into an unlearnable version for segmentation tasks.                                                                                                                                                                                                                    \\
HarmonyCloak   [\href{https://mosis.eecs.utk.edu/publications/meerza2024harmonycloak.pdf}{\textcolor{blue}{Paper}}]                                                                                 & S\&P                 & 2025          & Propose a defense mechanism using generative AI models to prevent   exploitative use of artwork, particularly in instrumental contexts.                                                                                                                                                                                                                                                                                       \\
COIN   [\href{https://arxiv.org/pdf/2311.18403}{\textcolor{blue}{Paper}},   \href{https://github.com/wxldragon/COIN}{\textcolor{magenta}{Code}}]                                                    & AAAI                 & 2025          & Propose a method to detect CUDA perturbations and break their protective   effects through reverse engineering.                                                                                                                                                                                                                                                                                                               \\
SALUD   [\href{https://openreview.net/pdf?id=I4Lq2RJ0eJ}{\textcolor{blue}{Paper}},   \href{https://github.com/MLsecurityLab/HowFarAreFromTrueUnlearnability.git}{\textcolor{magenta}{Code}}]        & ICLR                 & 2025          & Propose unlearnability distance, based on the distribution of parameters   in clean and poisoned models, to measure data unlearnability, aiming to   promote community awareness of the capability boundaries of existing   unlearnable methods.                                                                                                                                                                              \\ \bottomrule
\end{tabular}}
\end{table*}

\bibliographystyle{IEEEtran}
\bibliography{mybibfile}

@inproceedings{deng2009imagenet,
  title={Imagenet: A large-scale hierarchical image database},
  author={Deng, Jia and Dong, Wei and Socher, Richard and Li, Li-Jia and Li, Kai and Fei-Fei, Li},
  booktitle={2009 IEEE conference on computer vision and pattern recognition},
  pages={248--255},
  year={2009},
  organization={Ieee}
}

@inproceedings{lin2014microsoft,
  title={Microsoft coco: Common objects in context},
  author={Lin, Tsung-Yi and Maire, Michael and Belongie, Serge and Hays, James and Perona, Pietro and Ramanan, Deva and Doll{\'a}r, Piotr and Zitnick, C Lawrence},
  booktitle={Computer vision--ECCV 2014: 13th European conference, zurich, Switzerland, September 6-12, 2014, proceedings, part v 13},
  pages={740--755},
  year={2014},
  organization={Springer}
}

@inproceedings{karras2019style,
  title={A style-based generator architecture for generative adversarial networks},
  author={Karras, Tero and Laine, Samuli and Aila, Timo},
  booktitle={Proceedings of the IEEE/CVF conference on computer vision and pattern recognition},
  pages={4401--4410},
  year={2019}
}

@Article{Everingham10,
   author = "Everingham, M. and Van~Gool, L. and Williams, C. K. I. and Winn, J. and Zisserman, A.",
   title = "The Pascal Visual Object Classes (VOC) Challenge",
   journal = "International Journal of Computer Vision",
   volume = "88",
   year = "2010",
   number = "2",
   month = jun,
   pages = "303--338",
}

@article{achiam2023gpt,
  title={Gpt-4 technical report},
  author={Achiam, Josh and Adler, Steven and Agarwal, Sandhini and Ahmad, Lama and Akkaya, Ilge and Aleman, Florencia Leoni and Almeida, Diogo and Altenschmidt, Janko and Altman, Sam and Anadkat, Shyamal and others},
  journal={arXiv preprint arXiv:2303.08774},
  year={2023}
}

@inproceedings{radford2021learning,
  title={Learning transferable visual models from natural language supervision},
  author={Radford, Alec and Kim, Jong Wook and Hallacy, Chris and Ramesh, Aditya and Goh, Gabriel and Agarwal, Sandhini and Sastry, Girish and Askell, Amanda and Mishkin, Pamela and Clark, Jack and others},
  booktitle={International conference on machine learning},
  pages={8748--8763},
  year={2021},
  organization={PMLR}
}

@inproceedings{ramesh2021zero,
  title={Zero-shot text-to-image generation},
  author={Ramesh, Aditya and Pavlov, Mikhail and Goh, Gabriel and Gray, Scott and Voss, Chelsea and Radford, Alec and Chen, Mark and Sutskever, Ilya},
  booktitle={International conference on machine learning},
  pages={8821--8831},
  year={2021},
  organization={PMLR}
}

@incollection{hill2022secretive,
  title={The secretive company that might end privacy as we know it},
  author={Hill, Kashmir},
  booktitle={Ethics of Data and Analytics},
  pages={170--177},
  year={2022},
  publisher={Auerbach Publications}
}

@inproceedings{somepalli2023diffusion,
  title={Diffusion art or digital forgery? investigating data replication in diffusion models},
  author={Somepalli, Gowthami and Singla, Vasu and Goldblum, Micah and Geiping, Jonas and Goldstein, Tom},
  booktitle={Proceedings of the IEEE/CVF conference on computer vision and pattern recognition},
  pages={6048--6058},
  year={2023}
}

@inproceedings{birhane2021large,
  title={Large image datasets: A pyrrhic win for computer vision?},
  author={Birhane, Abeba and Prabhu, Vinay Uday},
  booktitle={2021 IEEE Winter Conference on Applications of Computer Vision (WACV)},
  pages={1536--1546},
  year={2021},
  organization={IEEE}
}

@inproceedings{bourtoule2021machine,
  title={Machine unlearning},
  author={Bourtoule, Lucas and Chandrasekaran, Varun and Choquette-Choo, Christopher A and Jia, Hengrui and Travers, Adelin and Zhang, Baiwu and Lie, David and Papernot, Nicolas},
  booktitle={2021 IEEE symposium on security and privacy (SP)},
  pages={141--159},
  year={2021},
  organization={IEEE}
}

@article{goodfellow2014explaining,
  title={Explaining and harnessing adversarial examples},
  author={Goodfellow, Ian J and Shlens, Jonathon and Szegedy, Christian},
  journal={arXiv preprint arXiv:1412.6572},
  year={2014}
}

@inproceedings{PGD,
  title={Towards Deep Learning Models Resistant to Adversarial Attacks},
  author={Madry, Aleksander and Makelov, Aleksandar and Schmidt, Ludwig and Tsipras, Dimitris and Vladu, Adrian},
  booktitle={International Conference on Learning Representations},
  year={2018}
}

@inproceedings{EM,
  title={Unlearnable Examples: Making Personal Data Unexploitable},
  author={Huang, Hanxun and Ma, Xingjun and Erfani, Sarah Monazam and Bailey, James and Wang, Yisen},
  booktitle={International Conference on Learning Representations},
  year={2021}

}

@inproceedings{REM,
  title={Robust Unlearnable Examples: Protecting Data Privacy Against Adversarial Learning},
  author={Fu, Shaopeng and He, Fengxiang and Liu, Yang and Shen, Li and Tao, Dacheng},
  booktitle={International Conference on Learning Representations},
  year={2022}

}

@inproceedings{OPS,
  title={One-Pixel Shortcut: On the Learning Preference of Deep Neural Networks},
  author={Wu, Shutong and Chen, Sizhe and Xie, Cihang and Huang, Xiaolin},
  booktitle={The Eleventh International Conference on Learning Representations},
  year={2023}
}

@article{AT,
  title={Evaluating and understanding the robustness of adversarial logit pairing},
  author={Engstrom, Logan and Ilyas, Andrew and Athalye, Anish},
  journal={arXiv preprint arXiv:1807.10272},
  year={2018}
}

@inproceedings{CUDA,
  title={Cuda: Convolution-based unlearnable datasets},
  author={Sadasivan, Vinu Sankar and Soltanolkotabi, Mahdi and Feizi, Soheil},
  booktitle={Proceedings of the IEEE/CVF Conference on Computer Vision and Pattern Recognition},
  pages={3862--3871},
  year={2023}
}

@article{GoingGrayscale,
  title={Going grayscale: The road to understanding and improving unlearnable examples},
  author={Liu, Zhuoran and Zhao, Zhengyu and Kolmus, Alex and Berns, Tijn and van Laarhoven, Twan and Heskes, Tom and Larson, Martha},
  journal={arXiv preprint arXiv:2111.13244},
  year={2021}
}

@inproceedings{UnlearnableClusters,
  title={Unlearnable clusters: Towards label-agnostic unlearnable examples},
  author={Zhang, Jiaming and Ma, Xingjun and Yi, Qi and Sang, Jitao and Jiang, Yu-Gang and Wang, Yaowei and Xu, Changsheng},
  booktitle={Proceedings of the IEEE/CVF Conference on Computer Vision and Pattern Recognition},
  pages={3984--3993},
  year={2023}
}

@inproceedings{TransferableUEs,
  title={Transferable Unlearnable Examples},
  author={Ren, Jie and Xu, Han and Wan, Yuxuan and Ma, Xingjun and Sun, Lichao and Tang, Jiliang},
  booktitle={The Eleventh International Conference on Learning Representations},
  year={2023}
}

@inproceedings{ye2024ungeneralizable,
  title={Ungeneralizable examples},
  author={Ye, Jingwen and Wang, Xinchao},
  booktitle={Proceedings of the IEEE/CVF Conference on Computer Vision and Pattern Recognition},
  pages={11944--11953},
  year={2024}
}

@article{wang2024provably,
  title={Provably unlearnable data examples},
  author={Wang, Derui and Xue, Minhui and Li, Bo and Camtepe, Seyit and Zhu, Liming},
  journal={arXiv preprint arXiv:2405.03316},
  year={2024}
}

@inproceedings{DetectionAndDefense,
  title={Detection and defense of unlearnable examples},
  author={Zhu, Yifan and Yu, Lijia and Gao, Xiao-Shan},
  booktitle={Proceedings of the AAAI Conference on Artificial Intelligence},
  volume={38},
  number={15},
  pages={17211--17219},
  year={2024}
}

@inproceedings{liu2024StableUnlearnable,
  title={Stable unlearnable example: Enhancing the robustness of unlearnable examples via stable error-minimizing noise},
  author={Liu, Yixin and Xu, Kaidi and Chen, Xun and Sun, Lichao},
  booktitle={Proceedings of the AAAI Conference on Artificial Intelligence},
  volume={38},
  number={4},
  pages={3783--3791},
  year={2024}
}

@inproceedings{UnSeg,
  title={UnSeg: One Universal Unlearnable Example Generator is Enough against All Image Segmentation},
  author={Sun, Ye and Zhang, Hao and Zhang, Tiehua and Ma, Xingjun and Jiang, Yu-Gang},
  booktitle={The Thirty-eighth Annual Conference on Neural Information Processing Systems},
  year={2025}
}

@inproceedings{jiang2024UnlearnableTimeSeries,
  title={Unlearnable Examples for Time Series},
  author={Jiang, Yujing and Ma, Xingjun and Erfani, Sarah Monazam and Bailey, James},
  booktitle={Pacific-Asia Conference on Knowledge Discovery and Data Mining},
  pages={213--225},
  year={2024},
  organization={Springer}
}

@inproceedings{chen2024oneFroAll,
  title={One for all: A universal generator for concept unlearnability via multi-modal alignment},
  author={Chen, Chaochao and Zhang, Jiaming and Li, Yuyuan and Han, Zhongxuan},
  booktitle={Forty-first International Conference on Machine Learning},
  year={2024}
}

@article{wang2024unlearnable3DpointClouds,
  title={Unlearnable 3D point clouds: Class-wise transformation is all you need},
  author={Wang, Xianlong and Li, Minghui and Liu, Wei and Zhang, Hangtao and Hu, Shengshan and Zhang, Yechao and Zhou, Ziqi and Jin, Hai},
  journal={Advances in Neural Information Processing Systems},
  volume={37},
  pages={99404--99432},
  year={2024}
}

@inproceedings{liu2024multimodalUnlearnableMM,
  title={Multimodal unlearnable examples: Protecting data against multimodal contrastive learning},
  author={Liu, Xinwei and Jia, Xiaojun and Xun, Yuan and Liang, Siyuan and Cao, Xiaochun},
  booktitle={Proceedings of the 32nd ACM International Conference on Multimedia},
  pages={8024--8033},
  year={2024}
}

@article{meng2024semanticHiding,
  title={Semantic deep hiding for robust unlearnable examples},
  author={Meng, Ruohan and Yi, Chenyu and Yu, Yi and Yang, Siyuan and Shen, Bingquan and Kot, Alex C},
  journal={IEEE Transactions on Information Forensics and Security},
  year={2024},
  publisher={IEEE}
}

@inproceedings{liu2024gameUnlearnable,
  title={Game-theoretic unlearnable example generator},
  author={Liu, Shuang and Wang, Yihan and Gao, Xiao-Shan},
  booktitle={Proceedings of the AAAI Conference on Artificial Intelligence},
  volume={38},
  number={19},
  pages={21349--21358},
  year={2024}
}

@inproceedings{sun2024medicalUnlearnable,
  title={Medical Unlearnable Examples: Securing Medical Data from Unauthorized Training via Sparsity-Aware Local Masking},
  author={Sun, Weixiang and Liu, Yixin and Yan, Zhiling and Xu, Kaidi and Sun, Lichao},
  booktitle={ICML 2024 Next Generation of AI Safety Workshop},
  year={2024}
}

@inproceedings{yu2024purify,
  title={Purify Unlearnable Examples via Rate-Constrained Variational Autoencoders},
  author={Yu, Yi and Wang, Yufei and Xia, Song and Yang, Wenhan and Lu, Shijian and Tan, Yap-Peng and Kot, Alex},
  booktitle={International Conference on Machine Learning},
  pages={57678--57702},
  year={2024},
  organization={PMLR}
}

@inproceedings{liu2024metacloak,
  title={Metacloak: Preventing unauthorized subject-driven text-to-image diffusion-based synthesis via meta-learning},
  author={Liu, Yixin and Fan, Chenrui and Dai, Yutong and Chen, Xun and Zhou, Pan and Sun, Lichao},
  booktitle={Proceedings of the IEEE/CVF Conference on Computer Vision and Pattern Recognition},
  pages={24219--24228},
  year={2024}
}

@article{gong2025armor,
  title={ARMOR: Shielding Unlearnable Examples against Data Augmentation},
  author={Gong, Xueluan and Wang, Yuji and Chen, Yanjiao and Dong, Haocheng and Li, Yiming and Sun, Mengyuan and Li, Shuaike and Wang, Qian and Chen, Chen},
  journal={arXiv preprint arXiv:2501.08862},
  year={2025}
}

@inproceedings{wang2024efficientAvailabilityAttacks,
  title={Efficient availability attacks against supervised and contrastive learning simultaneously},
  author={Wang, Yihan and Zhu, Yifan and Gao, Xiao-Shan},
  booktitle={The Thirty-eighth Annual Conference on Neural Information Processing Systems},
  year={2024}
}

@article{qin2024apbench,
    title={{APB}ench: A Unified Availability Poisoning Attack and Defenses Benchmark},
    author={Tianrui Qin and Xitong Gao and Juanjuan Zhao and Kejiang Ye and Cheng-zhong Xu},
    journal={Transactions on Machine Learning Research},
    issn={2835-8856},
    year={2024},
    url={https://openreview.net/forum?id=igJ2XPNYbJ},
}

@inproceedings{fang2024rethinking,
  title={Re-thinking data availability attacks against deep neural networks},
  author={Fang, Bin and Li, Bo and Wu, Shuang and Ding, Shouhong and Yi, Ran and Ma, Lizhuang},
  booktitle={Proceedings of the IEEE/CVF Conference on Computer Vision and Pattern Recognition},
  pages={12215--12224},
  year={2024}
}

@inproceedings{peng2022learnabilityLOCK,
  title={LEARNABILITY LOCK: AUTHORIZED LEARNABILITY CONTROL THROUGH ADVERSARIAL INVERTIBLE TRANSFORMATIONS},
  author={Peng, Weiqi and Chen, Jinghui},
  booktitle={10th International Conference on Learning Representations, ICLR 2022},
  year={2022}
}

@inproceedings{zhang2024hiddenspeaker,
  title={Hiddenspeaker: Generate imperceptible unlearnable audios for speaker verification system},
  author={Zhang, Zhisheng and Huang, Pengyang},
  booktitle={2024 International Joint Conference on Neural Networks (IJCNN)},
  pages={1--8},
  year={2024},
  organization={IEEE}
}

@inproceedings{meerza2024harmonycloak,
  title={Harmonycloak: Making music unlearnable for generative ai},
  author={Meerza, Syed Irfan Ali and Liu, Jian and Sun, Lichao},
  booktitle={2025 IEEE Symposium on Security and Privacy (SP)},
  pages={85--85},
  year={2024},
  organization={IEEE Computer Society}
}

@article{dang2023flew,
  title={Flew Over Learning Trap: Learn Unlearnable Samples by Progressive Staged Training},
  author={Dang, Pucheng and Hu, Xing and Xu, Kaidi and Duan, Jinhao and Huang, Di and Han, Husheng and Zhang, Rui and Du, Zidong and Guo, Qi and Chen, Yunji},
  journal={arXiv preprint arXiv:2306.02064},
  year={2023}
}

@article{lin2024safeguarding,
  title={Safeguarding medical image segmentation datasets against unauthorized training via contour-and texture-aware perturbations},
  author={Lin, Xun and Yu, Yi and Xia, Song and Jiang, Jue and Wang, Haoran and Yu, Zitong and Liu, Yizhong and Fu, Ying and Wang, Shuai and Tang, Wenzhong and others},
  journal={arXiv preprint arXiv:2403.14250},
  year={2024}
}

@inproceedings{liu2024counteringInMark,
  title={Countering personalized text-to-image generation with influence watermarks},
  author={Liu, Hanwen and Sun, Zhicheng and Mu, Yadong},
  booktitle={Proceedings of the IEEE/CVF Conference on Computer Vision and Pattern Recognition},
  pages={12257--12267},
  year={2024}
}

@inproceedings{liu2023ImageShortcutSqueezing,
  title={Image shortcut squeezing: Countering perturbative availability poisons with compression},
  author={Liu, Zhuoran and Zhao, Zhengyu and Larson, Martha},
  booktitle={International conference on machine learning},
  pages={22473--22487},
  year={2023},
  organization={PMLR}
}

@inproceedings{zhou2024makingHarmfulBehaviorsUnlearnable,
  title={Making Harmful Behaviors Unlearnable for Large Language Models},
  author={Zhou, Xin and Lu, Yi and Ma, Ruotian and Wei, Yujian and Gui, Tao and Zhang, Qi and Huang, Xuan-Jing},
  booktitle={Findings of the Association for Computational Linguistics: ACL 2024},
  pages={10258--10273},
  year={2024}
}

@article{gokul2024poscuda,
  title={Poscuda: Position based convolution for unlearnable audio datasets},
  author={Gokul, Vignesh and Dubnov, Shlomo},
  journal={arXiv preprint arXiv:2401.02135},
  year={2024}
}

@article{meng2023user,
  title={User Identity Protection in EEG-Based Brain--Computer Interfaces},
  author={Meng, Lubin and Jiang, Xue and Huang, Jian and Li, Wei and Luo, Hanbin and Wu, Dongrui},
  journal={IEEE Transactions on Neural Systems and Rehabilitation Engineering},
  volume={31},
  pages={3576--3586},
  year={2023},
  publisher={IEEE}
}

@article{liu2024expshield,
  title={ExpShield: Safeguarding Web Text from Unauthorized Crawling and Language Modeling Exploitation},
  author={Liu, Ruixuan and Tran, Toan and Wang, Tianhao and Hu, Hongsheng and Wang, Shuo and Xiong, Li},
  journal={arXiv preprint arXiv:2412.21123},
  year={2024}
}

@inproceedings{huang2024leveraging,
  title={Leveraging imperfect restoration for data availability attack},
  author={Huang, Yi and Styborski, Jeremy and Lyu, Mingzhi and Wang, Fan and Kong, Adams},
  booktitle={European Conference on Computer Vision},
  pages={69--86},
  year={2024},
  organization={Springer}
}

@inproceedings{zhang2025segue,
  title={Segue: Side-information Guided Generative Unlearnable Examples for Facial Privacy Protection in Real World},
  author={Zhang, Zhiling and Zhang, Jie and Zhang, Kui and Zhou, Wenbo and Xu, Ting and Gao, Daiheng and Guo, Zixian and Guo, Qinglang and Zhang, Weiming and Yu, Nenghai},
  booktitle={ICASSP 2025-2025 IEEE International Conference on Acoustics, Speech and Signal Processing (ICASSP)},
  pages={1--5},
  year={2025},
  organization={IEEE}
}

@article{zhao2023unlearnableDiffusionModels,
  title={Unlearnable examples for diffusion models: Protect data from unauthorized exploitation},
  author={Zhao, Zhengyue and Duan, Jinhao and Hu, Xing and Xu, Kaidi and Wang, Chenan and Zhang, Rui and Du, Zidong and Guo, Qi and Chen, Yunji},
  journal={arXiv preprint arXiv:2306.01902},
  year={2023}
}

@inproceedings{li2023makeTextUnlearnable,
  title={Make Text Unlearnable: Exploiting Effective Patterns to Protect Personal Data},
  author={Li, Xinzhe and Liu, Ming},
  booktitle={Proceedings of the 3rd Workshop on Trustworthy Natural Language Processing (TrustNLP 2023)},
  pages={249--259},
  year={2023}
}

@article{liu2023unlearnableGraph,
  title={Unlearnable graph: Protecting graphs from unauthorized exploitation},
  author={Liu, Yixin and Fan, Chenrui and Zhou, Pan and Sun, Lichao},
  journal={arXiv preprint arXiv:2303.02568},
  year={2023}
}

@inproceedings{jiang2023unlearnableGiveAFalse,
  title={Unlearnable examples give a false sense of security: Piercing through unexploitable data with learnable examples},
  author={Jiang, Wan and Diao, Yunfeng and Wang, He and Sun, Jianxin and Wang, Meng and Hong, Richang},
  booktitle={Proceedings of the 31st ACM International Conference on Multimedia},
  pages={8910--8921},
  year={2023}
}

@article{qin2023learningTheUnlearnable,
  title={Learning the unlearnable: Adversarial augmentations suppress unlearnable example attacks},
  author={Qin, Tianrui and Gao, Xitong and Zhao, Juanjuan and Ye, Kejiang and Xu, Cheng-Zhong},
  journal={arXiv preprint arXiv:2303.15127},
  year={2023}
}

@inproceedings{dolatabadi2024devilAdvocate,
  title={The devil’s advocate: Shattering the illusion of unexploitable data using diffusion models},
  author={Dolatabadi, Hadi M and Erfani, Sarah and Leckie, Christopher},
  booktitle={2024 IEEE Conference on Secure and Trustworthy Machine Learning (SaTML)},
  pages={358--386},
  year={2024},
  organization={IEEE}
}

@article{sandoval2023WhatCanWeLearn,
  title={What can we learn from unlearnable datasets?},
  author={Sandoval-Segura, Pedro and Singla, Vasu and Geiping, Jonas and Goldblum, Micah and Goldstein, Tom},
  journal={Advances in Neural Information Processing Systems},
  volume={36},
  pages={75372--75391},
  year={2023}
}

@inproceedings{LearningFromCUDA,
  title={Learning From Convolution-based Unlearnable Datasets},
  author={Kim, Dohyun and Sandoval-Segura, Pedro},
  booktitle={The Third Workshop on New Frontiers in Adversarial Machine Learning},
  year={2024}

}

@article{hapuarachchi2024nonlinearTransformationsAgainst,
  title={Nonlinear Transformations Against Unlearnable Datasets},
  author={Hapuarachchi, Thushari and Lin, Jing and Xiong, Kaiqi and Rahouti, Mohamed and Ost, Gitte},
  journal={arXiv preprint arXiv:2406.02883},
  year={2024}
}

@article{zhu2025scaleupUnlearnableExamples,
  title={Scale-up Unlearnable Examples Learning with High-Performance Computing},
  author={Zhu, Yanfan and Lyngaas, Issac and Meena, Murali Gopalakrishnan and Koran, Mary Ellen I and Malin, Bradley and Moyer, Daniel and Bao, Shunxing and Kapadia, Anuj and Wang, Xiao and Landman, Bennett and others},
  journal={arXiv preprint arXiv:2501.06080},
  year={2025}
}

@article{li2023detectingCorruptingCUDA,
  title={Detecting and Corrupting Convolution-based Unlearnable Examples},
  author={Li, Minghui and Wang, Xianlong and Yu, Zhifei and Hu, Shengshan and Zhou, Ziqi and Zhang, Longling and Zhang, Leo Yu},
  journal={arXiv e-prints},
  pages={arXiv--2311},
  year={2023}
}

@inproceedings{HowFarAreWeFrom,
  title={How Far Are We from True Unlearnability?},
  author={Ye, Kai and Su, Liangcai and Qian, Chenxiong},
  booktitle={The Thirteenth International Conference on Learning Representations},
  year={2025}

}

@inproceedings{thudi2022necessity,
  title={On the necessity of auditable algorithmic definitions for machine unlearning},
  author={Thudi, Anvith and Jia, Hengrui and Shumailov, Ilia and Papernot, Nicolas},
  booktitle={31st USENIX security symposium (USENIX Security 22)},
  pages={4007--4022},
  year={2022}
}

@misc{logemann2018art,
  title={Art. 17 GDPR--Right to erasure (‘right to be forgotten’)--General Data Protection Regulation (GDPR)},
  author={Logemann, Thorsten},
  year={2018},
  publisher={Intersoft Consulting. https://gdpr-info. eu/art-17-gdpr/(accessed: 10.03. 2024)}
}

@article{gu2019badnets,
  title={Badnets: Evaluating backdooring attacks on deep neural networks},
  author={Gu, Tianyu and Liu, Kang and Dolan-Gavitt, Brendan and Garg, Siddharth},
  journal={IEEE Access},
  volume={7},
  pages={47230--47244},
  year={2019},
  publisher={IEEE}
}

@article{nguyen2022survey,
  title={A survey of machine unlearning},
  author={Nguyen, Thanh Tam and Huynh, Thanh Trung and Ren, Zhao and Nguyen, Phi Le and Liew, Alan Wee-Chung and Yin, Hongzhi and Nguyen, Quoc Viet Hung},
  journal={arXiv preprint arXiv:2209.02299},
  year={2022}
}

@article{liu2024machine,
  title={Machine unlearning in generative ai: A survey},
  author={Liu, Zheyuan and Dou, Guangyao and Tan, Zhaoxuan and Tian, Yijun and Jiang, Meng},
  journal={arXiv preprint arXiv:2407.20516},
  year={2024}
}

@article{wang2024machine,
  title={Machine unlearning: A comprehensive survey},
  author={Wang, Weiqi and Tian, Zhiyi and Zhang, Chenhan and Yu, Shui},
  journal={arXiv preprint arXiv:2405.07406},
  year={2024}
}

@article{liu2025threats,
  title={Threats, attacks, and defenses in machine unlearning: A survey},
  author={Liu, Ziyao and Ye, Huanyi and Chen, Chen and Zheng, Yongsen and Lam, Kwok-Yan},
  journal={IEEE Open Journal of the Computer Society},
  year={2025},
  publisher={IEEE}
}

@article{zhang2023review,
  title={A review on machine unlearning},
  author={Zhang, Haibo and Nakamura, Toru and Isohara, Takamasa and Sakurai, Kouichi},
  journal={SN Computer Science},
  volume={4},
  number={4},
  pages={337},
  year={2023},
  publisher={Springer}
}

@article{qu2023learn,
  title={Learn to unlearn: A survey on machine unlearning},
  author={Qu, Youyang and Yuan, Xin and Ding, Ming and Ni, Wei and Rakotoarivelo, Thierry and Smith, David},
  journal={arXiv preprint arXiv:2305.07512},
  year={2023}
}

@article{akhtar2018threat,
  title={Threat of adversarial attacks on deep learning in computer vision: A survey},
  author={Akhtar, Naveed and Mian, Ajmal},
  journal={Ieee Access},
  volume={6},
  pages={14410--14430},
  year={2018},
  publisher={IEEE}
}

@article{zhang2020adversarial,
  title={Adversarial attacks on deep-learning models in natural language processing: A survey},
  author={Zhang, Wei Emma and Sheng, Quan Z and Alhazmi, Ahoud and Li, Chenliang},
  journal={ACM Transactions on Intelligent Systems and Technology (TIST)},
  volume={11},
  number={3},
  pages={1--41},
  year={2020},
  publisher={ACM New York, NY, USA}
}

@article{chakraborty2021survey,
  title={A survey on adversarial attacks and defences},
  author={Chakraborty, Anirban and Alam, Manaar and Dey, Vishal and Chattopadhyay, Anupam and Mukhopadhyay, Debdeep},
  journal={CAAI Transactions on Intelligence Technology},
  volume={6},
  number={1},
  pages={25--45},
  year={2021},
  publisher={Wiley Online Library}
}

@article{akhtar2021advances,
  title={Advances in adversarial attacks and defenses in computer vision: A survey},
  author={Akhtar, Naveed and Mian, Ajmal and Kardan, Navid and Shah, Mubarak},
  journal={IEEE Access},
  volume={9},
  pages={155161--155196},
  year={2021},
  publisher={IEEE}
}

@article{li2022backdoor,
  title={Backdoor learning: A survey},
  author={Li, Yiming and Jiang, Yong and Li, Zhifeng and Xia, Shu-Tao},
  journal={IEEE transactions on neural networks and learning systems},
  volume={35},
  number={1},
  pages={5--22},
  year={2022},
  publisher={IEEE}
}

@article{gao2020backdoor,
  title={Backdoor attacks and countermeasures on deep learning: A comprehensive review},
  author={Gao, Yansong and Doan, Bao Gia and Zhang, Zhi and Ma, Siqi and Zhang, Jiliang and Fu, Anmin and Nepal, Surya and Kim, Hyoungshick},
  journal={arXiv preprint arXiv:2007.10760},
  year={2020}
}

@article{li2023backdoor,
  title={Backdoor attacks to deep learning models and countermeasures: A survey},
  author={Li, Yudong and Zhang, Shigeng and Wang, Weiping and Song, Hong},
  journal={IEEE Open Journal of the Computer Society},
  volume={4},
  pages={134--146},
  year={2023},
  publisher={IEEE}
}

@inproceedings{MitigatingUnauthorizedSpeechSynthesis,
    author = {Zhang, Zhisheng and Yang, Qianyi and Wang, Derui and Huang, Pengyang and Cao, Yuxin and Ye, Kai and Hao, Jie},
    title = {Mitigating Unauthorized Speech Synthesis for Voice Protection},
    year = {2024},
    isbn = {9798400712098},
    publisher = {Association for Computing Machinery},
    address = {New York, NY, USA},
    url = {https://doi.org/10.1145/3689217.3690615},
    doi = {10.1145/3689217.3690615},
    booktitle = {Proceedings of the 1st ACM Workshop on Large AI Systems and Models with Privacy and Safety Analysis},
    pages = {13–24},
    numpages = {12},
    location = {Salt Lake City, UT, USA},
    series = {LAMPS '24}
}

@inproceedings{
    wen2023is,
    title={Is Adversarial Training Really a Silver Bullet for Mitigating Data Poisoning?},
    author={Rui Wen and Zhengyu Zhao and Zhuoran Liu and Michael Backes and Tianhao Wang and Yang Zhang},
    booktitle={The Eleventh International Conference on Learning Representations },
    year={2023},
}

@inproceedings{jiang2023color,
  title={Color backdoor: A robust poisoning attack in color space},
  author={Jiang, Wenbo and Li, Hongwei and Xu, Guowen and Zhang, Tianwei},
  booktitle={Proceedings of the IEEE/CVF conference on computer vision and pattern recognition},
  pages={8133--8142},
  year={2023}
}

@article{li2022untargeted,
  title={Untargeted backdoor watermark: Towards harmless and stealthy dataset copyright protection},
  author={Li, Yiming and Bai, Yang and Jiang, Yong and Yang, Yong and Xia, Shu-Tao and Li, Bo},
  journal={Advances in Neural Information Processing Systems},
  volume={35},
  pages={13238--13250},
  year={2022}
}

@article{doan2021backdoor,
  title={Backdoor attack with imperceptible input and latent modification},
  author={Doan, Khoa and Lao, Yingjie and Li, Ping},
  journal={Advances in Neural Information Processing Systems},
  volume={34},
  pages={18944--18957},
  year={2021}
}

@article{jordan2015machine,
  title={Machine learning: Trends, perspectives, and prospects},
  author={Jordan, Michael I and Mitchell, Tom M},
  journal={Science},
  volume={349},
  number={6245},
  pages={255--260},
  year={2015},
  publisher={American Association for the Advancement of Science}
}

@article{kaplan2020scaling,
  title={Scaling laws for neural language models},
  author={Kaplan, Jared and McCandlish, Sam and Henighan, Tom and Brown, Tom B and Chess, Benjamin and Child, Rewon and Gray, Scott and Radford, Alec and Wu, Jeffrey and Amodei, Dario},
  journal={arXiv preprint arXiv:2001.08361},
  year={2020}
}

@article{villalobos2022will,
  title={Will we run out of data? an analysis of the limits of scaling datasets in machine learning},
  author={Villalobos, Pablo and Sevilla, Jaime and Heim, Lennart and Besiroglu, Tamay and Hobbhahn, Marius and Ho, Anson},
  journal={arXiv preprint arXiv:2211.04325},
  volume={1},
  year={2022},
  publisher={arXiv}
}

@article{jones2024ai,
  title={The AI revolution is running out of data. What can researchers do?},
  author={Jones, Nicola},
  journal={Nature},
  volume={636},
  number={8042},
  pages={290--292},
  year={2024},
  publisher={Nature}
}

@inproceedings{carlini2021extracting,
  title={Extracting training data from large language models},
  author={Carlini, Nicholas and Tramer, Florian and Wallace, Eric and Jagielski, Matthew and Herbert-Voss, Ariel and Lee, Katherine and Roberts, Adam and Brown, Tom and Song, Dawn and Erlingsson, Ulfar and others},
  booktitle={30th USENIX security symposium (USENIX Security 21)},
  pages={2633--2650},
  year={2021}
}

@inproceedings{shokri2017membership,
  title={Membership inference attacks against machine learning models},
  author={Shokri, Reza and Stronati, Marco and Song, Congzheng and Shmatikov, Vitaly},
  booktitle={2017 IEEE symposium on security and privacy (SP)},
  pages={3--18},
  year={2017},
  organization={IEEE}
}

@inproceedings{rasley2020deepspeed,
  title={Deepspeed: System optimizations enable training deep learning models with over 100 billion parameters},
  author={Rasley, Jeff and Rajbhandari, Samyam and Ruwase, Olatunji and He, Yuxiong},
  booktitle={Proceedings of the 26th ACM SIGKDD international conference on knowledge discovery \& data mining},
  pages={3505--3506},
  year={2020}
}

@article{brown2020language,
  title={Language models are few-shot learners},
  author={Brown, Tom and Mann, Benjamin and Ryder, Nick and Subbiah, Melanie and Kaplan, Jared D and Dhariwal, Prafulla and Neelakantan, Arvind and Shyam, Pranav and Sastry, Girish and Askell, Amanda and others},
  journal={Advances in neural information processing systems},
  volume={33},
  pages={1877--1901},
  year={2020}
}

@article{touvron2023llama,
  title={Llama: Open and efficient foundation language models},
  author={Touvron, Hugo and Lavril, Thibaut and Izacard, Gautier and Martinet, Xavier and Lachaux, Marie-Anne and Lacroix, Timoth{\'e}e and Rozi{\`e}re, Baptiste and Goyal, Naman and Hambro, Eric and Azhar, Faisal and others},
  journal={arXiv preprint arXiv:2302.13971},
  year={2023}
}

@article{guo2025deepseek,
  title={Deepseek-r1: Incentivizing reasoning capability in llms via reinforcement learning},
  author={Guo, Daya and Yang, Dejian and Zhang, Haowei and Song, Junxiao and Zhang, Ruoyu and Xu, Runxin and Zhu, Qihao and Ma, Shirong and Wang, Peiyi and Bi, Xiao and others},
  journal={arXiv preprint arXiv:2501.12948},
  year={2025}
}

@article{liu2024sora,
  title={Sora: A review on background, technology, limitations, and opportunities of large vision models},
  author={Liu, Yixin and Zhang, Kai and Li, Yuan and Yan, Zhiling and Gao, Chujie and Chen, Ruoxi and Yuan, Zhengqing and Huang, Yue and Sun, Hanchi and Gao, Jianfeng and others},
  journal={arXiv preprint arXiv:2402.17177},
  year={2024}
}

@article{garhart2023wasn,
  title={It wasn't me, it was the AI: Intellectual property and data privacy concerns with nonprofits' use of artificial intelligence systems},
  author={Garhart, Nate and Rowland, Cynthia},
  journal={Board \& Administrator for Administrators Only},
  volume={40},
  number={4},
  pages={1--2},
  year={2023},
  publisher={Wiley Online Library}
}

@article{oecd2025,
  author = {OECD},
  title = {Intellectual property issues in artificial intelligence trained on scraped data},
  journal = {OECD Artificial Intelligence Papers},
  number = {33},
  year = {2025},
  publisher = {OECD Publishing},
  address = {Paris},
  url = {https://doi.org/10.1787/d5241a23-en},
}

@article{picht2023ai,
  title={AI and IP: Theory to policy and back again--policy and research recommendations at the intersection of artificial intelligence and Intellectual Property},
  author={Picht, Peter Georg and Thouvenin, Florent},
  journal={IIC-International Review of Intellectual Property and Competition Law},
  volume={54},
  number={6},
  pages={916--940},
  year={2023},
  publisher={Springer}
}

@article{li2024digger,
  title={Digger: Detecting copyright content mis-usage in large language model training},
  author={Li, Haodong and Deng, Gelei and Liu, Yi and Wang, Kailong and Li, Yuekang and Zhang, Tianwei and Liu, Yang and Xu, Guoai and Xu, Guosheng and Wang, Haoyu},
  journal={arXiv preprint arXiv:2401.00676},
  year={2024}
}

@article{salem2018ml,
  title={Ml-leaks: Model and data independent membership inference attacks and defenses on machine learning models},
  author={Salem, Ahmed and Zhang, Yang and Humbert, Mathias and Berrang, Pascal and Fritz, Mario and Backes, Michael},
  journal={arXiv preprint arXiv:1806.01246},
  year={2018}
}

@article{martin2024artificial,
  title={Artificial intelligence and its implications for data privacy},
  author={Martin, Kelly D and Zimmermann, Johanna},
  journal={Current opinion in psychology},
  pages={101829},
  year={2024},
  publisher={Elsevier}
}

@article{regulation2018generalGDPR,
  title={General data protection regulation},
  author={Regulation, Protection},
  journal={Intouch},
  volume={25},
  pages={1--5},
  year={2018}
}

@article{bonta2022californiaCCPA,
  title={California consumer privacy act (CCPA)},
  author={Bonta, Rob},
  journal={Retrieved from State of California Department of Justice: https://oag. ca. gov/privacy/ccpa},
  year={2022}
}

@inproceedings{saravanan2010color,
  title={Color image to grayscale image conversion},
  author={Saravanan, Chandran},
  booktitle={2010 second international conference on computer engineering and applications},
  volume={2},
  pages={196--199},
  year={2010},
  organization={IEEE}
}

@inproceedings{hadsell2006dimensionality,
  title={Dimensionality reduction by learning an invariant mapping},
  author={Hadsell, Raia and Chopra, Sumit and LeCun, Yann},
  booktitle={2006 IEEE computer society conference on computer vision and pattern recognition (CVPR'06)},
  volume={2},
  pages={1735--1742},
  year={2006},
  organization={IEEE}
}

@article{liu2022bome,
  title={Bome! bilevel optimization made easy: A simple first-order approach},
  author={Liu, Bo and Ye, Mao and Wright, Stephen and Stone, Peter and Liu, Qiang},
  journal={Advances in neural information processing systems},
  volume={35},
  pages={17248--17262},
  year={2022}
}

@article{gong2021automatic,
  title={Automatic and harmless regularization with constrained and lexicographic optimization: A dynamic barrier approach},
  author={Gong, Chengyue and Liu, Xingchao and Liu, Qiang},
  journal={Advances in Neural Information Processing Systems},
  volume={34},
  pages={29630--29642},
  year={2021}
}

@inproceedings{wang2018non,
  title={Non-local neural networks},
  author={Wang, Xiaolong and Girshick, Ross and Gupta, Abhinav and He, Kaiming},
  booktitle={Proceedings of the IEEE conference on computer vision and pattern recognition},
  pages={7794--7803},
  year={2018}
}

@article{kullback1951information,
  title={On information and sufficiency},
  author={Kullback, Solomon and Leibler, Richard A},
  journal={The annals of mathematical statistics},
  volume={22},
  number={1},
  pages={79--86},
  year={1951},
  publisher={JSTOR}
}

@inproceedings{ruiz2023dreambooth,
  title={Dreambooth: Fine tuning text-to-image diffusion models for subject-driven generation},
  author={Ruiz, Nataniel and Li, Yuanzhen and Jampani, Varun and Pritch, Yael and Rubinstein, Michael and Aberman, Kfir},
  booktitle={Proceedings of the IEEE/CVF conference on computer vision and pattern recognition},
  pages={22500--22510},
  year={2023}
}

@inproceedings{kirillov2023segment,
  title={Segment anything},
  author={Kirillov, Alexander and Mintun, Eric and Ravi, Nikhila and Mao, Hanzi and Rolland, Chloe and Gustafson, Laura and Xiao, Tete and Whitehead, Spencer and Berg, Alexander C and Lo, Wan-Yen and others},
  booktitle={Proceedings of the IEEE/CVF international conference on computer vision},
  pages={4015--4026},
  year={2023}
}

@article{wolpaw2000brain,
  title={Brain-computer interface technology: a review of the first international meeting},
  author={Wolpaw, Jonathan R and Birbaumer, Niels and Heetderks, William J and McFarland, Dennis J and Peckham, P Hunter and Schalk, Gerwin and Donchin, Emanuel and Quatrano, Louis A and Robinson, Charles J and Vaughan, Theresa M and others},
  journal={IEEE transactions on rehabilitation engineering},
  volume={8},
  number={2},
  pages={164--173},
  year={2000}
}

@article{pham2000current,
  title={Current methods in medical image segmentation},
  author={Pham, Dzung L and Xu, Chenyang and Prince, Jerry L},
  journal={Annual review of biomedical engineering},
  volume={2},
  number={1},
  pages={315--337},
  year={2000},
  publisher={Annual Reviews 4139 El Camino Way, PO Box 10139, Palo Alto, CA 94303-0139, USA}
}

@article{tan2021survey,
  title={A survey on neural speech synthesis},
  author={Tan, Xu and Qin, Tao and Soong, Frank and Liu, Tie-Yan},
  journal={arXiv preprint arXiv:2106.15561},
  year={2021}
}

@article{yamagishi2021asvspoof,
  title={ASVspoof 2021: accelerating progress in spoofed and deepfake speech detection},
  author={Yamagishi, Junichi and Wang, Xin and Todisco, Massimiliano and Sahidullah, Md and Patino, Jose and Nautsch, Andreas and Liu, Xuechen and Lee, Kong Aik and Kinnunen, Tomi and Evans, Nicholas and others},
  journal={arXiv preprint arXiv:2109.00537},
  year={2021}
}

@article{bimbot2004tutorial,
  title={A tutorial on text-independent speaker verification},
  author={Bimbot, Fr{\'e}d{\'e}ric and Bonastre, Jean-Fran{\c{c}}ois and Fredouille, Corinne and Gravier, Guillaume and Magrin-Chagnolleau, Ivan and Meignier, Sylvain and Merlin, Teva and Ortega-Garc{\'\i}a, Javier and Petrovska-Delacr{\'e}taz, Dijana and Reynolds, Douglas A},
  journal={EURASIP Journal on Advances in Signal Processing},
  volume={2004},
  pages={1--22},
  year={2004},
  publisher={Springer}
}

@article{chowdhary2020natural,
  title={Natural language processing},
  author={Chowdhary, KR1442 and Chowdhary, KR},
  journal={Fundamentals of artificial intelligence},
  pages={603--649},
  year={2020},
  publisher={Springer}
}

@article{zhao2023survey,
  title={A survey of large language models},
  author={Zhao, Wayne Xin and Zhou, Kun and Li, Junyi and Tang, Tianyi and Wang, Xiaolei and Hou, Yupeng and Min, Yingqian and Zhang, Beichen and Zhang, Junjie and Dong, Zican and others},
  journal={arXiv preprint arXiv:2303.18223},
  volume={1},
  number={2},
  year={2023}
}

@inproceedings{ivison2023hint,
  title={HINT: Hypernetwork Instruction Tuning for Efficient Zero-and Few-Shot Generalisation},
  author={Ivison, Hamish and Bhagia, Akshita and Wang, Yizhong and Hajishirzi, Hannaneh and Peters, Matthew E},
  booktitle={Proceedings of the 61st Annual Meeting of the Association for Computational Linguistics (Volume 1: Long Papers)},
  pages={11272--11288},
  year={2023}
}

@article{scarselli2008graph,
  title={The graph neural network model},
  author={Scarselli, Franco and Gori, Marco and Tsoi, Ah Chung and Hagenbuchner, Markus and Monfardini, Gabriele},
  journal={IEEE transactions on neural networks},
  volume={20},
  number={1},
  pages={61--80},
  year={2008},
  publisher={IEEE}
}

@article{mustafa2022multimodal,
  title={Multimodal contrastive learning with limoe: the language-image mixture of experts},
  author={Mustafa, Basil and Riquelme, Carlos and Puigcerver, Joan and Jenatton, Rodolphe and Houlsby, Neil},
  journal={Advances in Neural Information Processing Systems},
  volume={35},
  pages={9564--9576},
  year={2022}
}

@article{ho2020denoising,
  title={Denoising diffusion probabilistic models},
  author={Ho, Jonathan and Jain, Ajay and Abbeel, Pieter},
  journal={Advances in neural information processing systems},
  volume={33},
  pages={6840--6851},
  year={2020}
}

@misc{krizhevsky2009learning,
  title={Learning multiple layers of features from tiny images},
  author={Krizhevsky, Alex and Hinton, Geoffrey and others},
  year={2009},
  publisher={Toronto, ON, Canada}
}

\vfill

\begin{IEEEbiography}[{\includegraphics[width=1in,height=1.25in,clip,keepaspectratio]{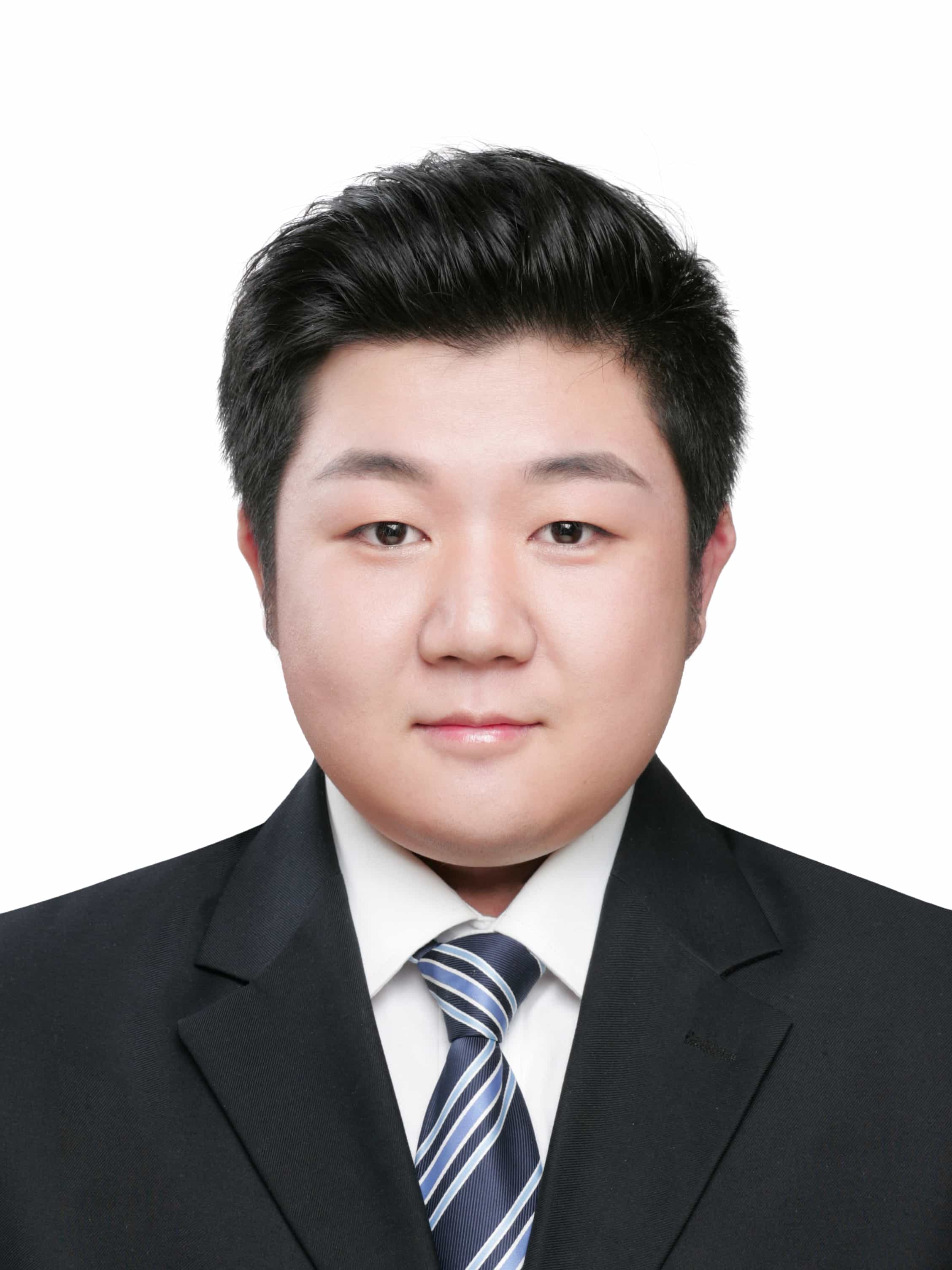}}]{Jiahao Li} received the B.S. degree in computational mathematics from Shandong University in 2020 and the Ph.D. degree in computer science from the Institute of Computing Technology (ICT), Chinese Academy of Sciences (CAS) in 2025. He is currently an assistant professor at the ICT, CAS. His current research focuses on AI safety, unlearnability, shortcut learning, anomaly detection, memory mechanism and quantum machine learning.
\end{IEEEbiography}

\begin{IEEEbiography}
[{\includegraphics[width=1in,height=1.25in,clip,keepaspectratio]{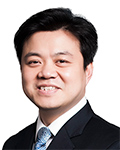}}]{Yiqiang Chen}
received the B.S. and M.S. degrees in computer science from Xiangtan University, Xiangtan, China, in 1996 and 1999, respectively, and the Ph.D. degree in computer science from the Institute of Computing Technology (ICT), Chinese Academy of Sciences (CAS), Beijing, China, in 2003. In 2004, he was a visiting scholar researcher with the Department of Computer Science, Hong Kong University of Science and Technology, Hong Kong. He is currently a professor and the director of the Research Center for Ubiquitous Computing Systems at the ICT, CAS. His research interests include artificial intelligence, pervasive computing, and human-computer interaction.
\end{IEEEbiography}

\begin{IEEEbiography}
[{\includegraphics[width=1in,height=1.25in,clip,keepaspectratio]{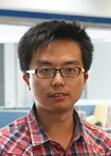}}]{Yunbing Xing}
received the B.S. and M.S. degrees in computer science from Northwestern Polytechnical University, Xi'an, China. He is a senior engineer at the Institute of Computing Technology, Chinese Academy of Sciences. His research interests include graphics rendering, video coding, and perceptual computing.
\end{IEEEbiography}

\begin{IEEEbiography}
[{\includegraphics[width=1in,height=1.25in,clip,keepaspectratio]{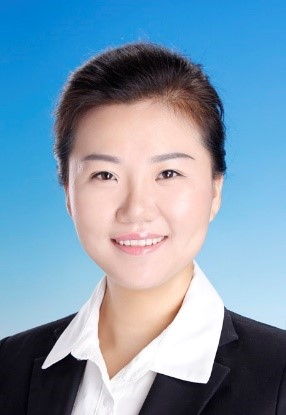}}]{Yang Gu} received the B.S. in computer science from Beijing University of Posts and Telecommunications, China in 2010, and Ph.D. degree in computer science from the Institute of Computing Technology (ICT), Chinese Academy of Sciences (CAS), Beijing, China, in 2016. She is currently an associate professor in the Research Center for Ubiquitous Computing Systems at ICT, CAS. Her research interests include intelligent sensing and digital health.
\end{IEEEbiography}

\begin{IEEEbiography}
[{\includegraphics[width=1in,height=1.25in,clip,keepaspectratio]{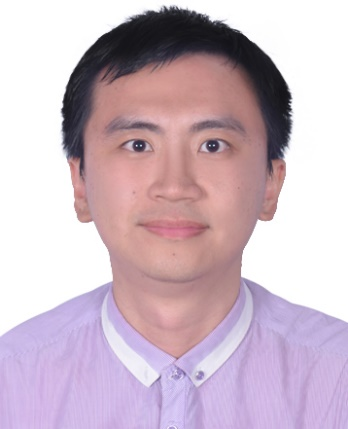}}]{Xiangyuan Lan} received the B.Eng. degree in computer science and technology from the South China University of Technology, China, in 2012, and the Ph.D. degree from the Department of Computer Science, Hong Kong Baptist University, Hong Kong in 2016, where he is currently a Research Assistant Professor. His current research interests include intelligent video surveillance and biometric security.
\end{IEEEbiography}

\vfill

\end{document}